\documentclass{article} 
\usepackage{collas2025_conferenc,times}
\usepackage{easyReview}

\usepackage{amsmath,amsfonts,bm}

\def\figref#1{figure~\ref{#1}}

\def\twofigref#1#2{figures \ref{#1} and \ref{#2}}
\def\threefigref#1#2#3{figures \ref{#1}, \ref{#2}, and \ref{#3}}

\def\secref#1{section~\ref{#1}}
\def\Secref#1{Section~\ref{#1}}

\def\appref#1{appendix~\ref{#1}}

\def\eqref#1{equation~\ref{#1}}

\def\Algref#1{Algorithm~\ref{#1}}

\def\1{\bm{1}}

\def\rb{{\textnormal{b}}}

\def\vx{{\bm{x}}}
\def\vy{{\bm{y}}}
\def\vz{{\bm{z}}}

\DeclareMathAlphabet{\mathsfit}{\encodingdefault}{\sfdefault}{m}{sl}
\SetMathAlphabet{\mathsfit}{bold}{\encodingdefault}{\sfdefault}{bx}{n}

\def\gE{{\mathcal{E}}}

\def\gM{{\mathcal{M}}}

\def\gR{{\mathcal{R}}}

\def\sA{{\mathbb{A}}}
\def\sB{{\mathbb{B}}}

\def\sH{{\mathbb{H}}}

\def\sL{{\mathbb{L}}}

\def\sN{{\mathbb{N}}}
\def\sO{{\mathbb{O}}}

\def\sR{{\mathbb{R}}}
\def\sS{{\mathbb{S}}}

\def\sV{{\mathbb{V}}}

\def\sX{{\mathbb{X}}}

\newcommand{\E}{\mathbb{E}}

\usepackage{hyperref}
\hypersetup{
    colorlinks=true,
    linkcolor=red,
    filecolor=magenta,
    urlcolor=blue,
    citecolor=purple,
    pdftitle={Overleaf Example},
    pdfpagemode=FullScreen,
    }

\usepackage{graphicx}
\usepackage{caption}
\usepackage{subcaption}
\usepackage{array}
\usepackage{graphicx}
\usepackage{booktabs}
\usepackage{fvextra}
\usepackage{amsmath}
\usepackage{amssymb}
\usepackage{stmaryrd}
\usepackage{multirow}
\usepackage{multicol}
\usepackage{cleveref}
\usepackage{pgf}
\usepackage{pgfplots}
\usepackage{algorithm}
\usepackage[noend]{algpseudocode}
\usepackage{wrapfig}
\usepackage{tcolorbox} 

\usepackage{mathtools}

\title{SECURE: Semantics-aware Embodied Conversation \\ under Unawareness for Lifelong Robot Learning}

\author{Rimvydas Rubavicius, Peter David Fagan, Alex Lascarides, Subramanian Ramamoorthy \\
School of Informatics\\
The University of Edinburgh\\
United Kingdom \\
\texttt{\{rimvydas.rubavicius,p.d.fagan,a.lascarides,s.ramamoorthy\}@ed.ac.uk}  
}

\collasfinalcopy 

\begin{document}

\maketitle

\begin{abstract}
This paper addresses a challenging interactive task learning scenario we call rearrangement under unawareness: an agent must manipulate a rigid-body environment without knowing a key concept necessary for solving the task and must learn about it during deployment. For example, the user may ask to ``put the two granny smith apples inside the basket'', but the agent cannot correctly identify which objects in the environment are ``granny smith'' as the agent has not been exposed to such a concept before. We introduce SECURE, an interactive task learning policy designed to tackle such scenarios. The unique feature of SECURE is its ability to enable agents to engage in semantic analysis when processing embodied conversations and making decisions. Through embodied conversation, a SECURE agent adjusts its deficient domain model by engaging in dialogue to identify and learn about previously unforeseen possibilities. The SECURE agent learns from the user's embodied corrective feedback when mistakes are made and strategically engages in dialogue to uncover useful information about novel concepts relevant to the task. These capabilities enable the SECURE agent to generalize to new tasks with the acquired knowledge. We demonstrate in the simulated Blocksworld and the real-world apple manipulation environments that the SECURE agent, which solves such rearrangements under unawareness, is more data-efficient than agents that do not engage in embodied conversation or semantic analysis.\footnote{Project website: \url{https://assistive-autonomy.github.io/secure}}
\end{abstract}

\section{Introduction}
\label{sec:introduction}

\begin{figure*}[ht]
    \centering
    \includegraphics[width=\textwidth]{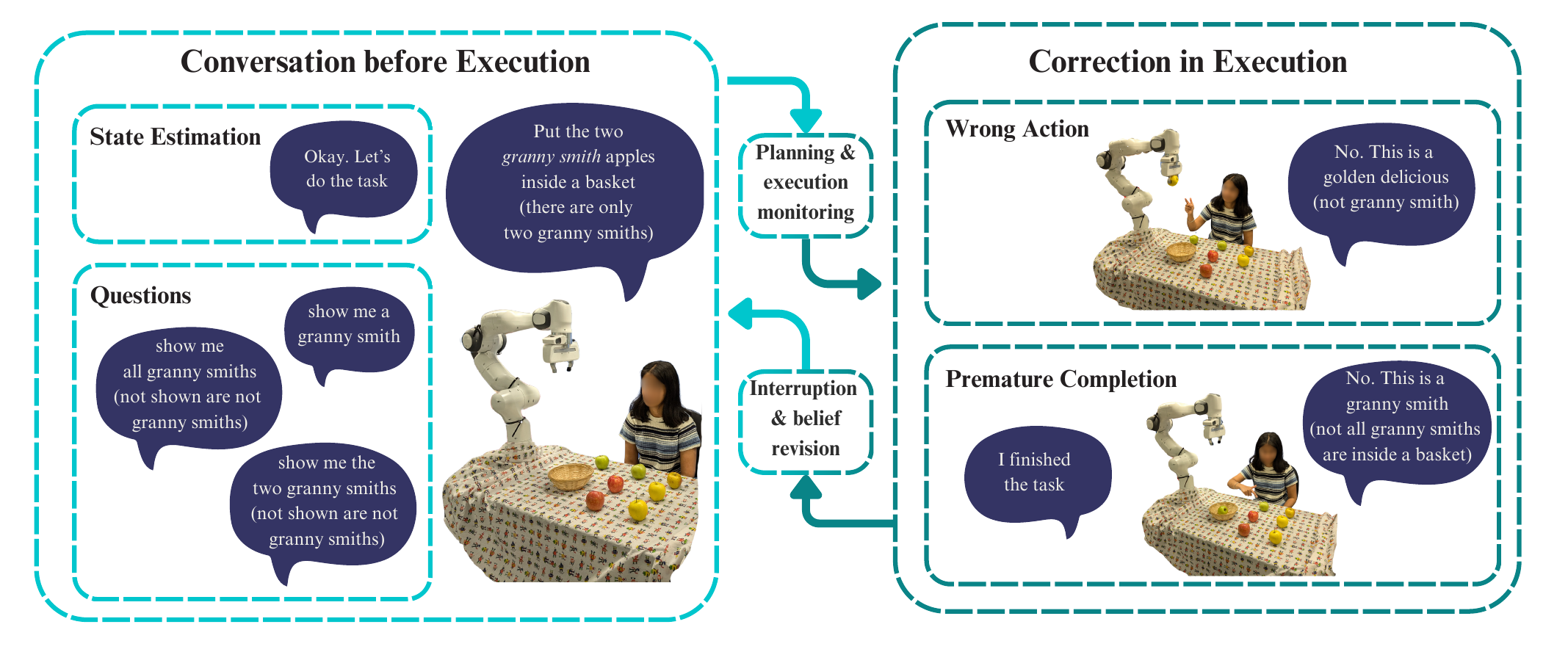}
    \caption{Framework overview. The agent interprets embodied conversation with the user in order to update its beliefs and solve tasks under unawareness: that is, an expression like ``granny smith'', which is a part of the user's instruction, is not in the agent's vocabulary, and the concept ``granny smith'' denotes is not a part of the agent's hypothesis space of possible domain models.  Our framework enables the agent to exchange embodied messages with the user before attempting to solve the task; the expert can also provide feedback on the agent's actions in the environment during task execution. For conversation before execution, the agent issues questions (e.g., ``show me a granny smith'') to reduce its uncertainty about the domain. When executing the task, the user responds to a suboptimal action with embodied corrective feedback, e.g., ``No. This is a golden delicious''. Such corrections can occur in case of a wrong action (e.g., picking ``golden delicious'' instead of ``granny smith'') or premature completion (declaring that the goal state is reached when it is not.). Such feedback exposes the agent's false beliefs (being confident but wrong about the state).  It triggers execution interruption and belief revision. In both cases, our framework processes embodied conversation in a semantics-aware manner: the user's messages are augmented with their logical consequences (shown in brackets in this figure). They are used to update the agent's beliefs, which in turn affects their decision-making.}
    \label{fig:overview}
\end{figure*}

A central theme in human-robot interaction~\citep{hri-book} is to use embodied conversation~\citep{DBLP:journals/aim/Cassell01} to instruct and teach robots to perform a variety of tasks~\citep{shridhar2021cliport,saycan2022arxiv,rt22023arxiv}. In this pursuit, models for vision, language, and action have been developed to design generalist robotic agents~\citep{shridhar2021cliport,saycan2022arxiv,DBLP:conf/corl/ZitkovichYXXXXW23,octo_2023}. One challenge for these agents is the interactive task learning (ITL) scenario~\citep{DBLP:journals/expert/LairdGAFJLSSTTW17}, in which the task instruction features a {\em neologism} (i.e., a newly coined expression) that denotes a concept the agent is unaware of but must understand to solve the task. For example, consider the scenario in \figref{fig:overview}. The user instructs the robot to ``Put the two granny smith apples inside a basket''.  But before deployment, the robot was not trained on and cannot correctly ground the concept ``granny smith'', let alone distinguish objects that are ``granny smith'' from other objects, including other kinds of apple, such as ``golden delicious''. Achieving such a fine-grained categorisation~\citep{DBLP:journals/pami/WeiSAWPTYB22} with limited prior experience might not be sufficiently captured in advance~\citep{DBLP:journals/corr/abs-2404-04125}. For instance, Grounding DINO~\citep{DBLP:journals/corr/abs-2303-05499} can accurately detect a variety of objects in the environment, including an excellent ability to localize fruits. However, in our real-world experiments, grounding DINO does not work out-of-the-box when asked to localize objects denoted by the phrase ``granny smith'' or ``granny smith apple'', presumably because it completely lacked any examples labelled ``granny smith'' during training. To cope with this scenario, the agent must update their domain conceptualisation with this unforeseen concept, thereby expanding the set of possible states, which in turn demands revisions to planning about subsequent interactions with the user and the environment.

More generally, the task in \figref{fig:overview} uses sensory observations to effect a rearrangement of rigid bodies into a state that satisfies the specified goal~\citep{DBLP:journals/corr/abs-2011-01975}. In this work, we consider a special case of rearrangement, in which the goal is given in a natural language instruction and the agent is not aware of at least one concept required to understand it. We refer to such tasks as {\em rearrangement under unawareness}. To solve them, the agent can engage in an {\em embodied conversation} with the user to learn {\em online} how to recognise unforeseen concepts by interactively grounding them. This requires incremental grounding (belief updates should occur with every user's message) and coping with an expanding hypothesis space (the dialogue may reveal unforeseen concepts). Consequently, the agent: (a) cannot assume mutual exclusivity between concepts; (b) must detect deficient beliefs about the hypothesis space; and (c) on discovering such deficiency, must adapt its prior beliefs to the newly expanded hypothesis space.

A principal way to process embodied conversation is to parse its utterances into a logical form in a symbolic logic that supports~{\em automated inference} \citep{DBLP:books/daglib/0031956}. This work studies how logical reasoning can enhance solving rearrangements under unawareness through interaction. We aim to make the agent {\em semantics-aware}: that is, besides the evidence acquired from directly processing the user's utterances, the agent also utilizes the {\em logical consequences} of the parsed output to aid learning and inference. Previous work showed the value of logical consequences in ITL stemming from the semantic analysis of individual messages~\citep{rubavicius-lascarides-2022-interactive,park-etal-2023-interactive} (e.g., ``put the two granny smiths\dots'' entails that there are two and only two objects that are ``granny smith'') as well as the semantic analysis of the overall discourse---in particular corrective feedback~\citep{appelgren:lascarides:2020} (e.g., in \figref{fig:overview}, the user highlights a mistaken pick, which implies that the picked object is not ``granny smith''). 
However, they don't \textit{combine} sentence and discourse semantics and lack state-of-the-art perception and testing in the real world.

To rectify this gap in the existing work, we develop the SECURE ITL framework---\textbf{S}emantics-aware \textbf{E}mbodied \textbf{C}onversation under \textbf{U}nawareness for Lifelong \textbf{R}obot L\textbf{e}arning. SECURE combines symbolic reasoning and neural grounding to process the embodied conversation. The agent continuously updates its beliefs, and hence its plans, as and when the user provides information through their dialogue.  It can discover unforeseen possibilities through the information exchange in the conversation, at which point SECURE revises its domain conceptualisation, triggering further belief updates and revised plans.  SECURE also learns {\em how} to engage in conversation: that is, it learns how to make decisions that resolve the dilemma between asking questions (which comes at a cost) vs.\ risking solving the task using current beliefs, for which failure is very costly. Furthermore, our agent can process the user's embodied corrective feedback, in response to the mistakes it makes when attempting to solve the task: its interpretation of correction serves to revise (false) beliefs, and furthermore may compel it to engage in further conversation before taking further actions in the environment. To quantify the value of semantic awareness, we evaluate SECURE in a simulated blocksworld and in a real-world apple manipulation on several instances of rearrangement under unawareness. The results show that augmenting the users' messages with their logical consequences aids ITL.

\section{Framework}
\label{sec:framework}

This section describes our ITL framework, including its components for knowledge representation and reasoning (\secref{sec:krr}), interactive symbol grounding (\secref{sec:grounding}), and processing of embodied conversation (\secref{sec:processing_embodied_conversation}).

\subsection{Knowledge Representation and Reasoning}
\label{sec:krr}

The agent and the user exchange information using embodied utterances. These utterances consist of natural language expressions~\(\sN\) and optional pointing gestures to objects~\(\sO\) in the environment. Each natural language expression is mapped to its logical form~\(\sL\) (see section~\ref{sec:processing_embodied_conversation}), which supports automated model checking and inference.

SECURE uses a finite first-order domain model \(\gM\)~\citep{gamut-intro-to-logic-volume-1} to reason about the environment. \(\gM\) is a triple: a set of objects \(\sO\); a set of predicates, a.k.a. the vocabulary \(\sV\); and an interpretation function \(I\colon \sV\mapsto \sO^*\) that maps predicates to denotations (for 1-place predicates it's a set of objects, for 2-place predicates a set of pairs of objects, and in general for \(n\)-place predicates a set of \(n\)-tuples of objects). The agent can evaluate statements in a natural language like ``\(o_1\) is a granny smith''\(\in\sN_{\mathrm{snt}}\), which has the logical form~\(\mathtt{grannysmith(o_1)}\in\sL_{\mathrm{snt}}\), to be true or false with respect to \(\gM\) using truth-conditional semantics. It can also identify objects denoted by various referential expressions \(r\in\sN_{\mathrm{ref}}\): e.g.,  ``the two granny smiths'', which has the logical form \(\Phi(r)={\tt \langle \_the\_2\_q\text{ }x.grannysmith(x)\rangle}\in\sL_{\mathrm{ref}}\), is evaluated against $\gM$ to yield either the empty set (reference failure) or a set consisting of the (unique) set of two objects in $\gM$ that are in the range of $I(\mathtt{grannysmith})$. We detail this (logical) reference semantics in \appref{app:logic}. The agent uses the identified objects to understand the task instruction \(t\in\sN_{\mathrm{task}}\) and, in turn, solve the task (e.g., from \figref{fig:overview}, when the two objects are identified as ``granny smith'', they are picked and placed in the basket.) The set of atomic predications (atoms) constructed from predicates followed by the appropriate number of constants that denote objects in the domain \(\sO\) is a set of ground atoms also known as Herbrand base \(\sH\).  The set of atoms that are entailed by \(\gM\) is denoted as \(\sH_{\gM} =\{\mathtt{a}\in\sH\mid\gM\models\mathtt{a}\}\).

The agent estimates the domain model~\(\hat{\gM}\) using its belief state \(b\in\sB\)~\citep{BOCHMAN2007557,algobook}.  This maintains knowledge that the agent has acquired throughout its lifecycle from various sources of evidence, including percepts, memory, and priors. Percepts are constructed from the observations of the current situation. They include a set of objects in the environment~\(\sO\), with corresponding object-centric embeddings~\(\vx\in\sR^{d}\) that are extracted from sensory observations in the environment and a domain theory~\(\Delta\) that is constructed (and updated) by processing the embodied conversation. Memory is a dynamic record of the agent's experience in its lifecycle. It includes the vocabulary~\(\sV\) and support~\(\sS\), which is a set of pairs consisting of an object-centric embedding~\(\vx\in\sR^d\) and the accompanying labels~\(\vy\in [0,1]^{|\sV|}\) with \(y^{(\mathtt{p})}\), indicating the degree of belief that an object with the corresponding embedding is denoted by the symbol \(\mathtt{p}\in\sV\). Priors that determine the initial beliefs of each atom are given by prior weights~\(w_p\colon\sH\mapsto[0,1]\),  parameterizing the Bernoulli random variable~\(\mathrm{Bern}\) for each atom. They can come from the initial estimation based on sensory observations (as in \figref{fig:overview}) or, in the case of pure unawareness, they are assigned \(0.5\) as the {\em default weight}, to capture complete ignorance about denotations of the symbol in question. Using the belief state, the agent can reason about the domain and make decisions. To do this, evidence is combined into {\em grounded weights} \(w_{g}\colon\sH\mapsto[0,1]\) that parameterise a probability distribution over domain models. In particular, using weights (both \(w_p\) and \(w_g\)), a probability of a well-formed formula \(\phi\in\sL_{\mathrm{snt}}\) can be computed using weighted model counting \(\mathrm{WMC}\)~\citep{DBLP:journals/ai/ChaviraD08} that counts and weights events (instances of \(\gM\)) that are valid under constraints (\(\gM\) entails \(\phi\)):
\begin{align}
\label{eq:wmc}
\mathrm{WMC}_{w}(\phi)  &= \sum_{\gM\colon \gM\models \phi } \text{ 
 }\prod_{\mathtt{a}\colon\mathtt{a}\in\sH} w(\mathtt{a})\cdot\1_{\mathtt{a}\in\sH_{\gM}} + (1-w(\mathtt{a}))\cdot(1-\1_{\mathtt{a}\in\sH_{\gM}} )
\end{align}
\(\mathrm{WMC}\) is used to compute various (probabilistic) queries given the current belief state, such as the conditional probability~\(\mathrm{CON}\) of a formula~\(\phi\), and the most probable domain model (MAP estimate):
\begin{align}
\label{eq:queries}
 \mathrm{CON}_{w}(\phi\mid b) = \frac{\mathrm{WMC}_{w}(\phi\wedge\tilde{\Delta})}{\mathrm{WMC}_{w}(\tilde{\Delta})}  \quad \hat{\gM} =\arg\max_{\gM} \mathrm{CON}_{w}(\gM\mid b) 
\end{align}
where \(\tilde{\Delta}\) is a conjunction of formulas in the domain theory~\(\Delta\) of the belief state~\(b\) and \(\mathrm{CON}_{w}(\gM\mid b)\) is the probability of \(\gM\)  given by the conditional probability of the conjunction of model atoms based on them being in \(\sH_{\gM}\), or not.

\subsection{Interactive Symbol Grounding}
\label{sec:grounding}

The grounding model is a function from sensory observations to atoms that hold in the domain.\footnote{We only consider one-place predicate grounding here. In principle, the grounding model could be extended to predicates of arbitrary arity. For instance, for two-place predicates involving spatial relationships, we could consider embeddings constructed from the pair of embeddings of the objects participating in this relationship.} It is operationalised as a probabilistic classifier~\(\omega_{b}\colon\sR^d\mapsto[0,1]^{|\sV|}\) that depends on the belief state \(b\in\sB\). Given an embedding \(\vx\in\sR^{d}\), corresponding to an object \(o\in\sO\), which is extracted from the environment's sensory observations, \(\omega_{b}\) predicts binary labels \(\hat{\vy}\in[0,1]^{|\sV|}\) with each element \(\hat{y}^{(\mathtt{p})}\) indicating a probability that an atom \(\mathtt{p(o)}\) constructed from predicate \(\mathtt{p}\) and the constant \(\tt o\) that denotes the object $o$ with an embedding \(\vx\) is a member of \(\sH_{\gM}\) (equivalent to the probability that ground-truth domain model $\gM$ entails $\mathtt{p(o)}$ based on the current belief state and object-centric embedding):
\begin{align*}
\hat{\vy} = \omega_{b}(\vx) \quad \hat{{y}}^{(\mathtt{p})} = P(\mathtt{p}(\mathtt{o})\in \sH_{\gM} \mid b)
\end{align*}

For interactive symbol grounding, SECURE uses a non-parametric grounding model of multilabel prototype networks~\citep{DBLP:journals/corr/abs-1911-07203,cano-santin-etal-2020-fast}. This uses the support~\(\sS\) of the belief state~\(b\) to make predictions. The unique approach of SECURE lies in using the domain theory~\(\Delta\), which is built by processing the latest user's (embodied) utterance to update \(\sS\) and, in turn, update the belief state. In particular, whenever the agent constructs a new well-formed formula~\(\phi\in\sL_{\mathrm{snt}}\), the belief updates~\(\mathrm{Update}\colon\sB\times\sL\mapsto\sB\) with new support constructed by updating the label values with the new conditional probabilities with respect to prior weights~\(\mathrm{CON}_{w_p}(\cdot\mid b)\) (see Appendix~\ref{app:algorithms} for details).

Using support~\(\sS\), the grounding model builds positive and negative support for each symbol \(\mathtt{p}\in \sV\), denoted as \(\sS^{(+\mathtt{p})}\) and \(\sS^{(-\mathtt{p})}\), respectively. Whether a pair goes into the support is decided by: the value of \(y^{(\mathtt{p})}\), the entropy of \(\mathrm{Bern(y^{(\mathtt{p})})}\), and the threshold~\(\tau\) for the overall noise level in support building:
\begin{align*}
\begin{split}
\sS^{(+\mathtt{p})} = \{ (\vx,\vy)\in \sS \mid y^{(\mathtt{p})} > \frac{1}{2} \wedge H(\mathrm{Bern(y^{(\mathtt{p})})}) \leq \tau \} \qquad \sS^{(-\mathtt{p})} = \{ (\vx,\vy)\in \sS \mid y^{(\mathtt{p})} < \frac{1}{2} \wedge H(\mathrm{Bern(y^{(\mathtt{p})})}) \leq \tau \} 
\end{split}   
\end{align*}
Using \(\sS^{(+\mathtt{p})}\) and \(\sS^{(-\mathtt{p})}\), positive and negative prototypes, denoted as \(\vz^{(+\mathtt{p})}\) and \(\vz^{(-\mathtt{p})}\) respectively, are computed by taking the weighted average of embeddings of the corresponding supports with the weight being the likelihood that the relevant object is (or respectively is not) denoted by $\mathtt{p}$ (i.e., the values $y^{(\mathtt{p})}$ and $(1-y^{(\mathtt{p})})$ respectively):
\begin{align*}
\begin{split}
\vz^{(+\mathtt{p})} = \frac{1}{|\sS^{(+\mathtt{p})}|}\sum_{(\vx,\vy)\in \sS^{(+\mathtt{p})}} y^{(\mathtt{p})}\vx \qquad \vz^{(-\mathtt{p})} = \frac{1}{|\sS^{(-\mathtt{p})}|}\sum_{(\vx,\vy)\in \sS^{(-\mathtt{p})}} (1-y^{(\mathtt{p})})\vx
\end{split}
\end{align*}
If, due to a lack of evidence (so far), the positive/negative support is empty---there is not enough evidence (yet) to deem any of the observed embeddings to be {\em good enough}---then positive/negative prototypes default to support elements with the largest/smallest entropy as the best guess of what exemplars are suitable for positive/negative support.

Using positive/negative prototypes, the probability that predicate \(\mathtt{p}\) is true of an object with embedding~\(\vx\) is computed using their cosine distance as an activation for a sigmoid:~\(\hat{y}^{(\mathtt{p})} = \sigma(\cos(\vz^{(-\mathtt{p})}-\vz^{(+\mathtt{p})},\vx))\). Grounding model predictions yield grounded weights \(w_g(\mathtt{p(o)}) \leftarrow \omega_b(\vx)^{(\mathtt{p})}\) that can be used for reasoning (\secref{sec:krr}).

The presented grounding model handles tasks under unawareness in the following way. Suppose that the agent observes a new symbol or neologism \(\mathtt{p}^*\) from the latest user's message. The agent becomes aware of \(\mathtt{p}^*\) by extending its vocabulary \(\sV \leftarrow \{\mathtt{p}^*\} \cup \sV\) in the belief state and by updating the belief state---in particular, the dimensionality of \(\vy\) increases by populating it with prior weights \(w_p\), which in case of pure unawareness is the default weight 0.5.

\subsection{Processing Embodied Conversation}
\label{sec:processing_embodied_conversation}

In processing embodied conversation, the agent performs semantic analysis (\secref{sec:semantic_analysis}), which contributes to decisions about action before attempting the task (\secref{sec:conversation_before_exectution}) and to understanding the user's corrections (\secref{sec:correction_in_execution}).

\subsubsection{Semantic Analysis}
\label{sec:semantic_analysis}

When the user utters an embodied signal, its semantic analysis yields a well-formed formula \(\phi\in\sL_{\mathrm{snt}}\), which gets added to $\Delta$ and triggers a belief update.  Semantic analyses include: sentence-level~\(\xi\colon\sN_{\mathrm{ref}}\times2^{\sO}\mapsto\sL_{\mathrm{snt}}\), stemming from generalised quantifier theory~\citep{Barwise1981GeneralizedQA}; and discourse-level~\(\zeta\colon\sN_{\mathrm{task}}\times\sN_{\mathrm{corr}}\times\sA\mapsto\sL_{\mathrm{snt}}\), stemming from coherence-based theories of discourse, in particular coherent corrections~\citep{lascarides:asher:2009}.

The sentence-level semantic analysis \(\phi\in\sL_{\mathrm{snt}}\) is constructed from a referential expression \(r\in\sN_{\mathrm{ref}}\) and objects in the environment denoting it known as the referent \(\gR\subseteq2^{\sO}\). This is a subset of \(2^\sO\), with each element being a set of objects denoted by the referential expression. E.g., when 3 objects are ``granny smith'' the referent of ``two granny smiths'' is a set of 3 elements, each containing 2 out of the 3 objects that are ``granny smith''. 
\(\phi\) includes clauses deduced from the surface form of \(r\) as well as negated clauses from the logical consequence of using a quantifier~\citep{BEAVER1997939}. For instance, if the user utters the referential expression \(r=\)``the one granny smith'' and points to $o$, this leads to the formula \(\xi(r, \gR)  = \mathtt{grannysmith(o)} \wedge \bigwedge_{o'\in\sO-o}\neg\mathtt{grannysmith(o')}\) getting added to $\Delta$.  The deduced negated clauses follow from the uniqueness condition on the quantifier ``the one''~\citep{Russell1917-RUSKBA-4}, which entails that no other object in the environment is denoted by the phrase ``granny smith''. 
 
For discourse-level semantic analysis, following~\citet{lascarides:asher:2009}, the user's correction~\(c\in\sN_{\mathrm{corr}}\) in response to the agent's action \(a\in\sA\) has two effects on the agent's beliefs: 1) the content of $c\in\sN_{\mathrm{corr}}$ is true; and 2) \(a\in\sA\) is not part of an optimal plan, which leads to further deductions that are dependent on the task \(t\in\sN_{\mathrm{task}}\) and what type of action $a$ is. \Secref{sec:correction_in_execution} looks in detail at which well-formed formulae are inferred and added to $\Delta$. 

\subsubsection{Conversation before Execution} 
\label{sec:conversation_before_exectution}

When solving a task under unawareness, the interaction begins with the user uttering a task instruction in natural language~\(t\in\sN_{\mathrm{task}}\). Before trying to solve the task using their (current) belief, this option being a sequence of $a_{\mathrm{act}}$ from the set of possible executable actions \(\sA_{\mathrm{exec}}\) that change the environment, the agent can choose instead to ask a question~\(a_{\mathrm{quest}}\in\sA_{\mathrm{quest}}\subset\sA\), to reduce uncertainty about the domain model $\hat{{\cal M}}$. The questions are of the form ``show me \(r\)'' where \(r\in\sN_{\mathrm{ref}}\) is a referential expression. The user responds to such questions by pointing to a referent or $r$, which, using sentence-level semantic analysis \(\xi\), yields a well-formed formula \(\phi\in\sL_{\mathrm{snt}}\) that triggers a belief update. \(\phi\) is not known in advance to the agent and can be conceptualized as a random variable parameterized by the agent's \(\phi\sim\mathrm{Result}(a)\). To maintain coherence, \(\sA_{\mathrm{quest}}\) depends on the task instruction: specifically, \(r\) in the question can feature only non-logical words used in the instruction, and its quantifier can be replaced with an existential or universal. For example, in the case of \(t=\)``move every red cylinder to the left of the one cube'', the agent can ask to show ``every red cylinder'', ``the one cube'', ``a red cylinder'', ``a cube'' and ``all cubes'', but not for instance ``a red cube'' as this referential expression may not have a referent in the environment.

To make strategic decisions about whether and how to engage in the embodied conversation, SECURE solves the following decision problem. Each agent's action (\(\sA = \sA_{\mathrm{quest}} \cup \sA_{\mathrm{exec}}\)) results in a well-formed formula \(\phi\in\sL_{\mathrm{snt}}\) being observed, followed by the belief update \(\mathrm{Update}(b,\phi)\). To quantify the value of each action, we take an information-theoretic approach and measure the {\em expected information gain}~\(I\colon\sB\times\sA\mapsto\sR^+\)~\citep{lindley1956measure, 10.5555/1121596} for actions: 
\begin{align}
\begin{split}
\label{eq:information_gain}
    I(b,a) =  H(b) - \E_{\phi\sim\mathrm{Result}(a)}[H(\mathrm{Update}(b,\phi))]    
\end{split}
\end{align}
The entropoy of belief states is defined using grounded weights~\(w_{g}\) to parameterise the \(\mathrm{Bern}\) random variable for each atom:
\begin{align}
\label{eq:belief_entropy}
    H(b) \coloneqq \sum_{\mathtt{a}\in\sH} H(\rb)\quad \rb\sim\mathrm{Bern}(w_g(\mathtt{a}))
\end{align}
The expectation in \eqref{eq:information_gain} is computed by marginalising over answers \(\phi\) consistent with \(b\), with the likelihood that \(\phi\) is the actual answer computed using conditional probabilities with grounded weights (\eqref{eq:queries}).

The value of information, of performing certain actions in the environment is in a trade-off with rewards of taking these actions. Questions have an inherent cost \(C\colon\sA_{\mathrm{quest}}\mapsto\sR^+\), which approximates the user's answering effort. It depends on two quantities: \(\mathrm{Obj}(r)\) that is the number of objects in the referent of \(r\) approximating the designation (pointing) effort, and \(\mathrm{Sym}(r)\) that is the number of predicates in \(\Phi(r)\), approximating reference resolution effort. These quantities are weighted by unit pointing cost \(C_{\mathrm{point}}\) and unit reference cost \(C_{\mathrm{ref}}\), resulting in the overall cost:
\begin{align}
\label{eq:cost}
C(q) = C_{\mathrm{point}}\mathrm{Obj}(r) + C_{\mathrm{ref}}\mathrm{Sym}(r)\quad r\in a_{\mathrm{quest}}
\end{align}
For example, question \(q=\)``show me the one red cube'' cost is \(C_{\mathrm{point}}+2C_{\mathrm{ref}}\) as its referential expression (\(r=\)``the one red cube'') features two predicates and answering involves pointing to one object. 

For action \(a_{\mathrm{act}}\in\sA\) and successful attempt to solve the task the reward of \(1\) is observed and \(-1\) otherwise, resulting to the following reward function \(R\colon\sA\times\sB\mapsto[-1,1]\):
\begin{align}
\label{eq:query-reward}
        R(a) = \begin{cases}
            1   &\text{ if } a=a_{\mathrm{act}} \text{ and task is solved correctly} \\
            -C(a) &\text{ if } a\in \sA_{\mathrm{quest}} \\
            -1  &\text{ otherwise } 
        \end{cases}
\end{align}

SECURE compares expected information gain to the expected reward of the current belief state:
\begin{align}
\label{eq:exp_reward}
    \E_b[R(a)] =
    \begin{cases}
    -\tilde{C}(a) &\text{if }  a\in \sA_{\mathrm{quest}} \\    
    2\cdot\mathrm{CON}_{w_g}{(\gM \mid b)} -1 &\text{otherwise}
    \end{cases}
\end{align}
where \(\tilde{C}\) is just like \(C\) except different for the universal quantifier as the number of objects in a referent is not known in advance as it ranges from 1 to \(|\sO|\), thus is approximated by the average, i.e., \(\frac{|\sO|-1}{2}\).

This leads to the overall action-value function \(Q\colon\sB\times\sA\mapsto\sR\): 
\begin{align}
\label{eq:Q-values}
Q(b,a) = \theta_1 I(b,a) + \theta_2 \E_{b}[R(a)] = \begin{bmatrix} \theta_1 \\ \theta_2 \end{bmatrix} \cdot \begin{bmatrix} I(b,a) \\ \E_{b}[R(a)] \end{bmatrix} = \theta^\top h(b,a)
\end{align}
where \(\theta=[\theta_1,\theta_2]^{\top}\) are parameters signifying preferences in exploration (engaging in embodied conversation) and exploitation (solving task under unawareness) and \(h(a,b)\coloneqq[I(b,a), \E_{b}[R(a)]]^{\top}\) known as preference~\citep{DBLP:books/lib/SuttonB98}. Action-value function~\(Q\) parameters \(\theta\) can be optimized using semi-gradient SARSA. See \appref{app:algorithms} for details.

\subsubsection{Correction in Execution}
\label{sec:correction_in_execution}

Upon choosing \(a_{\mathrm{act}}\in\sA\), planning with \(\hat{\gM}\) is used to compute the sequence of execution actions. For rearrangement, three execution actions \(\sA_{\mathrm{exec}}\subset\sA\) are considered: \(a_{\mathrm{pick}}\), for picking an object in the environment, \(a_{\mathrm{place}}\) for placing the picked object in the target location and \(a_{\mathrm{complete}}\) for declaring that the agent has solved the task.

When executing these actions, the agent can make mistakes (suboptimal actions in the environment) due to an incorrectly estimated domain model $\hat{{\cal M}}$. The user reacts to a mistake by providing embodied corrective feedback. Each correction consists of the corrective cue ``No.'', followed by a referential expression and designation that highlights the source of error, e.g.  ``No. This is a cylinder. (points to \(o_{\mathrm{corr}}\in\sO\))''. As mentioned in \secref{sec:semantic_analysis}, semantic analysis of the correction depends on the task instruction~\(t\in\sN_{\mathrm{task}}\), the executed action~\(a\in\sA_{\mathrm{exec}}\), and the correction itself~\(c\in\sN_{\mathrm{corr}}\). To illustrate the different discourse-level semantic analyses, we consider the task instruction \(t=\)``move every cube in front of a cylinder''.

When the action \(a_{\mathrm{pick}}\in\sA_{\mathrm{exec}}\) gets corrected, semantic analysis dictates that the picked object should {\em not} have been picked: specifically, the picked object is not denoted by the direct object referential expression~\(r_{\mathrm{direct}}\) in the task instruction~\(t\in\sN_{\mathrm{task}}\). For instance, when \(\mathrm{pick}\) is corrected with \(c=\)``No. This is a cylinder. (points to \(o_{\mathrm{corr}}\in\sO\))'', semantic analysis dictates that \(\mathtt{cylinder(o_{\mathrm{corr}})}\) and \(\neg\mathtt{cube(o_{\mathrm{corr}})}\) are both true. In general, the semantic analysis for \(\mathrm{pick}\) is as follows:
\begin{equation*}
    \zeta(a_{\mathrm{pick}},t,c)  = \xi(r_{\mathrm{corr}},\{\{o_{\mathrm{corr}}\}\}) \wedge \neg \xi(r'_{\mathrm{direct}},\{\{o_{\mathrm{corr}}\}\}) 
\end{equation*}
where \(r'_{\mathrm{direct}}\) is just like \(r_{\mathrm{direct}}\) but with an existential quantifier.

When the action \(a_{\mathrm{place}}\in\sA_{\mathrm{exec}}\) gets corrected, semantic analysis is similar to the \(\mathtt{pick}\) case, except this time the object designated in the correction indicates that it is not part of the referent of the indirect object referential expression~\(r_{\mathrm{indirect}}\) in~\(t\). In our illustrative example, if the correction \(c=\)``No. This is a sphere. (points to \(o_{\mathrm{corr}}\in\sO\))'' is produced in response to \(a_{\mathrm{place}}\in\sA_{\mathrm{exec}}\), then the agent would (again) be able to infer \(\mathtt{sphere(o_{\mathrm{corr}})}\) directly from correction. But this time, the agent would also infer \(\neg\mathtt{cylinder(o_{\mathrm{corr}})}\) because placing the object must result in the placed object being in the intended spatial relation to the indirect object. In general, the semantic analysis for \(\mathrm{place}\) is as follows:
\begin{equation*}
    \zeta(a_{\mathrm{place}},t,c)  = \xi(r_{\mathrm{corr}},\{\{o_{\mathrm{corr}}\}\}) \wedge \neg \xi(r'_{\mathrm{indirect}},\{\{o_{\mathrm{corr}}\}\}) 
\end{equation*}
where \(r'_{\mathrm{indirect}}\) is a referential expression just like \(r_{\mathrm{indirect}}\) but with a existential quantifier.

When the action \(a_{\mathrm{complete}}\in\sA_{\mathrm{exec}}\) gets corrected, the correction entails that the goal state is {\em not} reached.  To illustrate, consider correction \(c=\)``No. This is a cube (points to \(o_{\mathrm{corr}}\in\sO\))''  utterend in response to \(\mathrm{complete}\).  The formula \(\mathtt{cube(o_{\mathrm{corr}})}\) is a part of the logical form of this message.  So the agent must reject any domain model for which \(\neg\mathtt{cube(o_{\mathrm{corr}})}\) is true, including its previous MAP estimate. Furthermore, the fact that this logical form is a correction entails that the current state does not satisfy the goal: in other words, there is a cube that is {\em not} in front of a cylinder. Treating the message ``this is a cube (points to $o_{\mathrm{corr}}\in\sO$)'' as an {\em explanation} of the corrective cue ``No'' entails that it is $o_{\mathrm{corr}}$ that is the renegade cube, not in front of a cylinder.\footnote{With the ground truth to spatial relations, the agent could infer from this that every object in front of $o_{\mathrm{corr}}$ is not a cylinder. SECURE foregos exploiting this entailment, however, for the sake of computational efficiency.} In general, the semantic analysis for \(\mathrm{complete}\) is as follows:
\begin{equation*}
    \zeta(a_{\mathrm{complete}},t,c) = \xi(r_{\mathrm{corr}},\{\{o_{\mathrm{corr}}\}\}) 
\end{equation*}

\section{Experiments}
\label{sec:experiments}

The proposed ITL framework is evaluated in simulation and real-world experiments on a Franka Emika Panda robot. We compare SECURE's policy $\pi_{\mathrm{secure}}$  against two baselines $\pi_{\mathrm{simple}}$ and $\pi_{\mathrm{correct}}$, which are as follows:
\begin{itemize}
    \item \(\pi_{\mathrm{secure}}\): the agent can engage in embodied conversation using semantic analysis to reason about both the literal interpretation and logical consequences of different actions. It can also process correction in execution;
    \item \(\pi_{\mathrm{simple}}\): the agent can engage in conversation before execution but does not compute any logical consequences: questions ``show me the one cube'' and ``show me a cube'' have the same expected information gain, and the same belief state after the user's response. It can also process correction in execution;
    \item \(\pi_{\mathrm{correct}}\): the agent cannot engage in embodied conversation but processes correction in execution.
\end{itemize}

Each policy performance is measured using three performance metrics: cumulative reward \(\mathrm{cR}\) and commutative cost \(\mathrm{cC}\) as an extrinsic metric of task success as well ass mean micro F1 score \(\mathrm{mF1}\) between ground truth and estimated referents at the beginning of each task for symbols in the task instruction, as an intrinsic metric for grounding. See \appref{app:experiments} for implementation details.

\begin{figure*}[ht]
\begin{subfigure}{0.25\textwidth}
  \centering
  \includegraphics[width=0.95\linewidth]{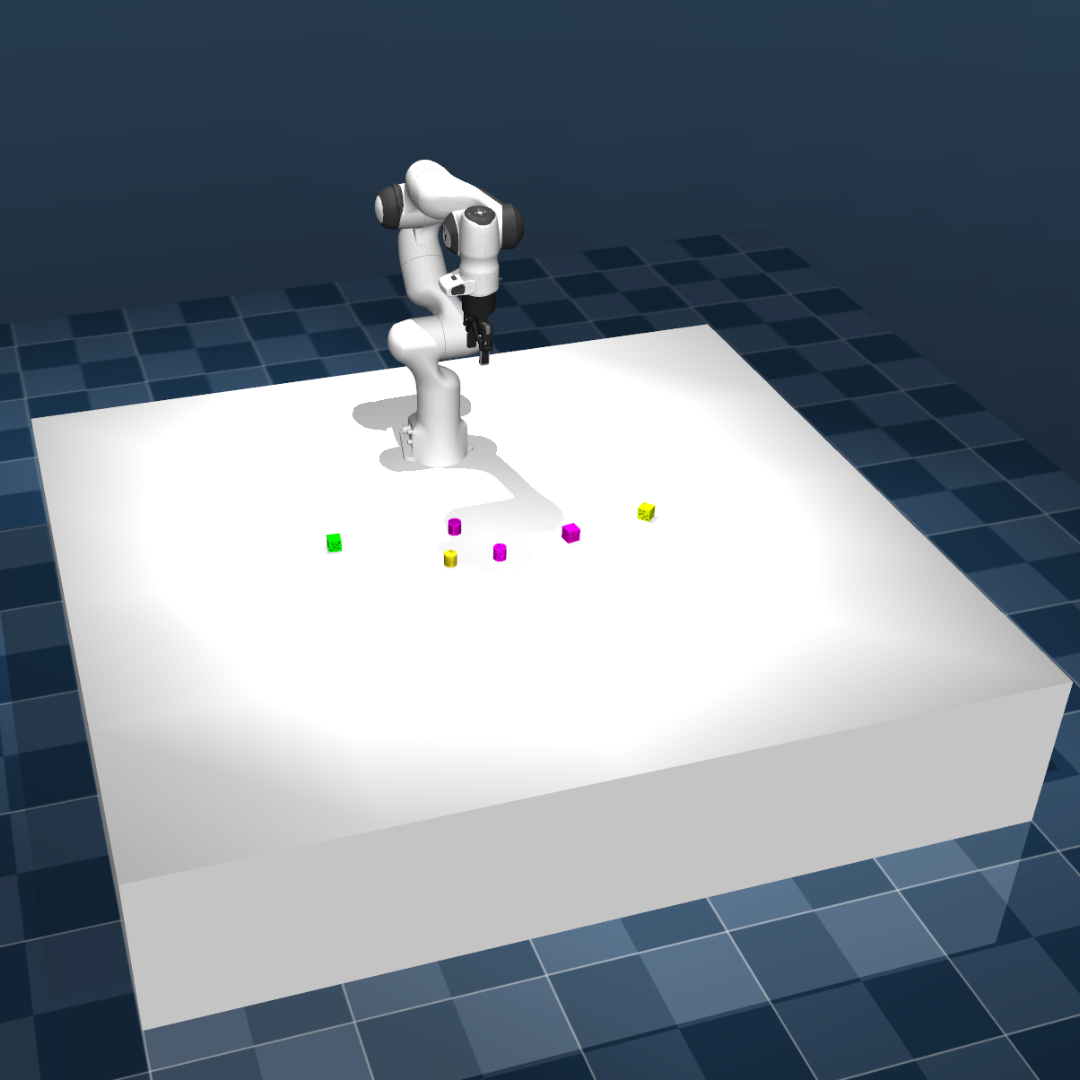}
  \caption{Bird-eye view}
  \label{fig:bird_eye}
\end{subfigure}%
\begin{subfigure}{0.25\textwidth}
  \centering
  \includegraphics[width=0.95\linewidth]{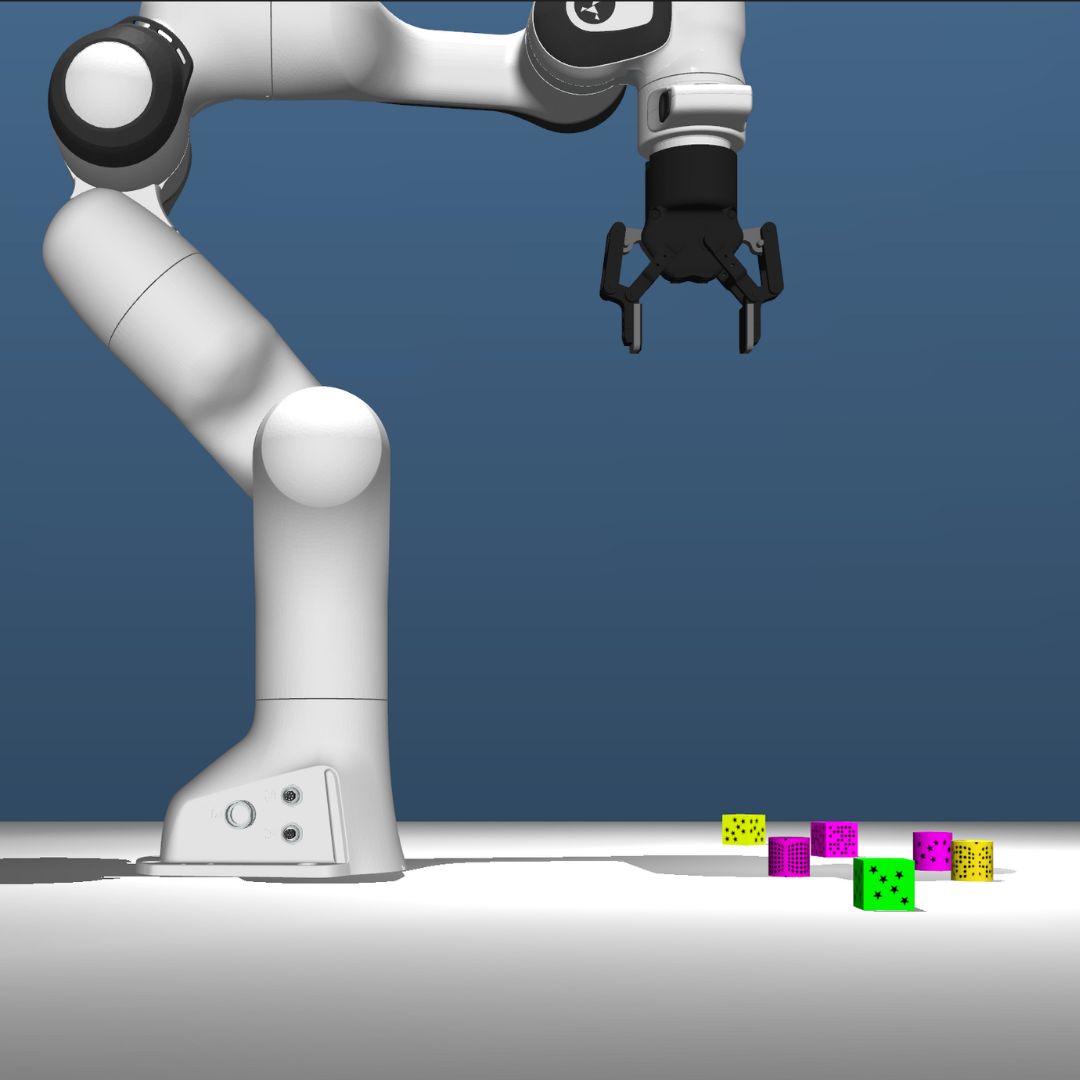}
  \caption{Side view}
  \label{fig:side_view}
\end{subfigure}%
\begin{subfigure}{0.25\textwidth}
  \centering
  \includegraphics[width=0.95\linewidth]{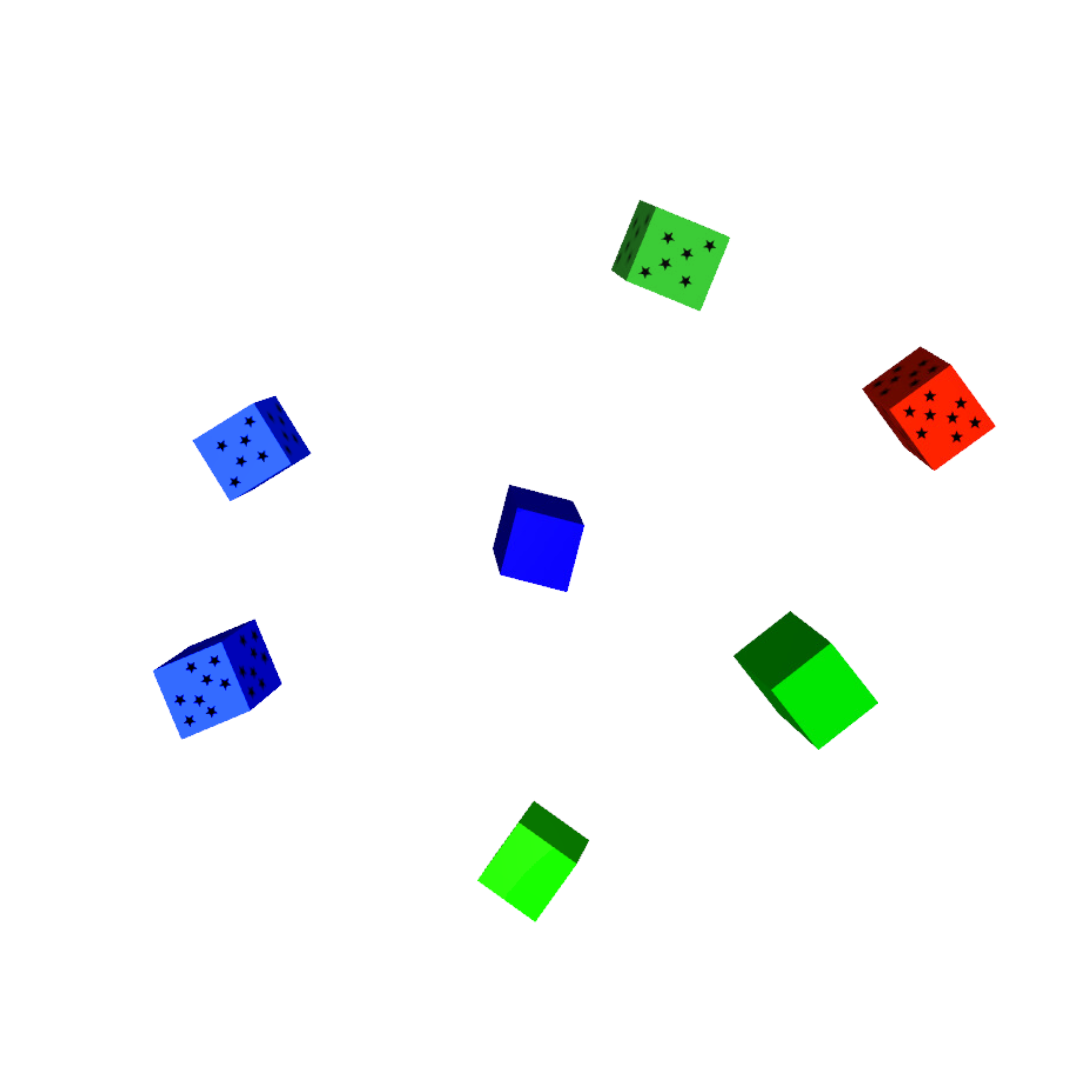}
  \caption{Top-down view}
  \label{fig:top_down}
\end{subfigure}%
\begin{subfigure}{0.25\textwidth}
  \centering
  \includegraphics[width=0.95\linewidth]{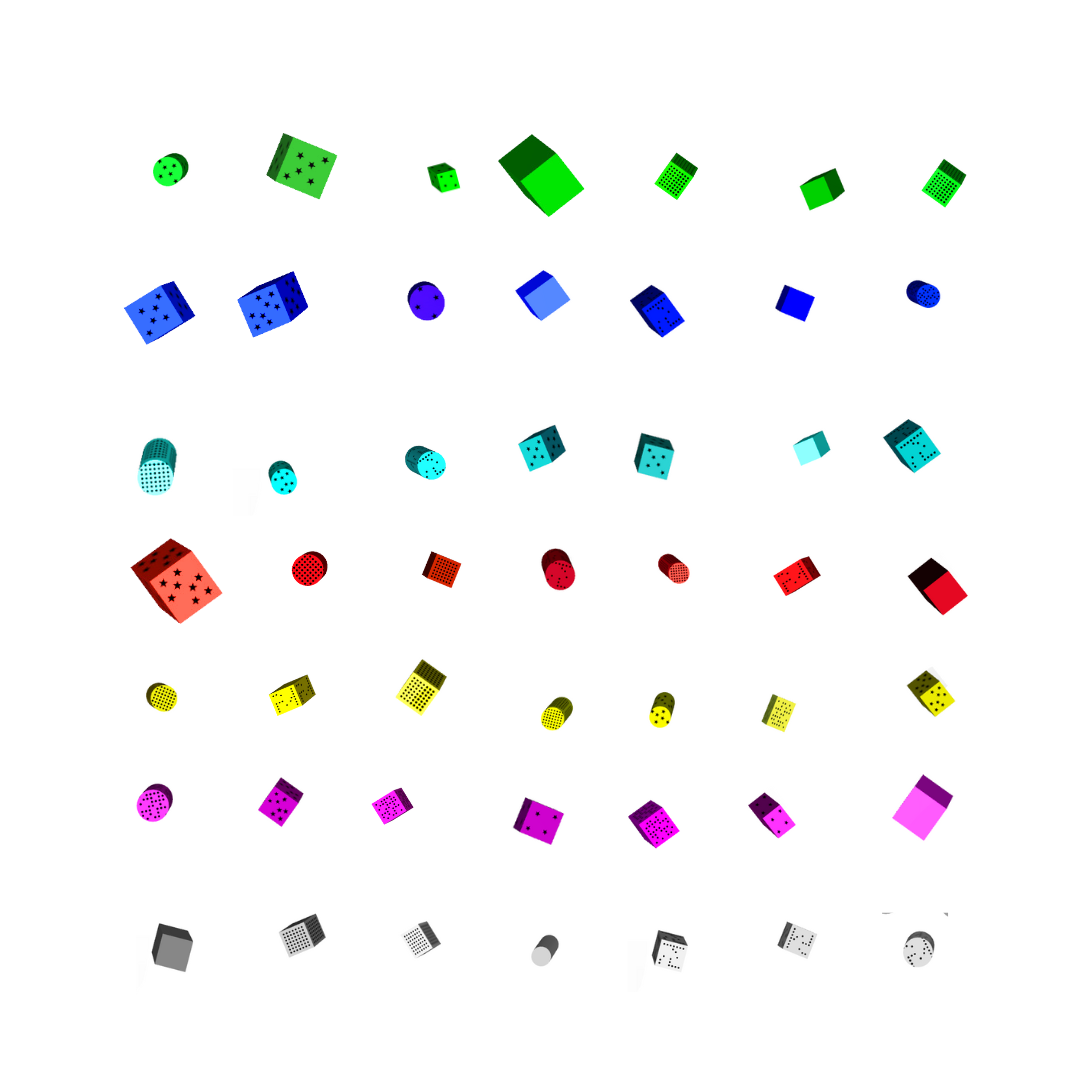}
  \caption{Object variability}
  \label{fig:variability}
\end{subfigure}%
\caption{Overview of MuJoCo~\citep{DBLP:conf/iros/TodorovET12} environment used for simulation experiments (\secref{sec:simulation_experiments}) including different views (see \threefigref{fig:bird_eye}{fig:side_view}{fig:top_down}) and overview of object variability (see \figref{fig:variability}). Each environment contains 6-7 rigid body objects. Each task instruction uses concepts of texture (plain, dotted, start), shape (cube, rectangle, cylinder), and colours (red, green, blue, cyan, grey, magenta, yellow) that agent is not aware of before the deployment as well predefined concepts of spatial relationships (in front of, to the right of, to the left of, behind) and quantifiers (a/an, every/all, and the \(n\), where \(n\) is natural number).}
\label{fig:simulation_experiments}
\end{figure*}

\subsection{Simulation Experiments}
\label{sec:simulation_experiments}

We conduct blocksworld manipulation experiments in simulation (see \figref{fig:simulation_experiments}). Each task instruction has the form ``move \(r_1\) \(rel\) \(r_2\)'' where \(rel\) is a spatial relationship and \(r_{1,2}\) are referential expressions consisting of a quantifier followed by one to three symbols (e.g. ``every red cube''). CodeLlama7B \citep{DBLP:journals/corr/abs-2308-12950} is used for semantic parsing to obtain logical forms of referential expressions and DINOv2~\citep{DBLP:journals/corr/abs-2304-07193} for feature extraction to obtain object-centric embeddings. The environment includes an oracle: it answers the agent's questions and corrects its suboptimal actions (see \secref{sec:correction_in_execution}). Each agent attempts to solve the task up to 5 times (one episode), with their policy evaluated over the same sequence of 60 episodes, averaged over 5 runs.

\begin{figure*}[ht]
\begin{subfigure}{\textwidth}
  \centering
  \includegraphics[width=\linewidth]{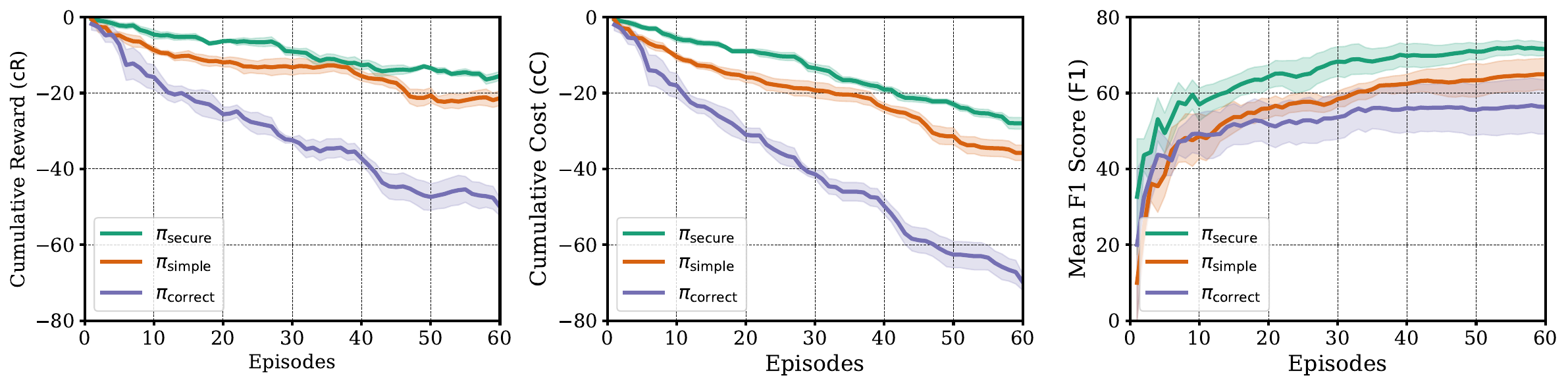}
  \caption{ITL experiment}
  \label{fig:sim_base_policies}
\end{subfigure}%
\hfill
\begin{subfigure}{\textwidth}
  \centering
  \includegraphics[width=\linewidth]{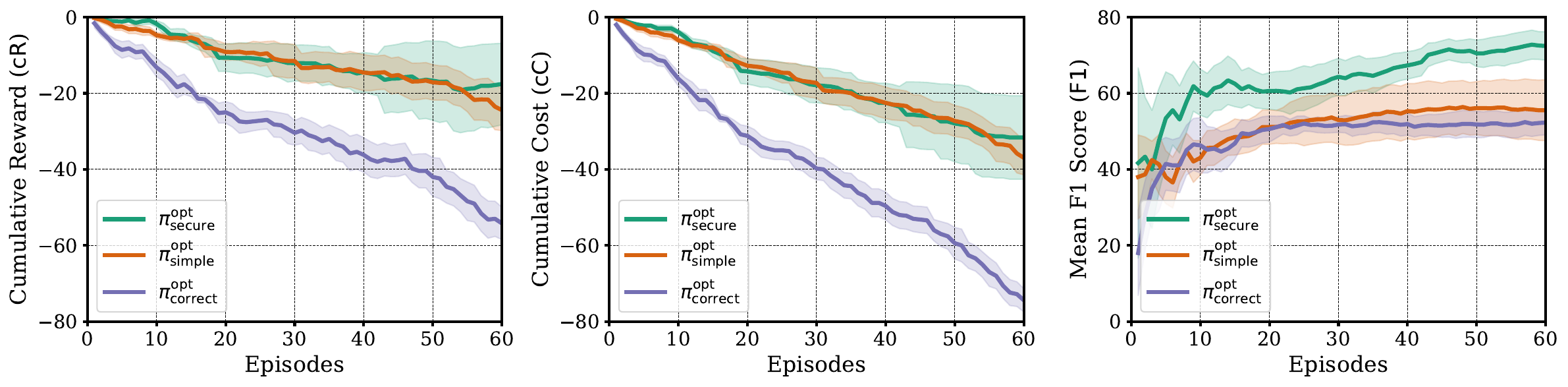}
  \caption{Optimistic belief experiment}
  \label{fig:sim_opt_policies}
\end{subfigure}
\hfill
\begin{subfigure}{\textwidth}
  \centering
  \includegraphics[width=\linewidth]{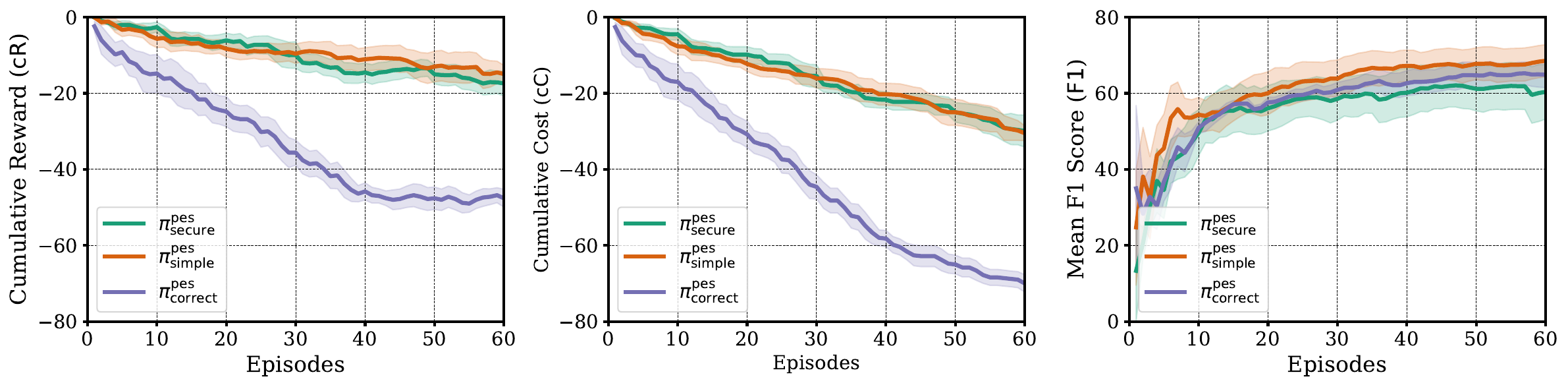}
  \caption{Pessimistic belief experiment}
  \label{fig:sim_pes_policies}
\end{subfigure}
\caption{Simulation experiment results. Learning curves are given with 95\% confidence interval.}
\label{fig:simu_results}
\end{figure*}

\subsubsection{ITL Experiment}
\label{sec:itl_experiments}

The results of the simulation experiment are given in \figref{fig:sim_base_policies}. The cumulative reward \(\mathrm{cR}\) and commutative cost \(\mathrm{cC}\) are taking negative values, indicating both the agent's questions and the user's corrections before succeeding (or not) in executing the task. Note that the \(\mathrm{cR}\) curve is steeper than \(\mathrm{cC}\) due to the positive rewards received from solving the task. Agents that engage in conversation before execution (i.e., $\pi_{\mathrm{secure}}$ and $\pi_{\mathrm{simple}}$) receive a higher and statistically different \(\mathrm{cR}\) and \(\mathrm{cC}\) compared with \(\pi_{\mathrm{correct}}\) (paired t-test p-values: \(\pi_{\mathrm{secure}}\text{ v.s. }\pi_{\mathrm{correct}}: 4\cdot10^{-26}\) and \(\pi_{\mathrm{simple}}\text{ v.s. }\pi_{\mathrm{correct}}: 4\cdot10^{-8}\)). This signifies the value of the learner having some control over the information they receive from the conversation before attempting to solve the task — a form of active learning.  
Comparing \(\mathrm{cR}\) for \(\pi_{\mathrm{secure}}\) and \(\pi_{\mathrm{simple}}\), we observe that for the majority of the tasks \(\pi_{\mathrm{secure}}\) performs better than \(\pi_{\mathrm{simple}}\) except for a brief task period 33-37 in which there's no statistically significant difference, indicating that both policies have similar embodied conversations and interactions in the environment. However, the \(\mathrm{mF1}\) of \(\pi_{\mathrm{secure}}\) performs on average 5\% better than \(\pi_{\mathrm{simple}}\) and 10\% better than \(\pi_{\mathrm{correct}}\). This signifies that it is beneficial to use agents whose decision-making and processing utilise the logical consequences of the user's embodied conversation messages.

\subsubsection{Optimistic and Pessimistic Belief Experiments} 
\label{sec:bias_belief}

We experiment with SECURE's ability to cope with (initial) biased beliefs that might be false. To evaluate this, we consider a variation on the experiments from \secref{sec:itl_experiments}, in which all textures (plain, dotted, starry) are assigned optimistic \(w_p^{\mathrm{opt}} = 0.7\) and pessimistic \(w_p^{\mathrm{pes}} = 0.3\) prior weights: these respectively impose an initial bias that all objects are (and respectively are not) plain, dotted and starry. Such bias encourages (respectively discourages) symbol prediction. The results are in \twofigref{fig:sim_opt_policies}{fig:sim_pes_policies}. When comparing \(\mathrm{cR}\) and \(\mathrm{cC}\) for agents with biased beliefs, \(\pi_{\mathrm{correct}}\) performs worse than agents that engage in embodied conversation. For \(\pi_{\mathrm{simple}}\) and \(\pi_{\mathrm{secure}}\), for the period of the first 20 tasks, \(\pi_{\mathrm{secure}}\) performs better under biased beliefs while later the performance difference is not conclusive, fluctuating between \(\pi_{\mathrm{simple}}\) or \(\pi_{\mathrm{secure}}\). The \(\mathrm{mF1}\) scores show that \(w_p^{\mathrm{opt}}\) hurts \(\pi_{\mathrm{simple}}\), making its grounding performance similar to \(\pi_{\mathrm{correct}}\), while \(\pi_{\mathrm{secure}}\)'s grounding performance is similar to that in \secref{sec:itl_experiments}. On the other hand, \(w_p^{\mathrm{pes}}\) does not yield any statistically significant difference among the agents. This is not surprising: the value derived from the semantic analysis diminishes because it predicts that atoms are false, which aligns with the pessimistic bias. We observe that \(\pi^{\mathrm{opt}}_{\mathrm{secure}}\) has high variance due to diverse dialogues, involving both attempts to solve the task without dialogue and questions that exploit semantic analysis, leading to more diverse belief states.

\subsection{Real-world Experiment}
\label{sec:real_world_experiment}

\begin{figure*}[ht]
\begin{subfigure}{0.25\textwidth}
  \centering
  \includegraphics[width=0.95\linewidth]{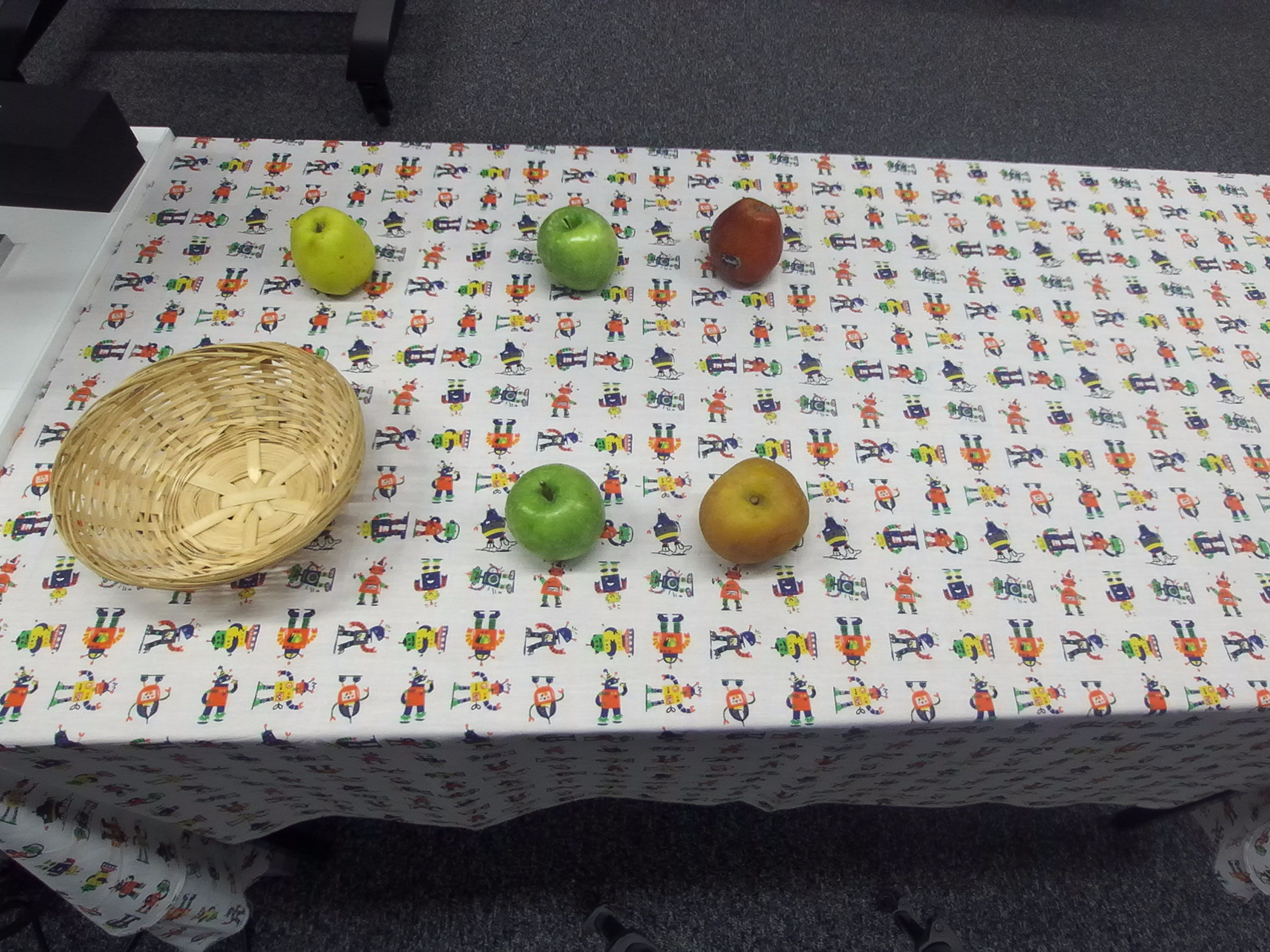}
  \caption{Environment 1}
  \label{fig:env1}
\end{subfigure}%
\begin{subfigure}{0.25\textwidth}
  \centering
  \includegraphics[width=0.95\linewidth]{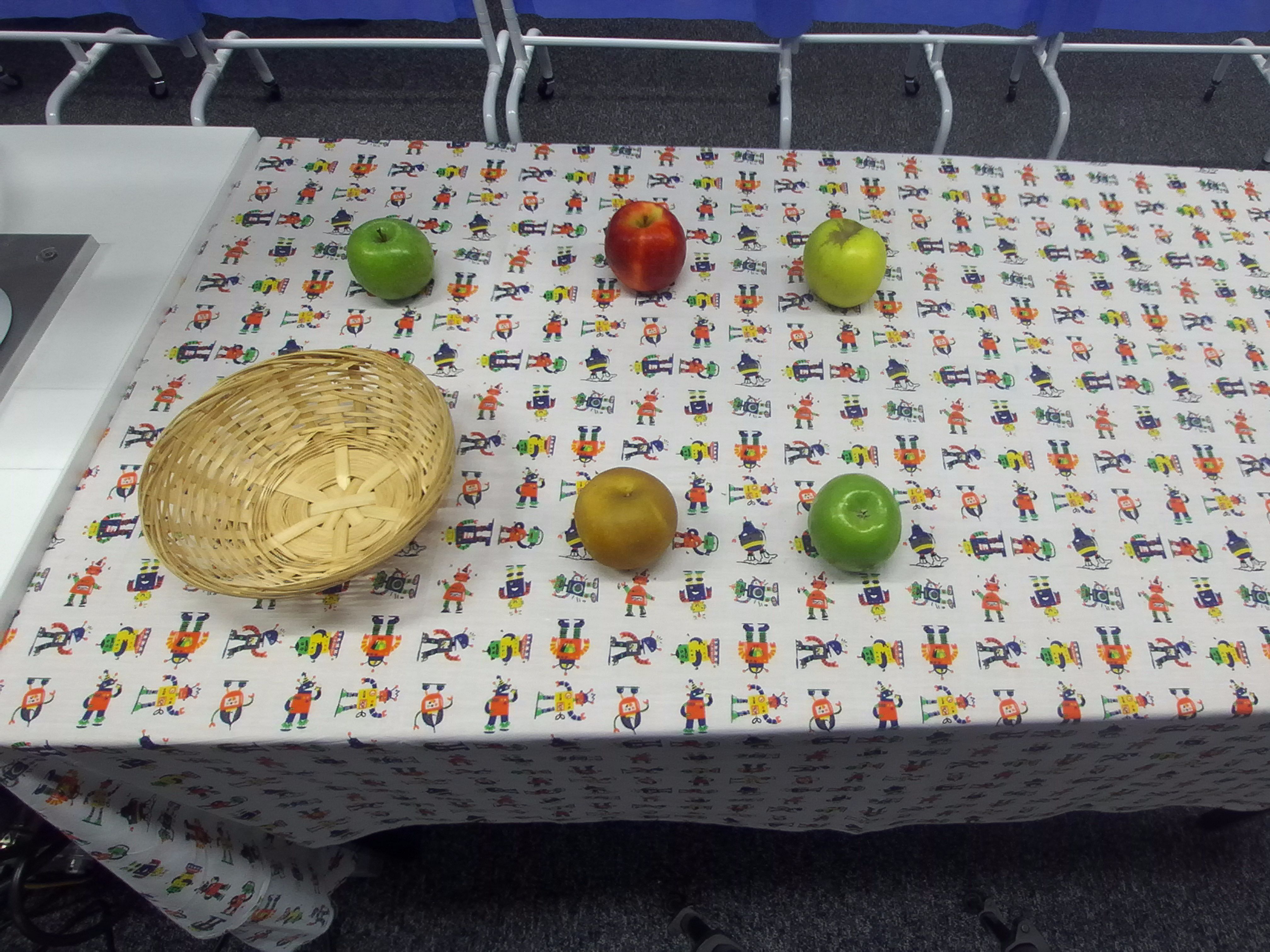}
  \caption{Environment 2}
  \label{fig:env2}
\end{subfigure}%
\begin{subfigure}{0.25\textwidth}
  \centering
  \includegraphics[width=0.95\linewidth]{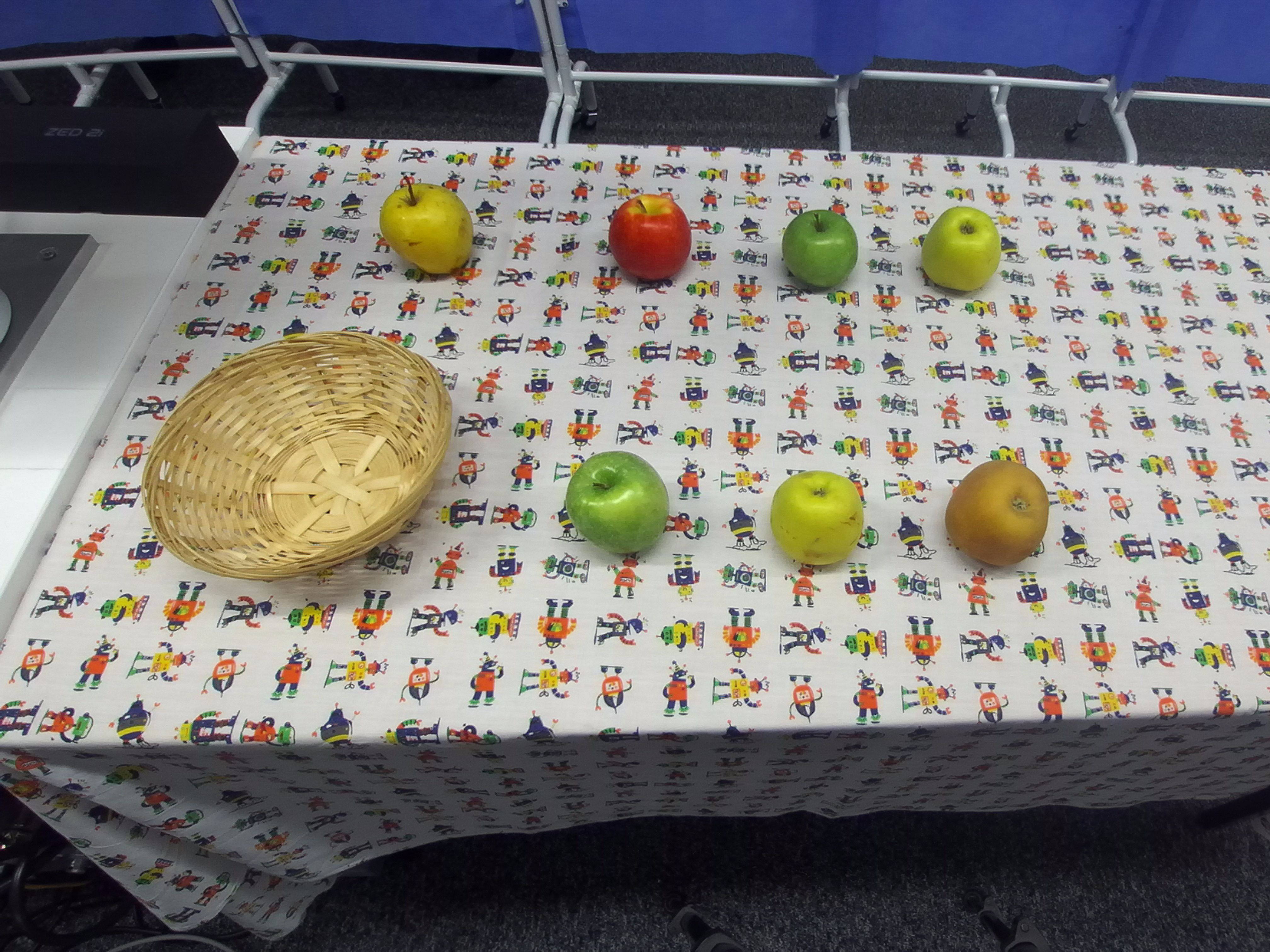}
  \caption{Environment 3}
  \label{fig:env3}
\end{subfigure}%
\begin{subfigure}{0.25\textwidth}
  \centering
  \includegraphics[width=0.95\linewidth]{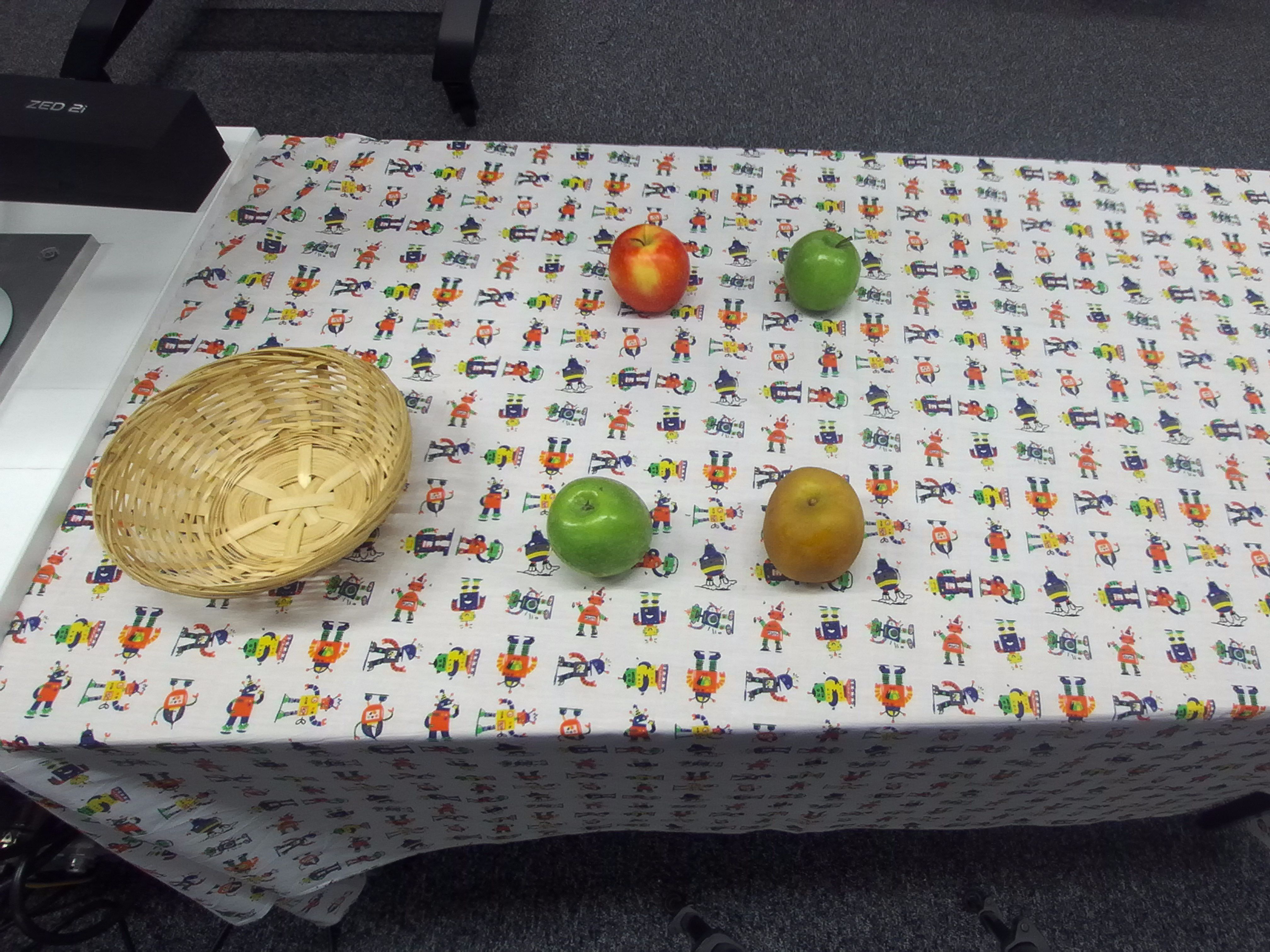}
  \caption{Environment 4}
  \label{fig:pred_env3}
\end{subfigure}%
\caption{Real-world experiment and task setup overview. Each of the 4 environments consists of different species of apples (granny smith, golden delicious, red delicious, pink lady, and russet) and a basket. The robot is instructed to ``put the two granny smith apples inside a basket''. All environments contain exactly two granny smiths while the other species vary, together with presenting different instances of the same species. The agent uses grounding DINO~\citep{DBLP:journals/corr/abs-2303-05499} to recognise and locate apples and the basket, but cannot accurately distinguish Granny Smiths from other types of apples. Note that using just ``granny smith'' grounding DINO fails to detect any object in all environments, suggesting limited concept awareness. To evaluate the proposed framework, we conduct experiments with all possible sequences of environments (24 in total) in which the agent starts unaware of concept \(\mathtt{grannysmith}\) and through interaction with the user (both questions and corrections) their belief state is updated. For evaluation we record final \(\mathrm{cR}\) and \(\mathrm{mF1}\) given in table~\ref{tab:real_world_results}.}
\label{fig:real_world_experiment}
\end{figure*}

\begin{wraptable}{r}{0.6\textwidth}
        \centering
\begin{tabular}{@{}rlll@{}}
\toprule
 Agent   & \(\mathrm{cR}\)   & \(\mathrm{cC}\) & \(\mathrm{mF1}\%\)   \\ \midrule
\(\pi_{\mathrm{correct}}\)   & -3.70 \(\pm\) 0.53 & -7.70 \(\pm\) 0.53 & 50.54  \(\pm\) 3.72 \\
\(\pi_{\mathrm{simple}}\)    & -3.45 \(\pm\) 0.72 & -7.45 \(\pm\) 0.72 & 45.11 \(\pm\) 3.08 \\
\(\pi_{\mathrm{secure}}\)   & \text{ }\textbf{0.47 \(\pm\) 0.44}\(^\dagger\) & \textbf{-3.625 \(\pm\) 0.44}\(^\dagger\)& \textbf{51.63 \(\pm\) 3.99}  \\ \bottomrule
\end{tabular}
\caption{Real-world experiment results with 95\% confidence interval. \(^\dagger\) indicates a statistically significant difference between agents.}
\label{tab:real_world_results}
\end{wraptable}

We conduct a real-world experiment on apple manipulation (see \figref{fig:real_world_experiment} for overview and \appref{app:real_world_experiment_details} for details). To detect apples, prior weights are assigned using grounding DINO~\citep{DBLP:journals/corr/abs-2303-05499}: when given the prompt ``granny smith apple'', this localises all the apples and outputs the similarity score between the prompt and localised environment region. This score is used as a prior weight for \(\mathtt{grannysmith}\). Similarly to the simulation experiments, the robot can attempt to solve the task 5 times before the end of the episode; the user monitors the execution and corrects if a mistake is made. Table~\ref{tab:real_world_results} shows the summary results. The policy \(\pi_{\mathrm{secure}}\) has on average highest \(\mathrm{cR}\), \(\mathrm{cC}\) and \(\mathrm{mF1}\). We observe these behaviours:

\begin{itemize}
    \item \(\pi_{\mathrm{secure}}\): is highly uncertain about the denotations of \(\mathtt{grannysmith}\) and keeps asking for evidence using the question ``show me the two granny smiths'', after which no reward is obtained for the task (-1 for the question and +1 for task success). After several such questions, the agent can correctly identify denotations and is confident enough to decide to attempt to solve the task;
    \item \(\pi_{\mathrm{simple}}\): is highly uncertain about the domain and often first opts to ask a question ``show me a granny smith'', which always results in several corrections on falsely picked fruits afterwards. Importantly, from the evidence gathered, \(\pi_{\mathrm{simple}}\) is unable to significantly change its grounding based on the evidence gathered from its previous tasks and always requires some kind of interaction with the user to solve the task. 
    \item \(\pi_{\mathrm{correct}}\): is attempting to solve the task from corrections, leading to the lowest \(\mathrm{cR}\), on average requiring 3 attempts to solve the task. At the same time, this gathers a lot of negative evidence, so after attempting to solve several tasks, this policy succeeds later in identifying referents as indicated on average higher \(\mathrm{mF1}\) than for \(\pi_{\mathrm{simple}}\).
\end{itemize}
For additional intuitions about agent behaviours, see appendix~\ref{app:real_world_experiment_details} for interaction traces for different agents in table~\ref{tab:traces} and videos in supplementary material illustrating belief updates using different interaction strategies.

\section{Related Work}
\label{sec:related_work}

\paragraph{Lifelong learning} SECURE as a framework for lifelong robots is designed for ITL scenarios, where there is natural interaction (embodied conversation in our case) between the agent and the user \citep{DBLP:journals/expert/LairdGAFJLSSTTW17}. By design, it is not at odds with generative multimodal models~\citep{DBLP:journals/corr/abs-2301-12597, driess2023palme}, but rather provides the means of structure-level domain adaptation when embodied agents are deployed in settings in which new concepts are frequently introduced or changed, requiring learning to adapt rapidly by using all possible incidental supervision signals~\citep{DBLP:conf/aaai/Roth17}. SECURE handles the issue of unawareness, specifically missing concepts from the hypothesis space of possible domain models, which must be discovered and learned about to solve the task. A related concept is catastrophic forgetting~\citep{MCCLOSKEY1989109,Ratcliff1990-RATCMO}, which describes the loss of previously acquired knowledge due to limitations on the agent's memory (i.e., the amount of experience stored). Both of these phenomena hinder lifelong learning but in different ways---e.g., unawareness stems from a lack of knowledge, while forgetting results from the loss of knowledge. SECURE focused on the underexplored problem of unawareness, while catastrophic forgetting was not addressed (our agent had no limit on the number of exemplars it could store).    

\paragraph{Language-conditioned Manipulation} Language can be used to specify manipulation goals~\citep{NEURIPS2020_9909794d,shridhar2021cliport,DBLP:conf/corl/IchterBCFHHHIIJ22} with grounded representations \citep{jiang2023vima,DBLP:conf/rss/GkanatsiosJXZAF23}, and can be highly modular and generalizable. Our framework enhances such work by providing a methodology to address unawareness. SECURE is most similar to \citet{DBLP:journals/corr/abs-2307-01928}, which explores how planners can be interactively repaired when predictions are confident but wrong. We go beyond this by: a) considering a complete embodied conversation that includes corrections and b) explicitly learning a dialogue policy rather than performing statistical tests~\citep{DBLP:journals/corr/abs-2107-07511}. 

\paragraph{Learning from Human Feedback} Human-centric agents have to align their behaviour with human preferences \citep{DBLP:conf/nips/ChristianoLBMLA17, DBLP:journals/corr/abs-1909-08593, DBLP:journals/corr/abs-2212-08073} by integrating learning and acting during the fine-tuning phase or continuously through online, human-in-the-loop interactions \citep{DBLP:journals/air/MosqueiraReyHABF23}. Traditional models primarily rely on sparse binary rewards, but SECURE leverages symbolic inference to utilise more nuanced embodied conversation messages, enabling it to learn and efficiently perform new tasks involving unforeseen concepts.

\paragraph{Neuro-symbolic Reasoning} There is a broad interest in combining neural networks with symbolic/classical algorithms~\citep{garnelo2019reconciling,10.3233/AIC-210084} with many ways to do so~\citep{Sheth2023NeurosymbolicA}. In this project, we integrated symbolic reasoning into neural network inference by leveraging semantic analysis of embodied conversation, which builds upon \citet{rubavicius-lascarides-2022-interactive}. In the experiments, we study agents that use different semantic analyses to quantify and qualify the benefits of semantic awareness. Large language models have shown the capacity to handle at least some aspects of semantic analysis~\citep{madusanka-etal-2023-quantifiers}, but quantifying that capacity is beyond the scope and focus of this work.

\section{Conclusion}
\label{sec:conclusion}
We have presented a framework for processing embodied conversation to solve rearrangement under unawareness, which broadly falls into the type of task expected to be handled in ITL. We have demonstrated the benefits of making the agent semantics-aware in interactive symbol grounding and task learning, as well as having a mechanism to address false beliefs. There are several directions of future work to consider, including handling and exploiting other semantics of conversation (e.g. generics or contrast) as well as using additional learning signals and knowledge sources.

\section{Limitations}
\label{sec:limitations}

This project assumes collaborative human-robot interaction in the context of solving rearrangements that require pick-and-place manipulation of rigid objects. Human-robot interaction may not always be collaborative, and in real-world scenarios, it often involves deception~\citep{6722151}, which conflicts with the reliance on the user's feedback. The semantic analysis of corrections is explicitly encoded for the set of execution actions (\secref{sec:correction_in_execution}), so in the case of additional execution actions (e.g., navigation action in the room), additional knowledge needs to be encoded or inferred following semantic analysis of corrections~\citep{lascarides:asher:2009}. Due to the object-centric view of the environment, SECURE inference complexity (in particular \(\mathrm{WMC}\)) increases with the number of objects of interest in the environment and the length of the embodied conversation. This, nevertheless, was not an issue in practice due to the limited number of objects and short dialogues. Nonetheless, scaling this to real-world scenarios would require approximate inference~\citep{schlandals_anytime_approximation}. Our experiments do not consider dynamic situations, and dialogues were limited to non-fluent properties of objects. In effect, the environment is treated as static, even though its fluent properties (e.g., the location of objects) can change when the agent interacts with the environment. Finally, we are only considering user feedback as a source for learning. Other learning signals, such as utilising data available on the Internet, could also potentially be used to bootstrap the interactive task learning process.

\section*{Acknowledgments}
\label{sec:acknowledgments}

This work was supported in part by the UKRI Centre for Doctoral Training in Natural Language Processing, and the UKRI (grant EP/S022481/1) and UKRI Turing AI World Leading
Researcher Fellowship on AI for Person-Centred and Teachable Autonomy (grant EP/Z534833/1).

\bibliography{references,anthology,anthology_p2}
\bibliographystyle{collas2025_conference}

\newpage
\appendix
\section{Logic of Natural Language Expressions}
\label{app:logic}

This appendix outlines the background on representing and reasoning with natural language expressions using logic.

\subsection{Logic of Sentences}
\label{sec:sentences}

Natural language sentences like ``every cube is a cuboid'' can be represented in first-order logic.  The syntax of the language of first-order logic is constructed recursively:  predicate symbols followed by the appropriate number of terms (i.e., a variable \(\mathtt{x}\in\sL_{\mathrm{var}}\) or a constant \(\mathtt{o}\in\sL_{\mathrm{const}}\)) are {\em well-formed formulae}, and these combine with boolean operators and connectives $\sL_{\mathrm{conn}}=\{\wedge, \vee, \neg, \rightarrow, \leftrightarrow\}$ and the quantifiers $\forall$ and $\exists$ in the usual way to create more complex well-formed formulae. First-order languages have been used extensively to capture natural language semantics~\citep{DBLP:books/daglib/0031956}: for example, ``every cube is a cuboid'' has the logical form \(\forall\text{ }\mathtt{x. cube(x)\rightarrow cuboid(x)}\).

SECURE utilises generalised quantifier theory to represent the meanings of natural language determiners such as ``the one'' and ``both'' ~\citep{Barwise1981GeneralizedQA}.\footnote{SECURE does not consider vague quantifiers~\citep{bradburn1979vague} like ``some'' or ``few'', which implicate soft constraints on objects in the domain model, as a consequence of pragmatic principles of cooperative conversation~\citep{grice:1975}. The interpretation of these quantifiers in cooperative conversation is outside SECURE capabilities but, in principle, could be captured by the agent following the same architecture.}  generalised quantifier theory extends the classical first-order language with quantifiers additional to $\forall$ and $\exists$, and it is designed to express the meanings of other quantitative sentences about the domain with the usage of quantifiers like e.g., {\em at least two}, {\em all but one}, {\em the one}, {\em both}, and so on.

Syntactically, a well-formed formula involving a generalised quantifier ${\tt Q}\in\sL_{quant}$ is of the form $\mathtt{Q\text{ }x.(\phi,\psi)}$, where ${\tt Q}$ binds the variable ${\tt x}\in\sL_{\mathrm{var}}$, and the restrictor $\phi$ and body $\psi$ are both well-formed formulae in which $\mathtt{x}$ is absent or free. Very roughly, the restrictor $\phi$ expresses the descriptive content of the noun phrase that introduced the determiner whose meaning is $\mathtt{Q}$, and the body $\psi$ expresses the content of `the rest' of the sentence (typically the verb phrase). For example, the natural language sentence ``the one red cube is $o_1$'', or more explicitly ``the one red cube is object $o_1$'', has the following logical form \(\mathtt{\_the\_one\_q\text{ }x.(red(x)\wedge cube(x), object(o_1))}\). The list of possible generalised quantifiers is given in Table~\ref{tab:quantifiers} column 1.

Formally, the set of syntactically well-formed formulae---the language of sentences \(\sL_{snt}\)---is defined recursively:
\begin{enumerate}
    \item If $\mathtt{p}$ is an $n$-place predicate symbol and $\mathtt{t_1,t_2,...,t_n}$ are terms, then $\mathtt{p(t_1,t_2,...,t_n)}$ is well-formed formula;
    \item If $\phi$ is well-formed formula, then $\neg\phi$ is a well-formed formula;
    \item If $\phi$ and $\psi$ are well-formed formulas, then $\phi\wedge\psi$, $\phi\vee\psi$, $\phi\rightarrow \psi$, $\phi\leftrightarrow \psi$ are all well-formed formulas;
    \item If $\phi$ and $\psi$ are well-formed formulas, known as {\em restrictor} and {\em body}, and $\phi$ and $\psi$ either do not feature the variable ${\tt x}$ or ${\tt x}$ is free in these formulae, and if $\mathtt{Q}$ is a quantifier, then $\mathtt{Q\text{ }x.}(\phi,\psi)$, is well-formed formula. For this kind of well-formed formula, the variable $\mathtt{x}$ is a \textit{bound variable} that is bound by ${\tt Q}$. 
\end{enumerate}

For convenience, we also assume that each object \(o\in \sO\) in the domain has a unique constant~\(\mathtt{o}\) that denotes it. 

Elements of \(\sL_\mathrm{snt}\) that are well-formed expressions in generalised quantifier theory are assigned a truth-conditional semantics with respect to a first-order domain model \(\gM\) consisting of a set of objects \(\sO\), vocabulary \(\sV\), and the interpretation function \(I\colon\sV\mapsto\sO^*\) as well as a variable assignment function \(g\), which maps variables to individuals in the model~\citep{Tarski1931SurLE}. The truth-conditional semantics define a valuation function \(\llbracket\cdot\rrbracket^{\gM,g}\colon\sL_\mathrm{snt}\mapsto\{0,1\}\), which in turn specifies when a domain model variable assignment pair $(\gM,g)$ satisfies a formula $\phi\in\sL_\mathrm{snt}$, written $\gM,g\models \phi$: 
\begin{equation*}
    \gM,g \models \phi \text{ if and only if } \llbracket \phi \rrbracket^{\gM,g} = 1 
\end{equation*}

If there are no free variables in \(\phi\in\sL_\mathrm{snt}\) then \(\llbracket \phi \rrbracket^{\gM,g} = \llbracket \phi \rrbracket^{\gM,g'}\) for all \(g,g'\) and  without a loss of generality such situation can be expressed as:
\begin{equation*}
        \gM   \models \phi \text{ if and only if } \forall g \text{ }\llbracket \phi \rrbracket^{\gM,g} = 1 
\end{equation*}

\begin{table}[]
\centering
\begin{tabular}{@{}llll@{}}
\toprule
Quantifier $\mathtt{Q}$ & Surface form       & Truth condition $\mathcal{Q}$                                        & Referent constructor $\langle \mathtt{Q}\rangle^{\gM}$                                        \\ \midrule
$\mathtt{\_exactly\_n\_q}$           & exactly \(n\)      & $ |\sR\cap \sB| = n$                             & $\{ \sA \subseteq \sO \mid |\sA| = n\}$                                          \\
$\mathtt{\_at\_most\_n\_q}$          & at most \(n\)      & $ |\sR\cap \sB| \leq n$                          & $\{ \sA \subseteq \sO \mid |\sA| \leq n\}$                                       \\
$\mathtt{\_at\_least\_n\_q}$         & at least \(n\)     & $ |\sR\cap \sB| \geq n$                          & $\{ \sA \subseteq \sO \mid |\sA| \geq n\}$                                       \\
$\mathtt{\_a\_q}$                    & a/an               & $ |\sR\cap \sB| \neq n$                          & $\{ \sA \subseteq \sO \mid |\sA| \leq 1\}$                                       \\
$\mathtt{\_every\_q}$                & all/every          & $ |\sR\cap \sB| = |\sR| \wedge |\sR| = 1$               & $\{ \sA \subseteq \sO \mid |\sA| = |\sO| \wedge |\sO| = 1\}$                              \\
$\mathtt{\_the\_n\_q}$               & the \(n\)          & $ |\sR\cap \sB| = n \wedge |\sR| = n $ & $\{ \sA \subseteq \sO \mid |\sA| = |\sO| \wedge |\sO| = n\}$       \\
$\mathtt{\_both\_q}$                 & both               & $ |\sR\cap \sB| = 2 \wedge |\sR| = 2 $ & $\{ \sA \subseteq \sO \mid |\sA| = |\sO| \wedge |\sO| = 2 \}$      \\
$\mathtt{\_all\_but\_n\_q}$          & all but \(n\)      & $ |\sR\cap \sB| = |\sR| - n \wedge |\sR| > n $           & $\{ \sA \subseteq \sO \mid |\sA| = |\sO|-n \wedge |\sO| > n \}$ \\
$\mathtt{\_n\_of\_the\_m\_q}$        & \(n\) of the \(m\) & $ |\sR\cap \sB| = n \wedge |\sR| = m $ & $\{ \sA \subseteq \sO \mid |\sA| = n \wedge |\sO| = m\}$                  \\ \bottomrule
\end{tabular}
\caption{generalised quantifiers (column 1), surface form (column 2), truth-conditions $\mathcal{Q}$ between the restrictor set  \(\sR\) and body set \(\sB\), used to evaluate well-formed formulas of the form \(\mathtt{Q\text{ }x.}(\phi,\psi)\) (column 3), and  referent constructor used to evaluate the logical form of referential expressions for the domain model \(\gM\) (column 4).}
\label{tab:quantifiers}
\end{table}

To define valuation function \(\llbracket\cdot\rrbracket^{\gM,g}\), \(\lambda\)-expressions of the form \(\lambda\text{ }\mathtt{x}.\phi\) are used in which \(\mathtt{x}\) is a variable that is free or absent from \(\phi\). Note that such \(\lambda\)-expressions are well-formed expressions but are not in \(\sL_\mathrm{snt}\).  Here, they are simply used for defining the truth conditions of formulae that feature generalised quantifiers.

Well-formed formulas in truth-conditional semantics are either evaluated as being true (1) or false (0). This evaluation is given recursively:
\begin{align*}
\begin{split}
    \llbracket {\tt a} \rrbracket^{\gM,g} &= \begin{cases} I({\tt a})\quad \text{if \(\tt a\) is a predicate symbol}\\ g({\tt a})\quad \text{if \({\tt  a}\) is a variable} \\ I({\tt a})\quad \text{if \(\tt a\) is a constant\footnotemark} \end{cases} \\
    \llbracket \mathtt{p(t_1,t_2,\dots,t_n)} \rrbracket^{\gM,g} &= 1 \text{ if }(\llbracket \mathtt{t}_1 \rrbracket^{\gM,g},\llbracket \mathtt{t}_2 \rrbracket^{\gM,g},\dots,\llbracket \mathtt{t}_n \rrbracket^{\gM,g}) \in \llbracket \mathtt{p} \rrbracket^{\gM,g} \\
    \llbracket \neg \phi \rrbracket^{\gM,g} &= 1 \text{ if }\llbracket \phi \rrbracket^{\gM,g} = 0 \\
    \llbracket \phi \wedge \psi \rrbracket^{\gM,g} &= 1 \text{ if }\llbracket \phi \rrbracket^{\gM,g} = 1 \text{ and }\llbracket \psi \rrbracket^{\gM,g} = 1 \\
    \llbracket \phi \vee \psi \rrbracket^{\gM,g} &= 1 \text{ if }\llbracket \phi \rrbracket^{\gM,g} = 1 \text{ or }\llbracket \psi \rrbracket^{\gM,g} = 1 \\
    \llbracket \phi \rightarrow \psi \rrbracket^{\gM,g} &= 1 \text{ if }\llbracket \phi \rrbracket^{\gM,g} = 0 \text{ or }\llbracket \psi \rrbracket^{\gM,g} = 1 \\
    \llbracket \phi \leftrightarrow \psi \rrbracket^{\gM,g} &= 1 \text{ if }\llbracket \phi \rightarrow \psi \rrbracket^{\gM,g} = 1 \text{ and }\llbracket \psi \rightarrow \phi \rrbracket^{\gM,g} = 1  \\
    \llbracket \lambda\text{ }\mathtt{x}.\phi \rrbracket^{\gM,g} &= \{o\in \sO \mid \llbracket \phi \rrbracket^{\gM,g[\mathtt{x}/\mathtt{o}]} = 1 \} \\
    \llbracket \mathtt{Q\text{ }x.(\phi,\psi)} \rrbracket^{\gM,g} &= \mathcal{Q} (\llbracket \lambda\text{ }\mathtt{x}.\phi \rrbracket^{\gM,g}, \llbracket \lambda\text{ }\mathtt{x}.\psi \rrbracket^{\gM,g}) 
\end{split}
\end{align*}
\footnotetext{As mentioned earlier, we assume a unique constant denotes each object, and so without loss of generality, we can assume $I(\mathtt{a})=a$, where $a\in \sO \wedge \sO\in\gM$ is the unique object denoted by $\mathtt{a}$.}

where \(g[\mathtt{x}/\mathtt{o}]\) is a variable assignment function just like \(g\) except that \(g[\mathtt{x}/\mathtt{o}](\mathtt{x}) = \mathtt{o}\) and \(\mathcal{Q}\) is a truth-condition (Table~\ref{tab:quantifiers} column 3) used to evaluate well-formed formulas that feature generalised quantifiers. The value of these truth conditions depends on two sets defined by the \(\lambda\)-expression, commonly referred to as the restrictor set \(\sR\) and the body set \(\sB\). 

\subsection{Logic of Referential Expressions}
\label{sec:refexps}

Natural language expressions that denote objects in the environment are known as referential expressions \(r\)~\citep{Kripke1980NamingAN}. Linguistically, they are expressed by noun phrases: a noun phrase can consist of a determiner and a noun with pre- and post-modifiers (e.g., ``the one red cube that is to the left of every green sphere''); a pronoun or demonstrative (e.g., ``it'', ``this'', ``that''), or proper name (e.g., ``Edinburgh'', ``Kim'', `` Mr. Smith''). SECURE focuses on referential expressions that are of the first kind: i.e.; they include a determiner and a noun with pre-and/or post-modifiers. We denote the logical form of referential expression \(r\) as \(\Phi(r)\) and the language of these well-formed expressions as \(\sL_\mathrm{ref}\). 

Since \(r\) is a noun phrase and not a full sentence like the ones discussed in \secref{sec:sentences}, \(\Phi(r)\) consists of a generalised quantifier \(\mathtt{Q}\) (corresponding to \(r\)’s determiner), its restrictor \(\phi\) (corresponding to \(r\)’s adjectives and nouns) but the body of the quantifier is ‘missing’. In semantic analysis, this is traditionally captured using \(\lambda\)-calculus with logical forms like \(\mathtt{\lambda\text{ }\psi.\mathtt{Q}\text{ }x.(\phi,\psi)}\) but this is not what we want to capture to process embodied conversation.  The \(\lambda\)-term denotes the set of properties satisfied by the objects that satisfy \(\phi\), with quantifier \(\mathtt{Q}\) imposing constraints on the relationship between the denotations of \(\phi\) and \(\psi\) (and sometimes also on $\phi$ itself as well). In our scenario, however, we need to identify the set of referent(s) that are denoted by the referring expression \(r\), not the properties satisfied by those referents. Because of these requirements, we express the logical form \(\Phi(r)\) of the referential expression \(r\) another way and interpret it using the theory of generalised quantifiers~\citep{Barwise1981GeneralizedQA}. Specifically, we represent the referential expression ``the one red cube'' with the following logical form \(\langle\mathtt{\_the\_one\_q\text{ }x.red(x)\wedge cube(x)}\rangle\). More generally, the logical forms of referential expressions are of the form \(\langle \mathtt{Q\text{ }x.\phi}\rangle\), where \(\phi\in\sL_\mathrm{snt}\) is a well-formed formula with only one free variable \(\mathtt{x}\). Intuitively, this formula captures the content of the described objects and $\mathtt{Q}$ captures the content of the determiner.  The notation \(\langle\cdot\rangle\) is a convenient way to differentiate these logical forms from well-formed formulae and in particular, the logical forms of sentences.

The logical form of a referential expression $\Phi(r)\in\sL_\mathrm{ref}$ is evaluated for the domain model \(\gM\) to yield the referent \(\gR\). An element of $\gR$ is a {\em set} of objects because \(r\) may denote more than one object (e.g., any denotation of ``two cubes'' consists of two objects). $\cal{R}$ itself is a set of (potentially) more than one element because $\cal{M}$ may have more than one set of objects that are denoted by the referring expression 
$r$: e.g., if there is more than one object in $\gM$ that is a cube, then ``a cube'' has more than one referent in $\gM$. For example, given the domain model $\gM$ with objects \(\sO=\{o_1,o_2\}\) and interpretation function \(I(\mathtt{cube}) = \sO\), the expression ``all cubes'' should denote $\{\{o_1,o_2\}\}$ (there is only one set denoted by ``every cube'', and that is the set of all objects in the domain that are cubes), while ``a cube'' denotes $\{ \{o_1\}, \{o_2\}\}$ (i.e., there are two sets of objects that are denoted by this referential expression). Thus more formally,  \(\gR\) is always a subset of the power set of \(\sO\): \(\gR\subseteq 2^{\sO}\). These examples reveal two factors that affect the referent of $\Phi(r)$. First, objects in a denotation of $\Phi(r)$ satisfy the restrictor of the generalised quantifier.\footnote{This work ignore group nouns like ``committee''} Secondly, the generalised quantifier imposes its conditions on $\gR$, in particular on the cardinality of each set in $\gR$ that is a denotation, and for some quantifiers there is also a constraint on the relationship between a denotation and all the objects $\sO$, which is equivalent to a constraint on the cardinality of $\gR$ itself.  For instance, each set in a referent for ``at least two cubes'' must have a cardinality of at least two; for ``exactly two cubes'' the cardinality must be equal to 2; and ``the two cubes'' and ``both'' impose the additional constraint that not only should each denotation have cardinality 2, but also this denotation is {\em unique}~\citep{Russell1917-RUSKBA-4} (i.e., $\gR$ itself has cardinality 1).  Further, the referent for ``every cube'' is unique because it is the (unique) maximal set of objects in $\gM$ that satisfy the restrictor (again, $\gR$ must have cardinality 1). Constraints imposed by different quantifiers are lead to logical consequences about the environment~\citep{BEAVER1997939}. Note that such constraints are not English-specific: Finnish ``molempi'' has the same constraint as English ``both''. With this in mind, we obtain a formal definition of the reference semantics for an arbitrary logical form of the referential expression $\Phi(r)=\langle\mathtt{Q\text{ }x.}\phi \rangle$ by first defining a projection of $\gM$ onto a smaller domain model $\sigma(\gM,\phi,\mathtt{x})$, which consists of {\em all and only} those objects $o\in \sO$ that satisfy $\phi[\tt{x/o}]$ (i.e., the formula $\phi$ with each occurrence of $\tt{x}$  substituted with the unique constant $\tt{o}$ that denotes $o\in \sO$):
\begin{align}
\label{eq:projection}
\begin{split}
\sigma(\gM,\phi,\mathtt{x}) & = \sigma((\sO,\sV, I),\phi,\mathtt{x}) = (\sO',\sV, I')\\
\sO' &=\{o\in \sO \mid \gM\models \phi[\tt{x/o}]\} \\
I'&=I \downarrow \sO' \mbox{ (i.e., $I$ projected onto $\sO'$)}
\end{split}
\end{align}

We then define the reference semantics of $\Phi(r)$ for this projected domain model. As discussed before, the truth-conditional semantics of generalised quantifiers are defined using the restrictor $\sR$ and body $\sB$. But in our scenario, consisting only of referring expressions, the body $\sB$ is not present for truth-conditional evaluation.  However, by evaluating the content of the quantifier for the smaller projected model and making $\sB=\sO'$ (i.e., the objects in that projected model), the reference semantics can focus solely on how they constrain the {\em cardinalities} of these sets to achieve the above-desired effects for constructing the referent \(\gR\) Table~\ref{tab:quantifiers} column 4 shows how the referent is defined using set constructors. These constraints and the definition of model projection in Eq.~\ref{eq:projection} yield the reference semantics (Eq.~\ref{eq:reference_semantics}) for the logical form $\Phi(r)$ with the desired properties:
\begin{equation*}
    \label{eq:reference_semantics}
    \gR = \Phi(r)^{\gM} = \langle \mathtt{Q}\text{ }{\mathtt{x}}.\phi\rangle^{\gM} = \langle\mathtt{Q}\rangle^{\sigma(\gM,\phi,\mathtt{x})}
\end{equation*}
where \(\langle\mathtt{Q}\rangle^{\gM}\) is a referent constructor, utilising the condition specific to the (generalised) quantifier (and defined in Table~\ref{tab:quantifiers} column 4).

\newpage
\section{Algorithmic Details}
\label{app:algorithms}

\Algref{alg:update} outlines the belief state update procedure using a well-formed formula, constructed from the semantic analysis of embodied conversation and algorithm~\ref{alg:sarsa} outlines dialogue strategy optimization procedure using a semi-gradient SARSA with hyperparameter values given in table~\ref{tab:sarsa_hyperparameters}.

\begin{algorithm}
\caption{\(\mathrm{Update}\) (Belief state update with well-formed formula)}
\label{alg:update}
\begin{algorithmic}[1]
\Require Belief state~\(b\in\sB\) consisting of a set of objects~\(\sO\) set of object-centric embeddings~\(\sX\) (one-to-one corresponce with objects), vocabulary of predicates~\(\sV\), support of embedding-label vector pairs~\(\sS\), domain theory build throughout the embodied conversation~\(\Delta\), prior weights~\(w_p\), and grounded weights~\(w_g\) as well as a well-formed formula \(\phi\in\sL_{\mathrm{snt}}\).
\State \(\sS' \leftarrow \emptyset\) \Comment{Initialize new support}
\State \(\Delta \leftarrow \Delta \cup  \{\phi\}\) \Comment{Update domain theory}
\For{\(o,\vx\) in \(\mathrm{Zip}(\sO,\sX)\)} \Comment{Iterate over corresponding object-embedding pairs}
\State \(\mathtt{o} \leftarrow \mathrm{Const}(o)\) \Comment{Get constant of an object \(o\in\sO\)}
\State \(y \leftarrow \mathrm{Dictionary}()\) \Comment{Dictionary for symbol predictions}
\For{\(\mathtt{p}\) in \(\sV\)} \Comment{Iterate over predicate symbols}
\State \(y[\mathtt{p}] \leftarrow  \mathrm{CON}_{w_p}(\mathtt{p(o)}\mid b)\) \Comment{Conditional probability with prior weights for an atom \(\mathtt{p(o)}\)}
\EndFor
\State \(\vy \leftarrow \mathrm{Vectorize}(y)\) \Comment{Construct semantic vector of size \(|\sV|\)}
\State \(\sS' \leftarrow \sS' \cup (\vx,\vy)\) \Comment{Add exemplar to new support}
\EndFor
\State \(\sS \leftarrow \sS'\)\Comment{Update support}
\For{\(o,\vx\) in \(\mathrm{Zip}(\sO,\sX)\)} \Comment{Iterate over corresponding object-embedding pairs}
\For{\(\mathtt{p}\) in \(\sV\)} \Comment{Iterate over predicate symbols}
\State \(w_g(\mathtt{p(o)}) \leftarrow \omega_b(\vx)^{(\mathtt{p})}\) \Comment{Grounding model prediction}
\EndFor
\EndFor
\Return \(b\) 
\end{algorithmic}
\end{algorithm}

\begin{algorithm}
\caption{Dialogue Strategy Optimisation using Episodic Semi-gradient SARSA}
\label{alg:sarsa}
\begin{algorithmic}[1]
\Require{belief state \(b\in\sB\), state-action value function \(Q\) parameterised by \(\theta\), number of tasks per environment \(m\), learning rate \(\alpha\), discount factor \(\gamma\), epsilon \(\epsilon\).}
\While{ not converged}
\State \(\gE \leftarrow \mathrm{GetEnvironment()}\)
\For{ \(t\) in \(\mathrm{GetTasks}(\gE,m)\)} \Comment{Get \(m\) tasks in environment \(\gE\)}
\State \(b \leftarrow \mathrm{Update}(b,\Phi(t))\) \Comment{Update belief state (Algorithm~\ref{alg:update})}
\State \(\mathbb{A}_{\mathrm{quest}} \leftarrow \mathrm{GetQuestions}(t)\) \Comment{Coherent questions for \(t\)}
\State \(\sA \leftarrow \{a_{\mathrm{act}}\} \cup \mathbb{A}_{\mathrm{quest}}\) \Comment{Construct action space}
\While{\(\mathrm{True}\)} \Comment{Training loop}
\State \(a \leftarrow \arg\max_{a\in \sA}Q(b,a)\) \Comment{Greedy action}
\State \(R, \phi \leftarrow \mathrm{Act}(\mathcal{E},a)\) \Comment{Act in \(\mathcal{E}\) and construct \(\phi\) via semantic analysis}
\State \(b' \leftarrow \mathrm{Update}(b,\phi) \) \Comment{Update belief state (Algorithm~\ref{alg:update})}
\If{\(|R| = 1\)} \Comment{Terminating state} 
\State \(\delta \leftarrow R- Q(s,a)\)
\State \(\theta \leftarrow \theta + \alpha\delta\nabla_{\theta} Q(s,a)\)
\State \textbf{break}
\Else
\State \(a' \leftarrow \begin{cases}
    \arg\max_{a\in \sA}Q(b',a) \text{ with probability } 1 -\epsilon \\
    \mathrm{Choose}(\sA) \text{ otherwise} 
\end{cases}\) \Comment{\(\epsilon\)-greedy action}
\State \(\delta \leftarrow R + \gamma Q(b',a') - Q(b,a)\)
\State \(\theta \leftarrow \theta + \alpha\delta\nabla_{\theta} Q(b,a)\)
\EndIf
\State \(b \leftarrow b'\) \Comment{Update belief state}
\EndWhile
\EndFor
\EndWhile
\end{algorithmic}
\end{algorithm}

\begin{table}[]
\centering
\begin{tabular}{@{}lll@{}}
\toprule
Hyparameter                     & Symbol       & Value \\ \midrule
Number of tasks per environment & \(m\)        & 1     \\
Learning rate                   & \(\alpha\)   & 0.1   \\
Discount factor                 & \(\gamma\)   & 0.99  \\
Exploration factor              & \(\epsilon\) & 0.1   \\ \bottomrule
\end{tabular}
\caption{Hyperparameter values used for optimizing dialogue strategies for \(\pi_{\mathrm{secure}}\) and \(\pi_{\mathrm{simple}}\) with semi-gradient SARSA following algorithm~\ref{alg:sarsa}.}
\label{tab:sarsa_hyperparameters}
\end{table}

\newpage
\section{Experiment Details}
\label{app:experiments}
The following appendix provides additional details for simulation and real-world experiments.

\subsection{Simulation Experiments}
\label{app:simulation_experiment_details}

Simulation experiments are conducted in the MuJoCo simulator in rearranging rigid objects. 

For obtaining the logical forms of referential expressions that are part of embodied conversation messages (task instructions, questions, corrections) we used a simple semantic parser based on CodeLlama7B~\citep{DBLP:journals/corr/abs-2308-12950}\footnote{\url{https://huggingface.co/codellama/CodeLlama-7b-Instruct-hf}} with in-context learning and prompt given in \figref{fig:prompt}. For the grounding model, we utilize object-centric embeddings by first using ground-truth object locations from the simulator to localize patches in the environment's top-down view (as in \figref{fig:top_down}), from which the feature vectors \(\vx\in\sR^{384}\) are obtained using DINOv2~\citep{DBLP:journals/corr/abs-2304-07193}\footnote{\url{https://huggingface.co/facebook/dinov2-small}} encoder. In inference, \(\tau=0.65\) is set as the threshold for adding examples to the positive/negative support, which results in an example being added to the negative support if \(\hat{y}^{(\mathtt{p})}\leq 0.354\) and added to the positive support if \(\hat{y}^{(\mathtt{p})}\geq 0.646\).

Agent's dialogue strategy is given by action-value function (Eq.\ref{eq:Q-values}) which is optimized offline. To learn dialogue strategy parameters \(\theta=[\theta_1,\theta_2]^{\top}\) for \(\pi_{\mathrm{secure}}\) and \(\pi_{\mathrm{simple}}\) (\(\pi_{\mathrm{correct}}\) does not make interaction-level decisions and just uses corrective feedback), we used semi-gradient SARSA \citep{rummery:cuedtr94} for policy optimization using tasks in simulation whose instructions feature subsets of domain concepts. In particular, we use all quantifiers and spatial relationships elicited in \figref{fig:simulation_experiments} but only a subset of properties (plain, cube, red, green, and blue). For the simulation experiments, we set unit costs of symbol designation and reference resolution to be the same and small to encourage exploration when learning new concepts. In particular we set \(C_\mathrm{point}=C_\mathrm{ref}=0.1\). 

When the agent decides to rearrange the environment, given by choosing an action \(a_{\mathrm{act}}\in\sA\), this manipulation is achieved using an operational space controller \cite{DBLP:journals/trob/Khatib87} and a rudimentary logic for setting control targets based on the pose of the object of interest. After each action, the oracle is checked, and in case of the suboptimal action, a correction is issued, and the previous action is undone (e.g. returning the picked object to the original position).

\begin{figure}
    \centering
    \begin{tcolorbox}[title=Prompt for Semantic parsing.]
    [INST] You are a helper in a virtual assistant. You are asked to translate referential expressions to their logical forms. Referential Expressions like ``the one red block'' have a logical form  \(\mathtt{\langle\_the\_1\_q\text{ }x. red(x) \wedge block(x)\rangle}\). Logical forms follow the pattern \(\langle\mathtt{[Quantifier] [Variable].\text{ }[Formula]}\rangle\) \(\mathtt{[Quantifier]}\) is a quantifier. We consider the following quantifiers:
    \begin{itemize}
        \item existential with surface form: ``a/an'' and denoted as logic symbol: \(\mathtt{\_a\_q}\)
        \item universal with surface form: ``every/all'' and denoted as a logic symbol: \(\mathtt{\_every\_q}\)
        \item uniqueness with surface form: ``the n'' and denoted as a logic symbol: \(\mathtt{\_the\_n\_q}\), where n is a natural number like ``one'', ``two'', etc.
    \end{itemize}
\(\mathtt{[Variable]}\) is a variable, e.g. \(\mathtt{x}\),  \(\mathtt{x1}\),  \(\mathtt{x12}\). Note that \(\mathtt{[Variable]}\) is the only free variable in \(\mathtt{[Formula]}\) not bound by a quantifier. if the formula has only one variable, it should be named \(\mathtt{x}\).

\(\mathtt{[Formula]}\) is a formula of predicate logic. Each formula is constructed recursively:
\begin{itemize}
    \item Predicates like \(\mathtt{red(x)}\), \(\mathtt{above(x1,x2)}\), \(\mathtt{left(x1,x3)}\) is a well-formed formula \(\mathtt{[Formula]}\)
    \item Negation of \(\mathtt{[Formula]}\) like \(\mathtt{neg(red(x))}\) is a well-formed formula \(\mathtt{[Formula]}\)
    \item Logical conjunction of \(\mathtt{[Formula]}\)s like \(\mathtt{red(x)\wedge block(x)}\) is a well-formed formula \(\mathtt{[Formula]}\)
    \item Logical disjunction of \(\mathtt{[Formula]}\)s like \(\mathtt{red(x)\vee block(x)}\) is a well-formed formula \(\mathtt{[Formula]}\)
    \item Logical implication of \(\mathtt{[Formula]}\)s like \(\mathtt{red(x)\rightarrow block(x)}\) is a well-formed formula \(\mathtt{[Formula]}\)
    \item Logical structure for generalised quantifiers \(\mathtt{[Quantifier][Variable].([Formula],[Formula])}\) like \(\mathtt{\_the\_1\_q x. (red(x), block(x))}\) is a well-formed formula \(\mathtt{[Formula]}\)
\end{itemize}
Here are some examples of what these logical forms of referential expressions look like: \\
RefExp, LF-RefExp,\\
``a block.'', \(\mathtt{\_a\_q\text{ }x.block(x)\rangle}\)\\
``the one block.'', \(\mathtt{\langle\_the\_1\_q\text{ }x. block(x)\rangle}\)\\
``the two plain objects.'', \(\mathtt{\langle\_the\_2\_q\text{ }x. plain(x) \wedge object(x)\rangle}\)\\
``every magenta sphere.'' \(\mathtt{\langle\_every\_q\text{ }x. magenta(x) \wedge sphere(x)\rangle}\) \\
``not a block above a sphere.'', \(\mathtt{\langle\_a\_q\text{ }x.neg(\_a\_q\text{ }x1.(sphere(x1),block(x) \wedge above(x,x1)))\rangle}\) \\
``a sphere to the left of every green cone.'', \\ \(\mathtt{\langle\_a\_q\text{ }x. \_every\_q\text{ }x1.(green(x1) \wedge cone(x1),sphere(x) \wedge left(x,x1))\rangle}\), \\
``every sphere to the left of every green object.'', \\ \(\mathtt{\langle\_every\_q\text{ }x. \_every\_q\text{ }x1.(green(x1) \wedge object(x1),sphere(x) \wedge left(x,x1))\rangle}\),\\
``a sphere to the right of the two green cones.'', \\ \(\mathtt{\langle\_a\_q\text{ }x. \_the\_2\_q\text{ }x1.(green(x1) \wedge cone(x1),sphere(x) \wedge right(x,x1))\rangle}\) \\
``the one sphere in front of every green cone.'', \\ \(\mathtt{<\_the\_1\_q\text{ }x. \_every\_q\text{ }x1.(green(x1) \wedge cone(x1),sphere(x) \wedge front(x,x1))\rangle}\)\\
``the two spheres behind a green cone.'', \\ \(\mathtt{\langle\_the\_2\_q\text{ }x. \_a\_q\text{ }x1.(green(x1) \wedge cone(x1),sphere(x) \wedge behind(x,x1))\rangle}\)\\
Now, please translate the following referential expression to its logical form. Just give the logical form. No extra information.
Referential expression: \{refexp\} [/INST]    
    \end{tcolorbox}
    \caption{Prompt for parsing referential expressions to their logical forms. \{refexp\ denotes the referential expression to be parsed using this prompt.}
    \label{fig:prompt}
\end{figure}

\subsection{Real-world Experiment}
\label{app:real_world_experiment_details}

For real-world experiments, we modify several aspects of the experimental setup compared to simulation experiments while keeping the rest the same. In the real-world, we use the ZED 2i stereo camera (just the left feed) as the visual sensor. Using it, grounding DINO~\citep{DBLP:journals/corr/abs-2303-05499}\footnote{\url{https://huggingface.co/IDEA-Research/grounding-dino-tiny}}: an open-set grounding model is given the prompt ``granny smith apples'' that returns candidate patches from the environment. We set the threshold for prediction retrieval to 0.3 (so regions that are identified to locate ``granny smith apples'' below 0.3 are not considered). Each patch has an associated object-centric embedding \(\vx\in\sR^{256}\). For the apple manipulation, we set  \(C_\mathrm{point}=0.2\) and \(C_\mathrm{ref}=0.6\) to make querying more expensive and encourage the agent to query only on high uncertainty situations---these arise at the beginning of interaction when a neologism is encountered. When the agent decides to execute the plan, it performs pick-and-place moves using the MoveIt~\citep{chitta2016moveit} to plan and execute collision-free trajectories with pick-and-place locations determined using Transporter Networks~\citep{DBLP:conf/corl/ZengFTWCAAKDSL20} trained on generic pick-and-place tasks. 

Tables~\ref{tab:envs_labels_preds} and~\ref{tab:traces} provide additional visualization of the real-world environment, task setup, and qualitative behaviour differences via learning traces.

\begin{table}[h]
    \centering
    \vspace{10pt}
    \begin{subfigure}{\textwidth}
        \centering
        \includegraphics[width=0.33\textwidth]{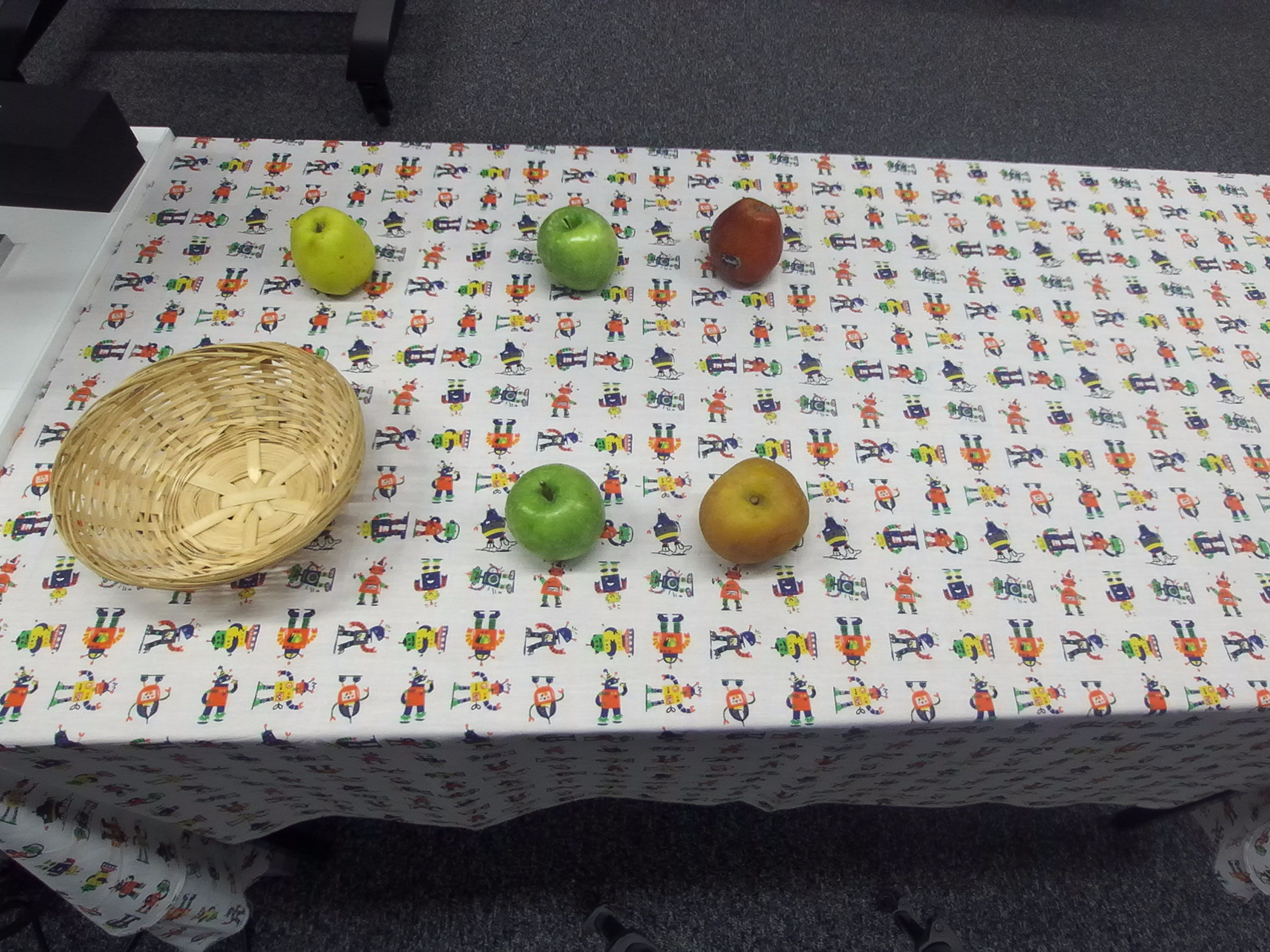}
        \includegraphics[width=0.33\textwidth]{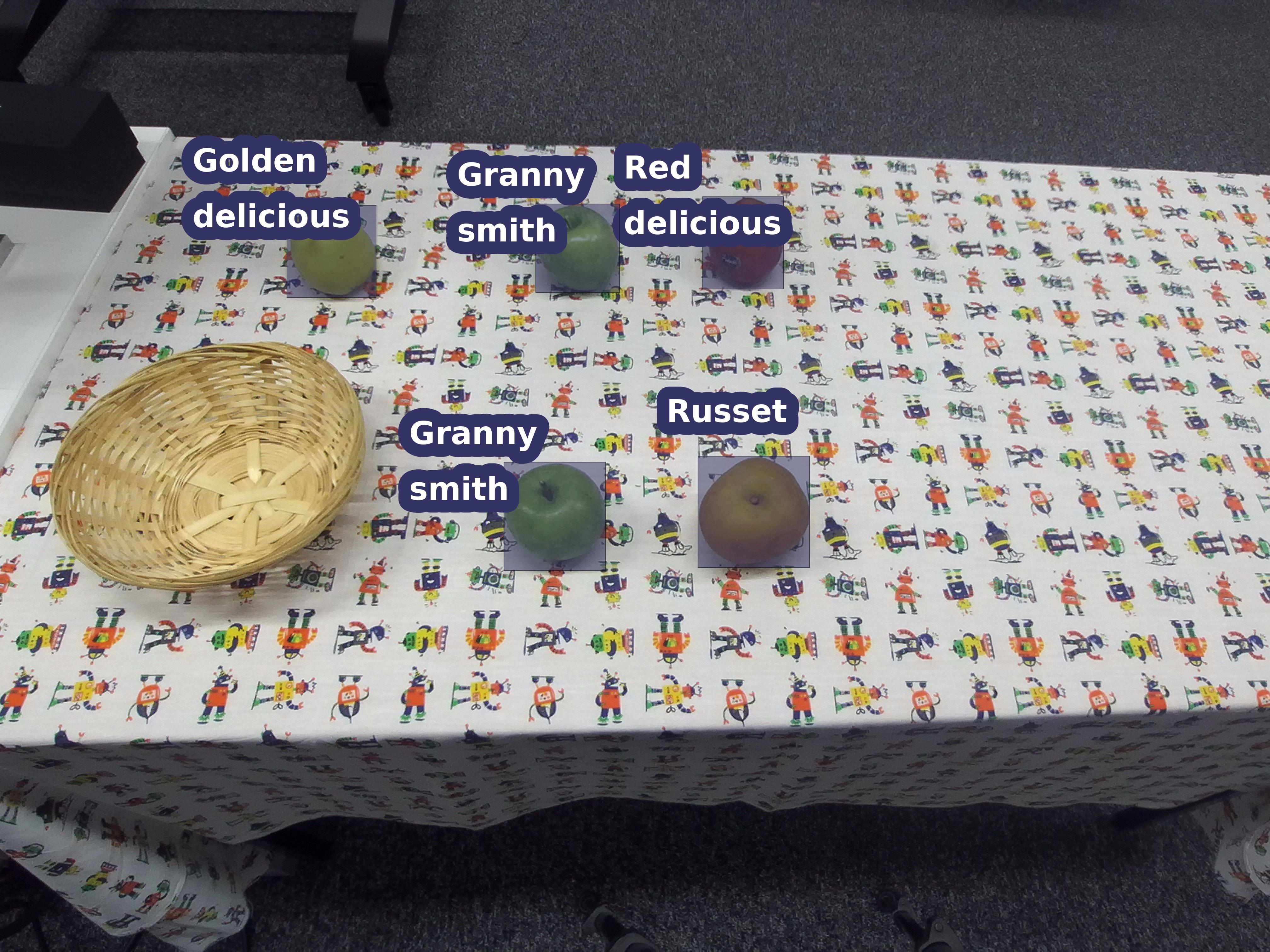}
        \includegraphics[width=0.33\textwidth]{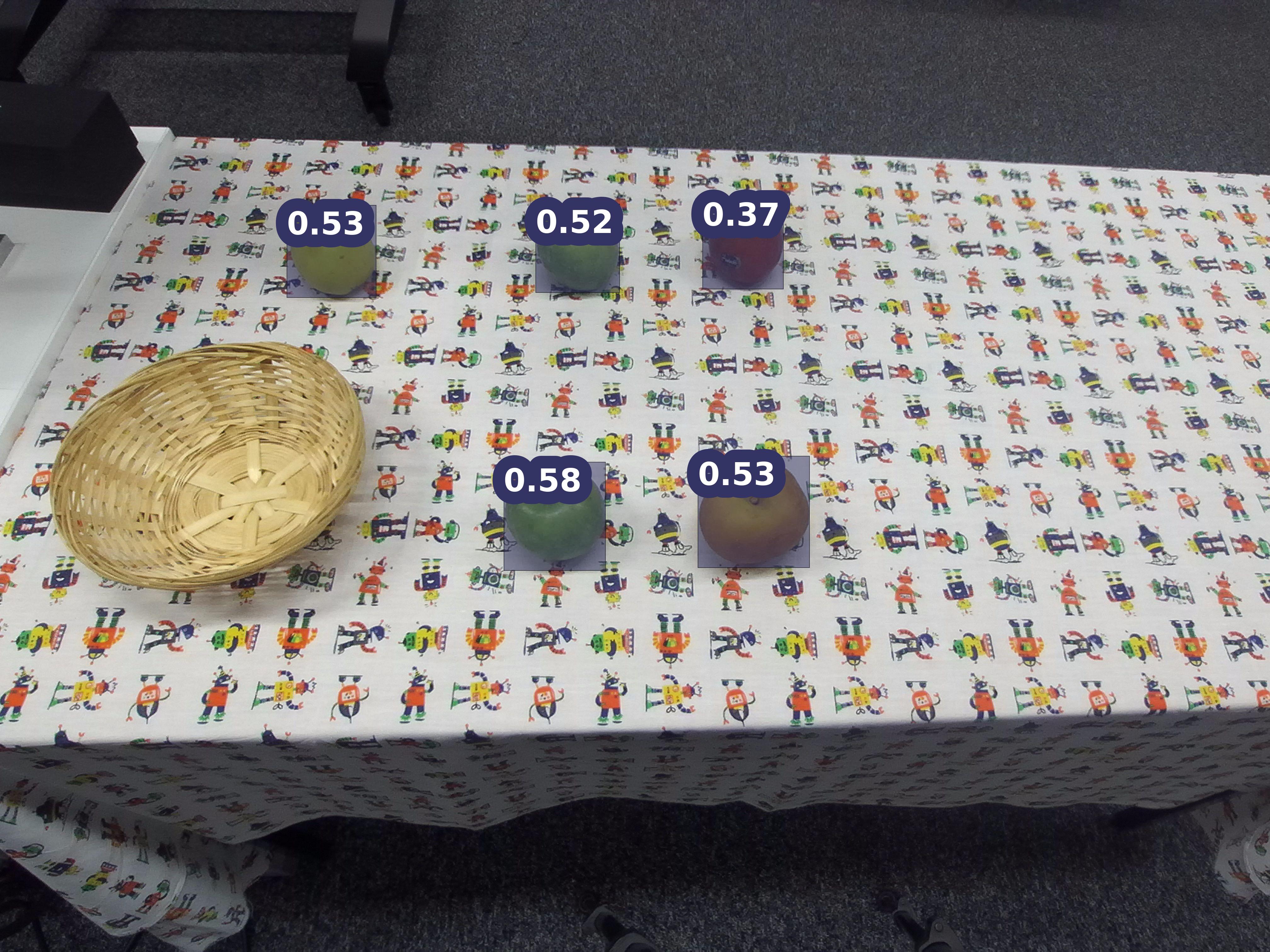}
        \caption{Environment 1}
    \end{subfigure}
    \vspace{10pt}
    \begin{subfigure}{\textwidth}
        \centering
        \includegraphics[width=0.33\textwidth]{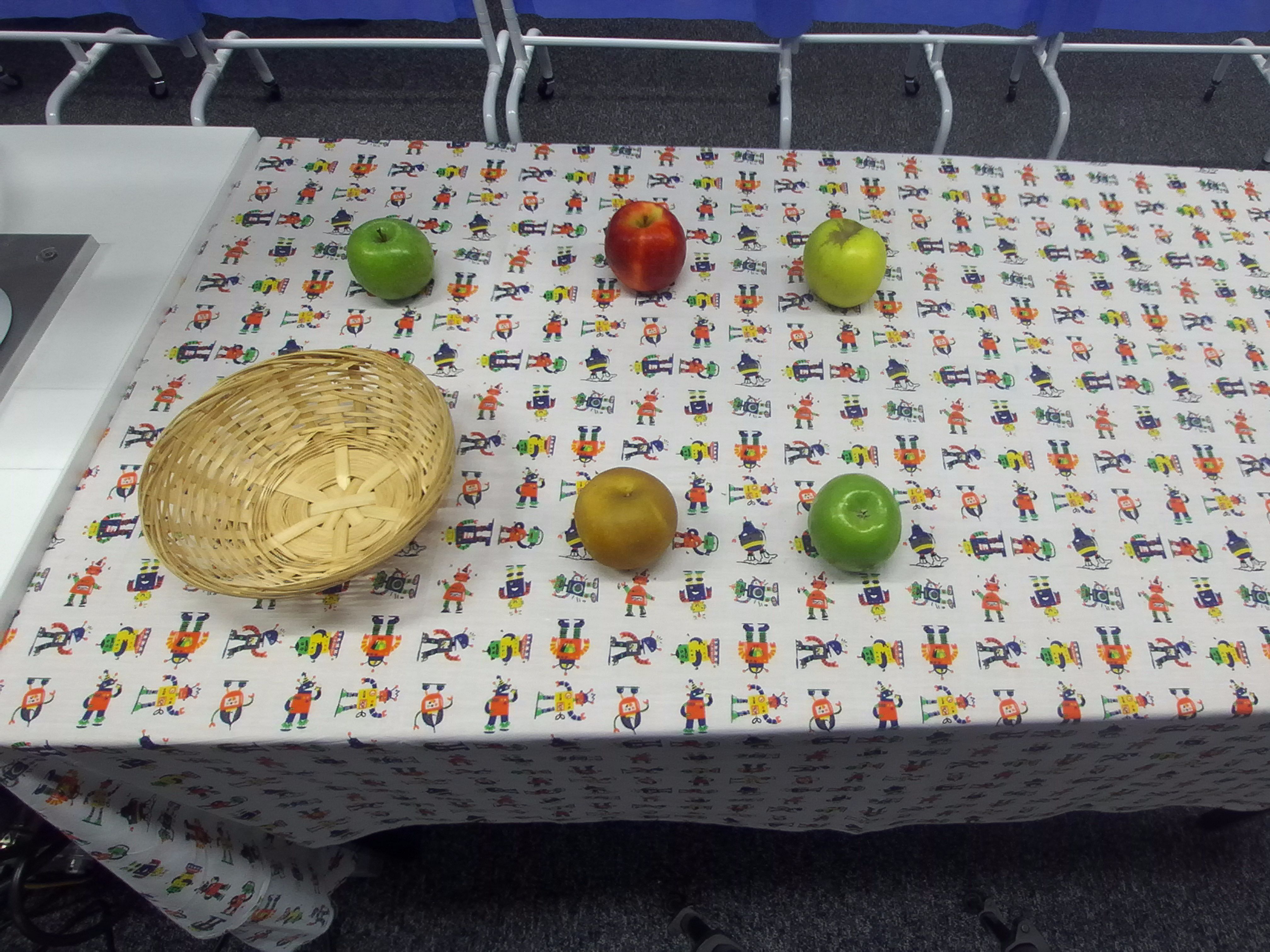}
        \includegraphics[width=0.33\textwidth]{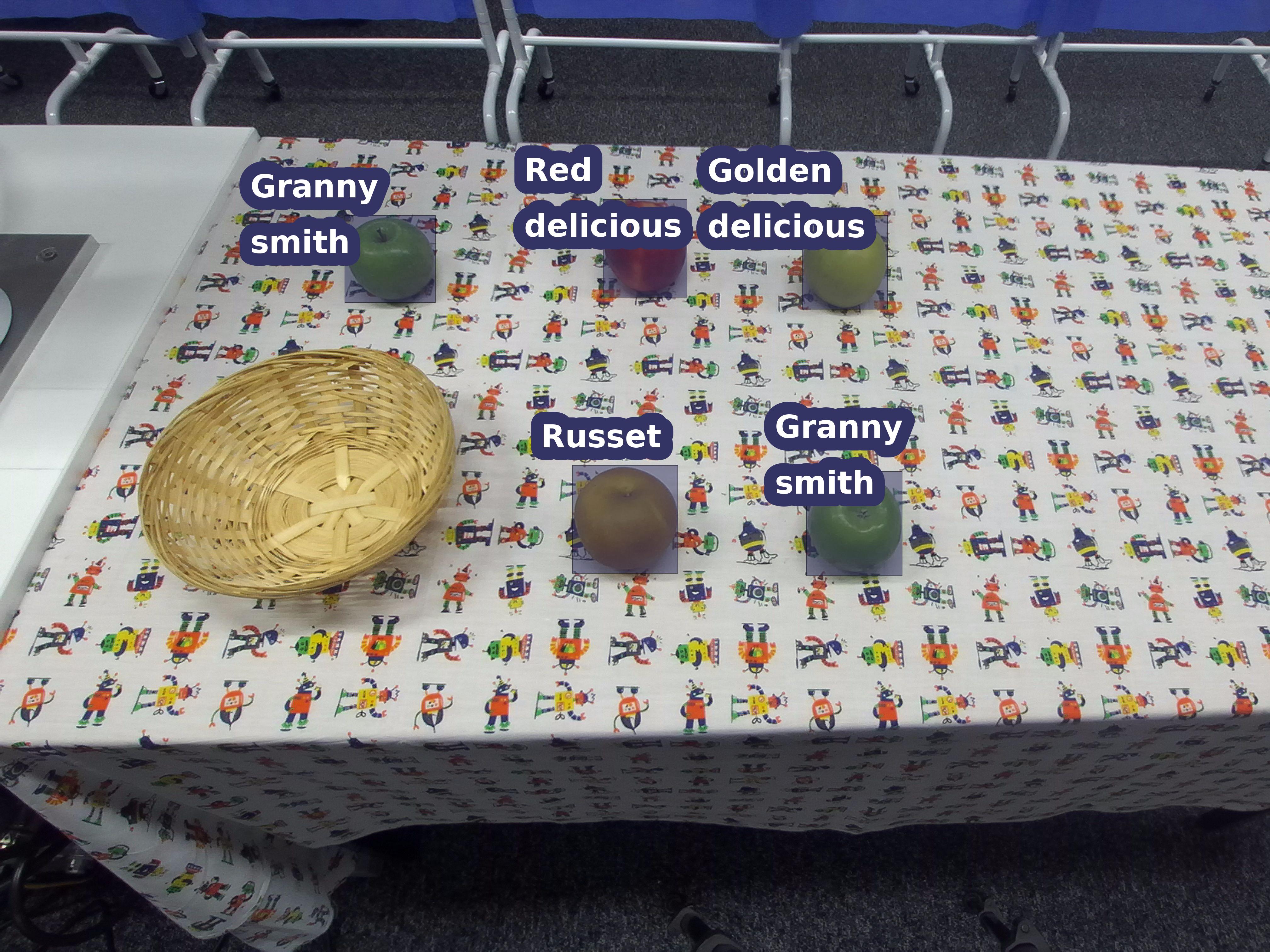}
        \includegraphics[width=0.33\textwidth]{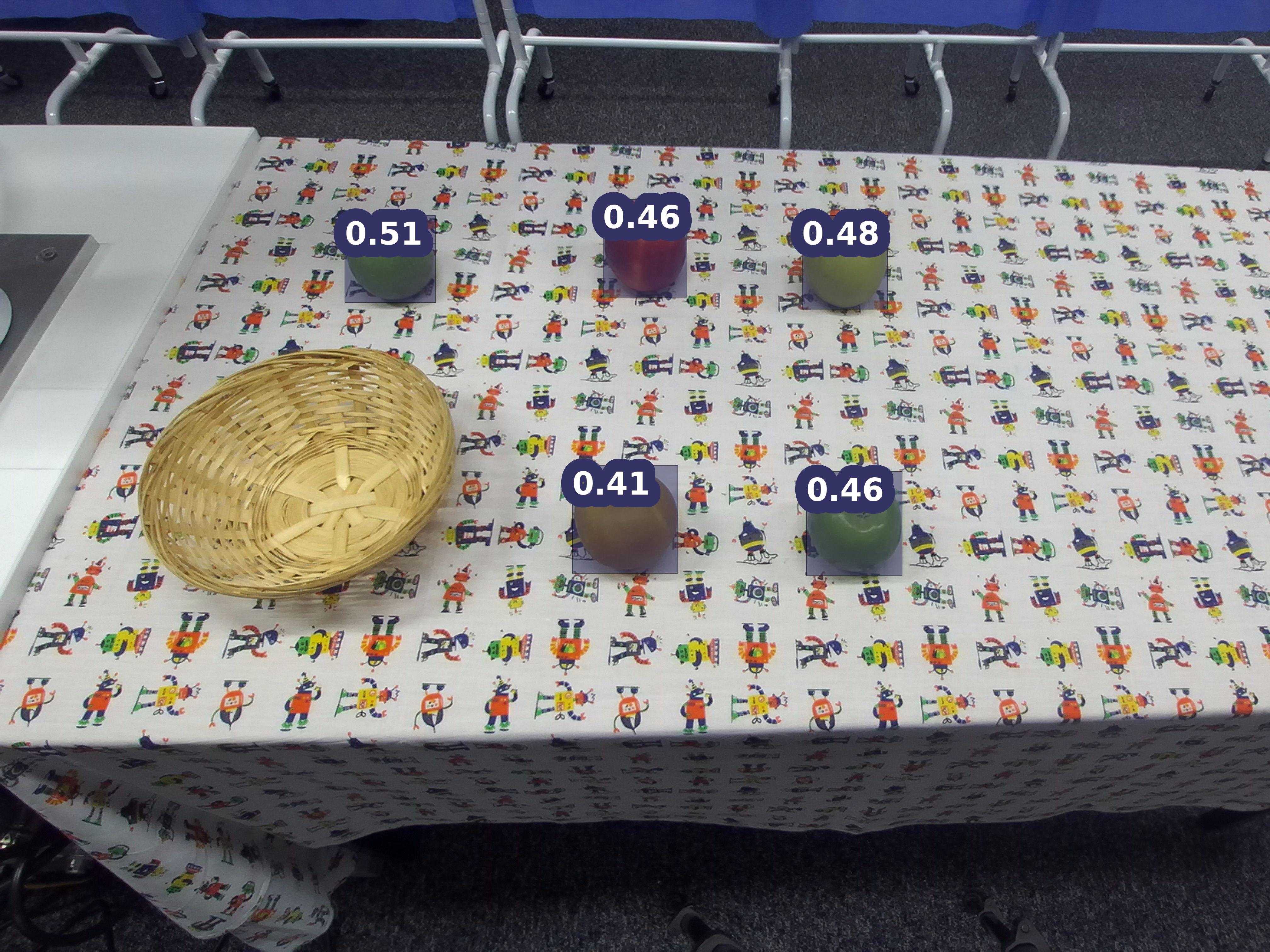}
        \caption{Environment 2}
    \end{subfigure}
    \vspace{10pt}
    \begin{subfigure}{\textwidth}
        \centering
        \includegraphics[width=0.33\textwidth]{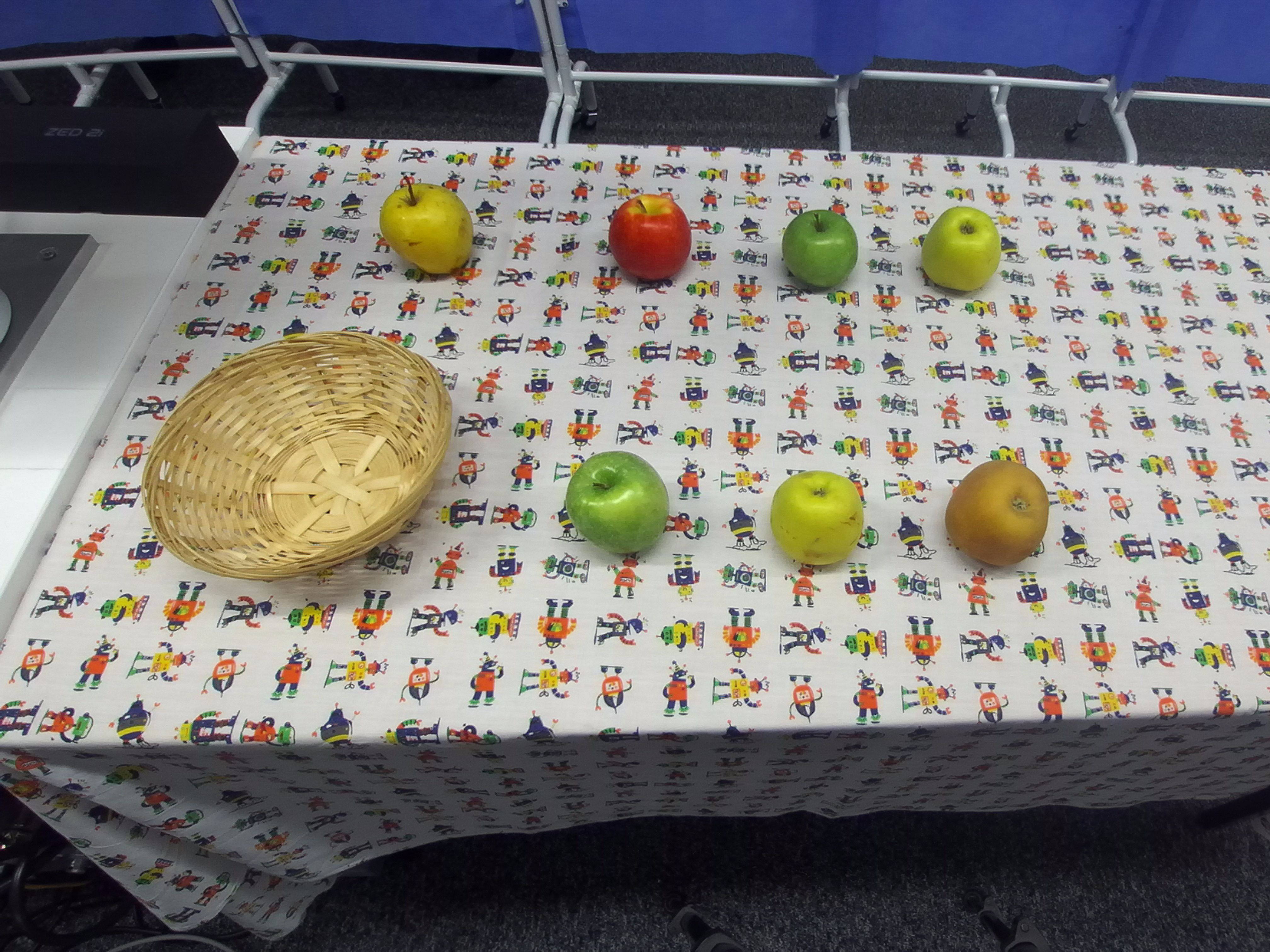}
        \includegraphics[width=0.33\textwidth]{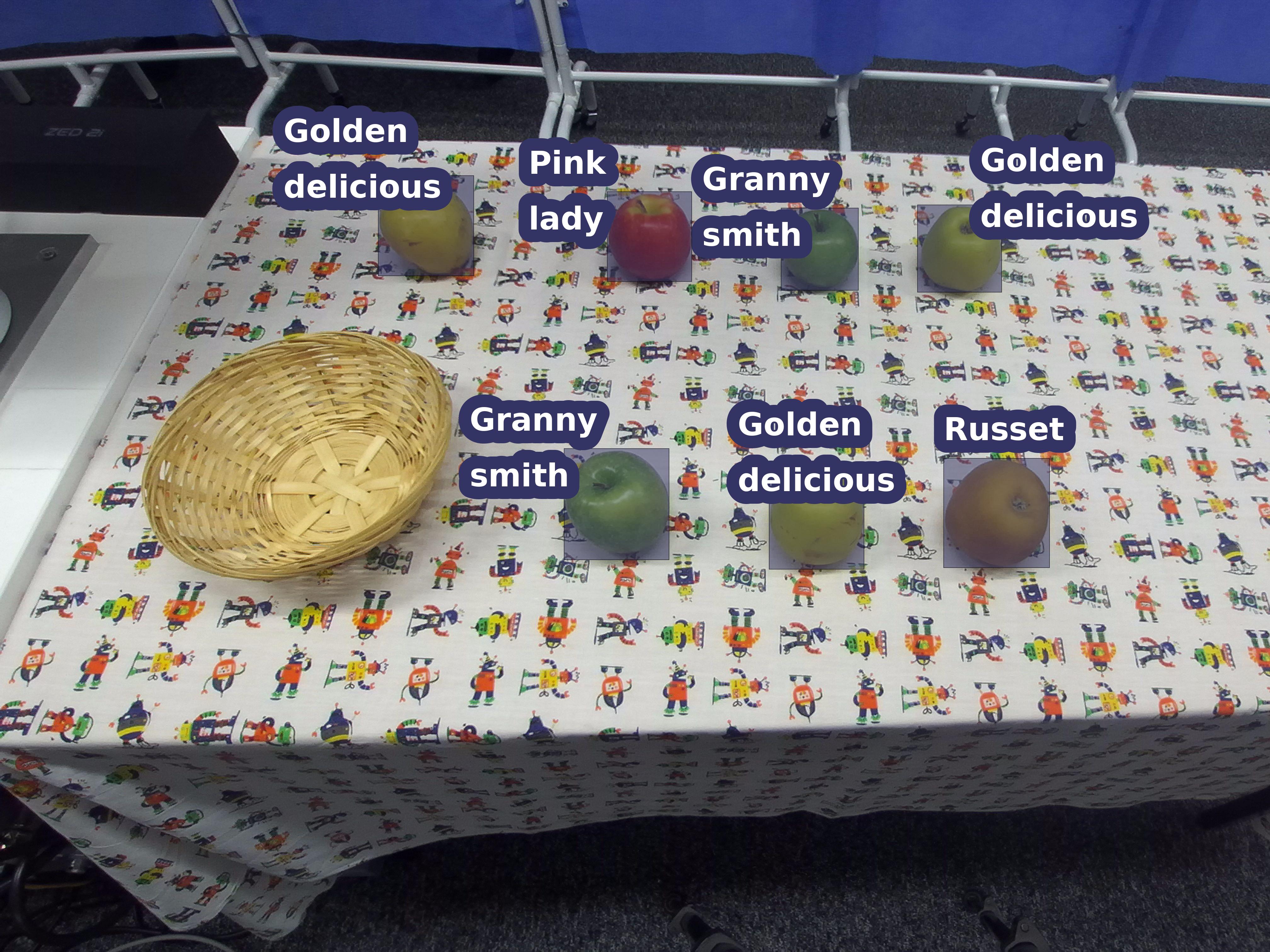}
        \includegraphics[width=0.33\textwidth]{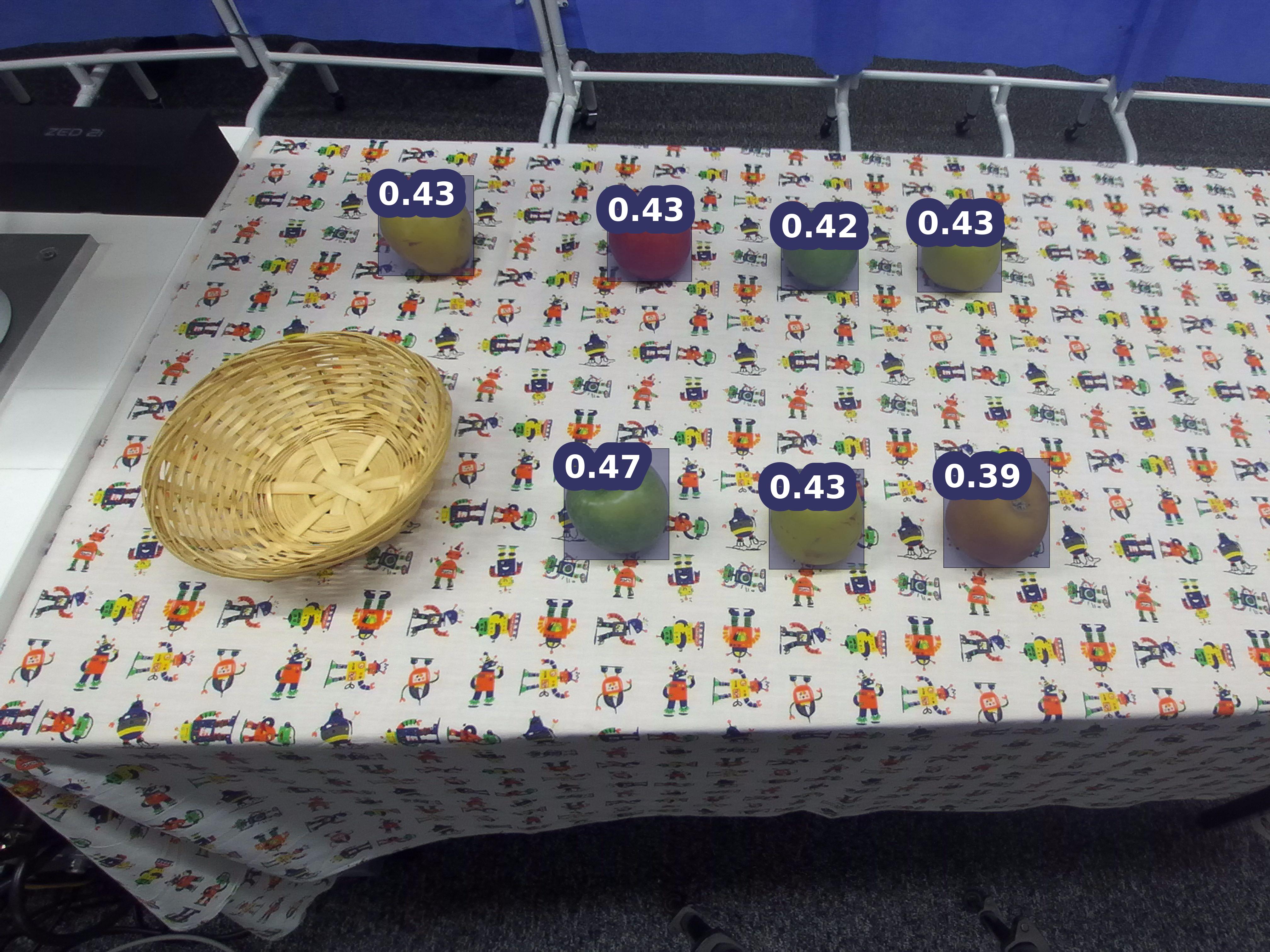}
        \caption{Environment 3}
    \end{subfigure}
    \vspace{10pt}
    \begin{subfigure}{\textwidth}
        \centering
        \includegraphics[width=0.33\textwidth]{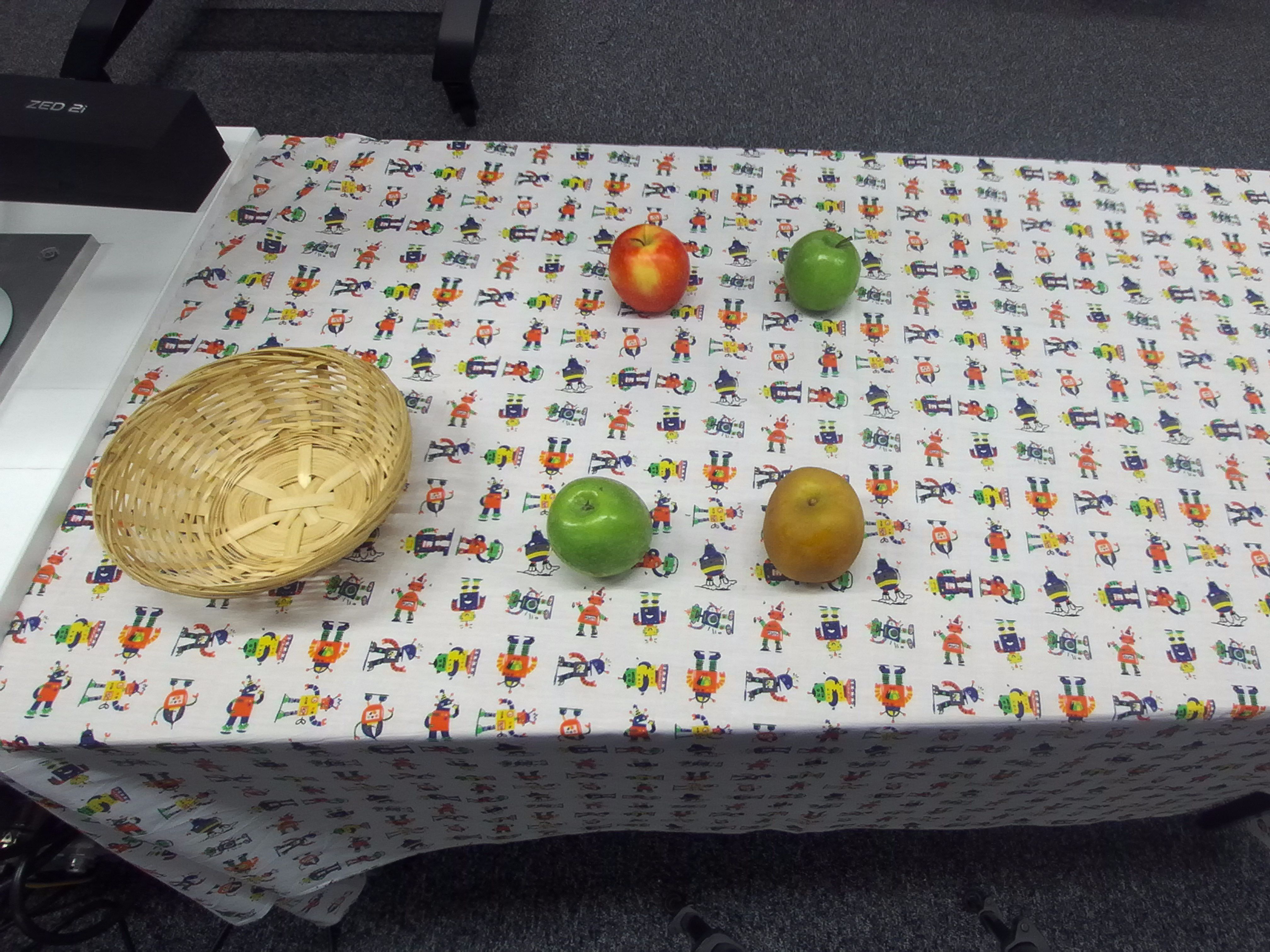}
        \includegraphics[width=0.33\textwidth]{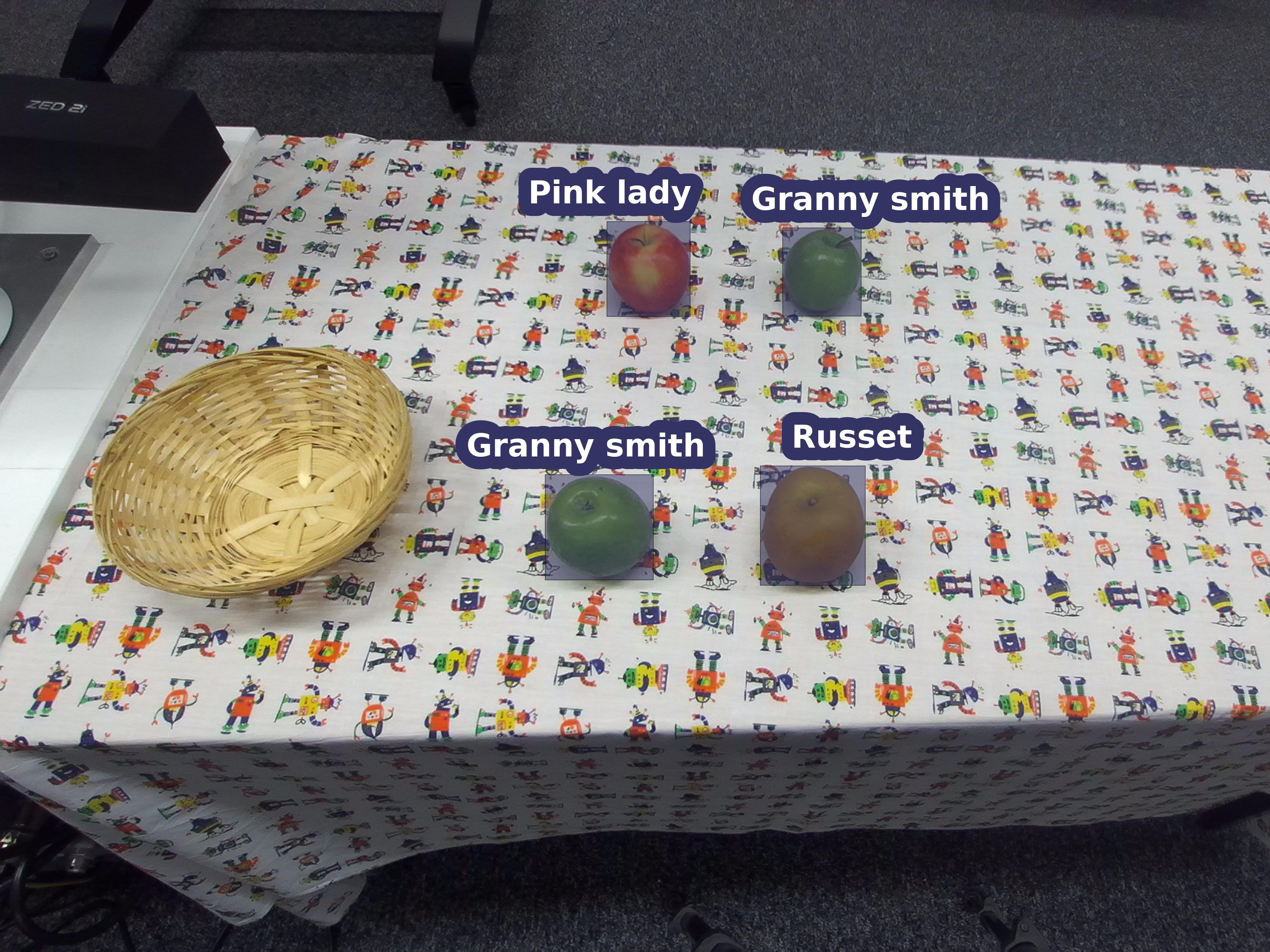}
        \includegraphics[width=0.33\textwidth]{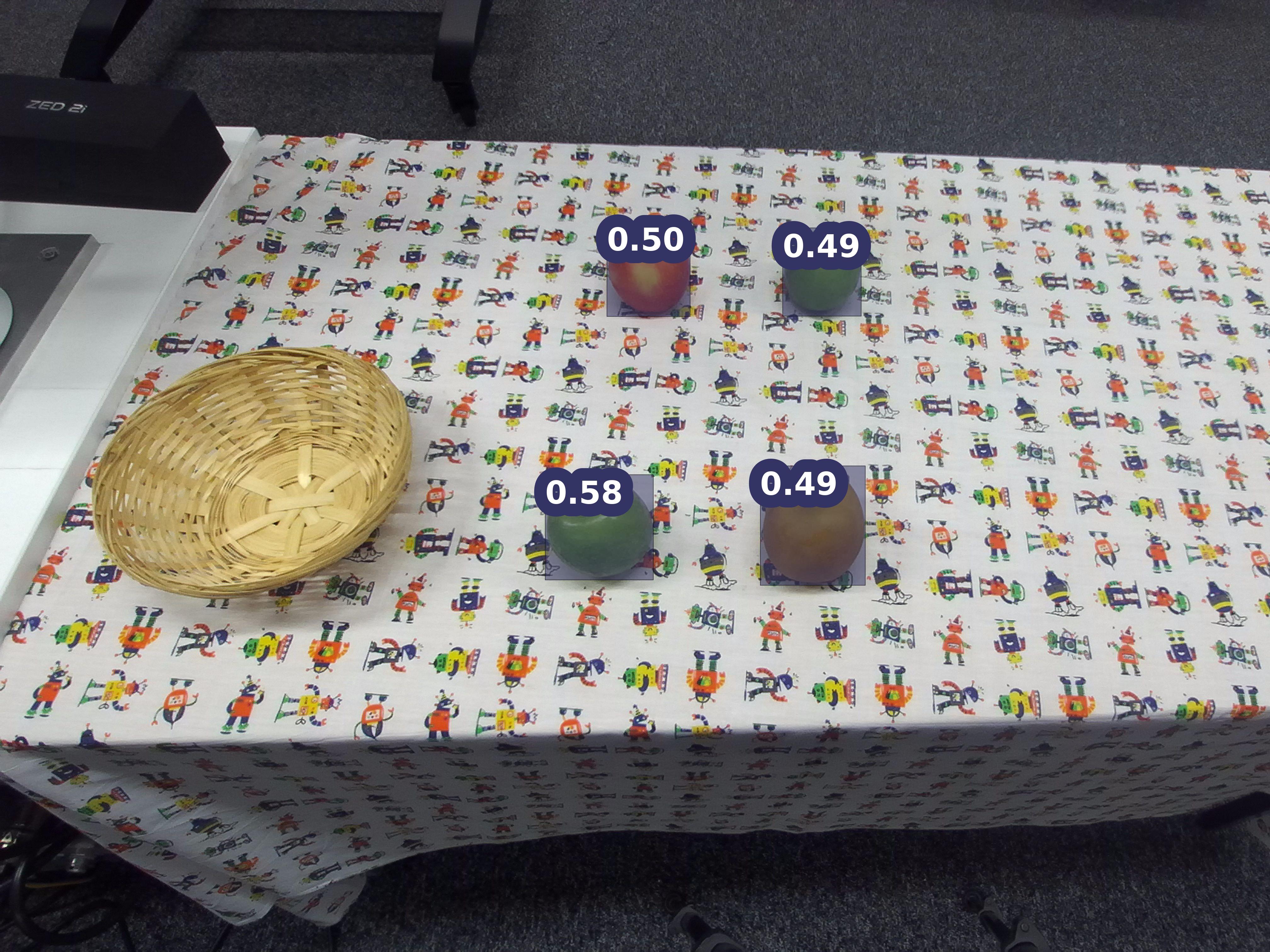}
        \caption{Environment 4}
    \end{subfigure}

    \caption{Visualisation of apple manipulation environments, their ground-truth labels, and initial grounding DINO predictions \textit{before} interaction with the environment and the user via embodied conversation.}
    \label{tab:envs_labels_preds}

\end{table}

\begin{table}[]
\centering
\renewcommand{\arraystretch}{1.5} 
\begin{tabular}{@{}m{0.25\textwidth} m{0.4\textwidth} m{0.25\textwidth}@{}}
\toprule
Initial belief state & Embodied conversation & Updated belief state \\ \midrule

\multicolumn{3}{c}{\(\pi_{\mathrm{correct}}\)} \\

\includegraphics[width=0.23\textwidth]{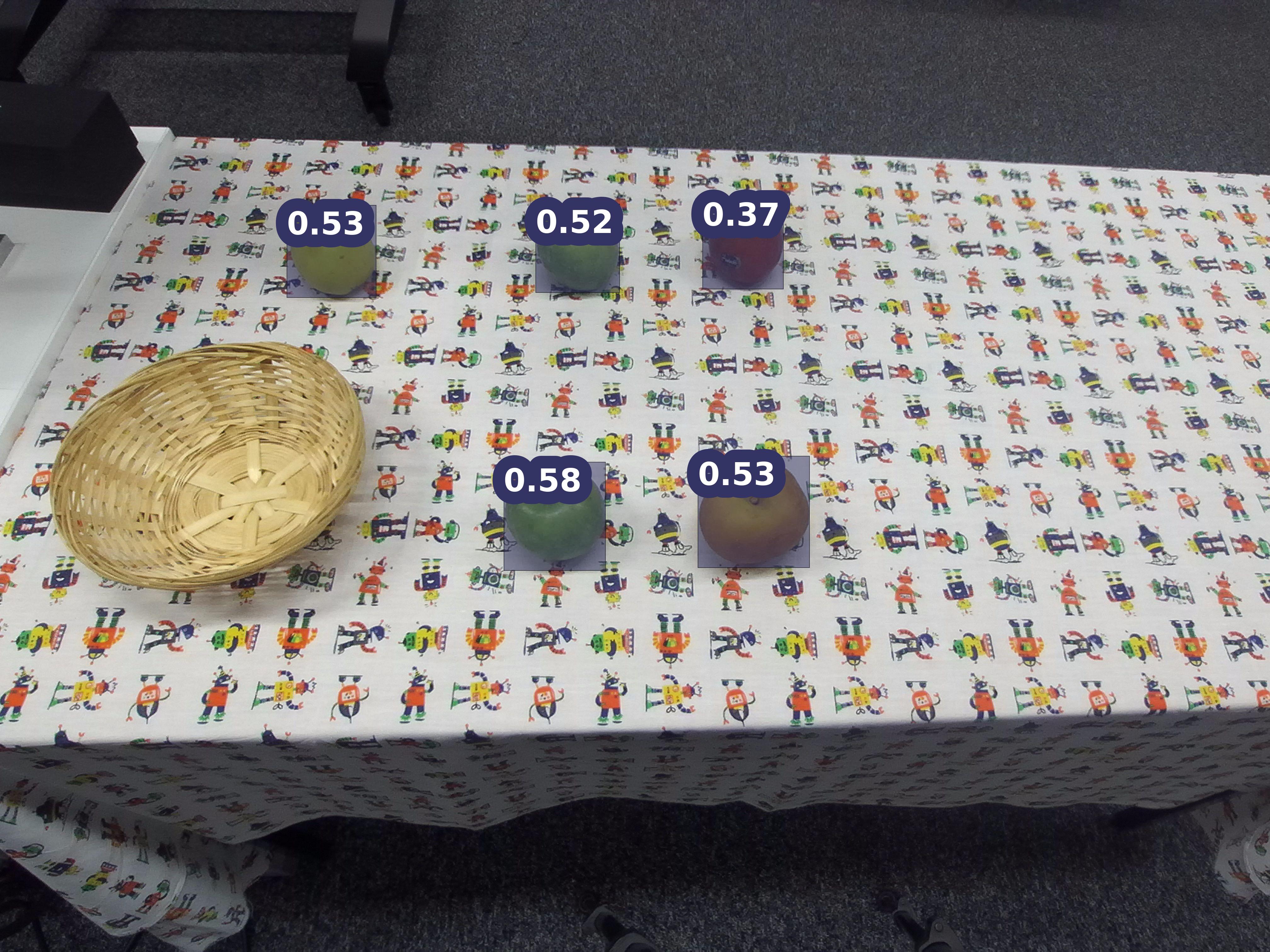} &
\fontsize{8}{9} \selectfont
User: Put the two granny smiths inside the basket. \newline
Agent: Okay. Let's try to achieve the task \newline
User: No. This is a golden delicious. \newline
Agent. Sorry. Let's try again. \newline
User: No. This is a russet. \newline
Agent. Sorry. Let's try again. \newline
User: No. This is a red delicious. \newline
Agent. Sorry. Let's try again. \newline
Agent: I have finished the task. \newline
User: Correct. 
&
\includegraphics[width=0.23\textwidth]{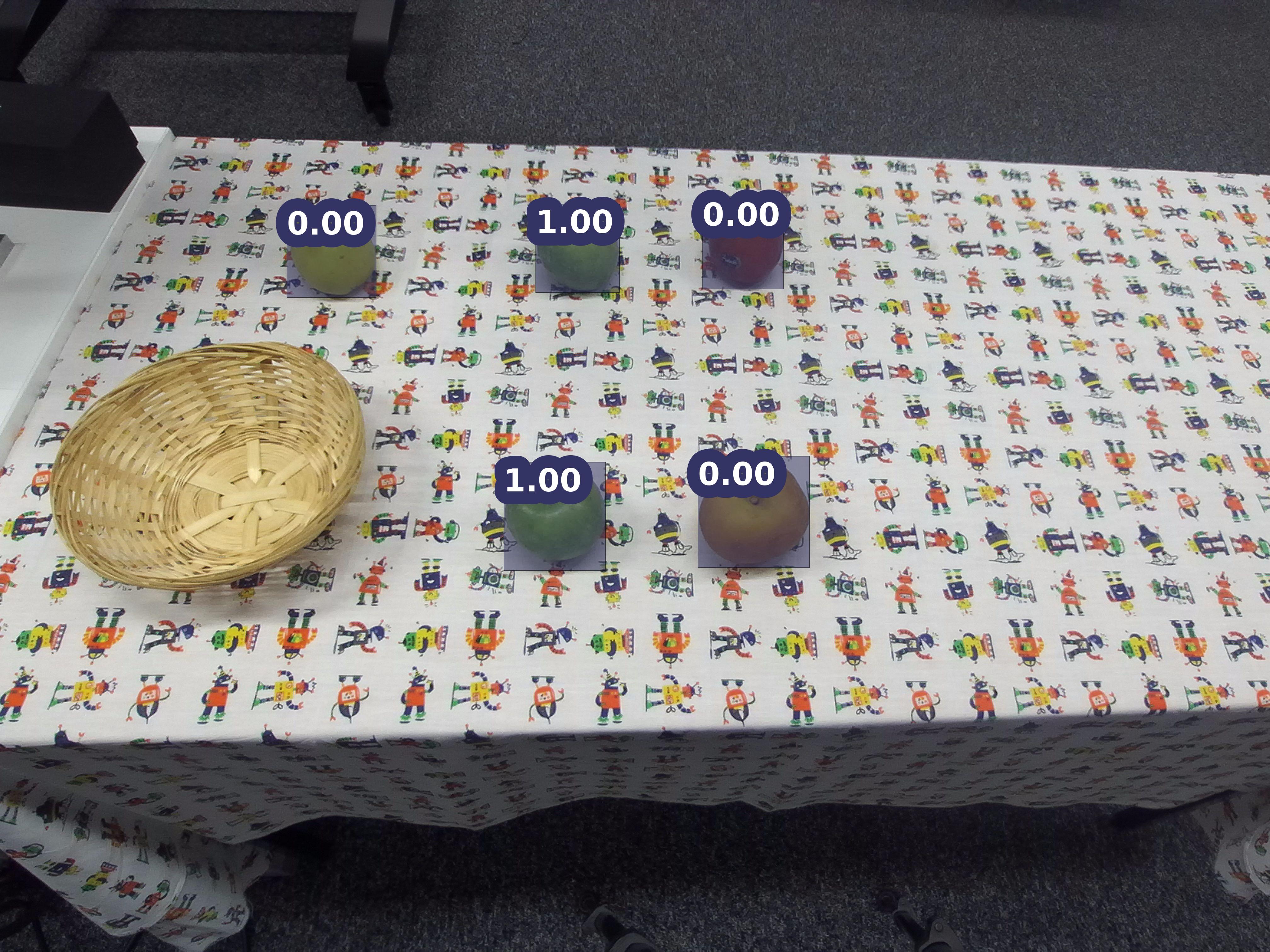} \\

\includegraphics[width=0.23\textwidth]{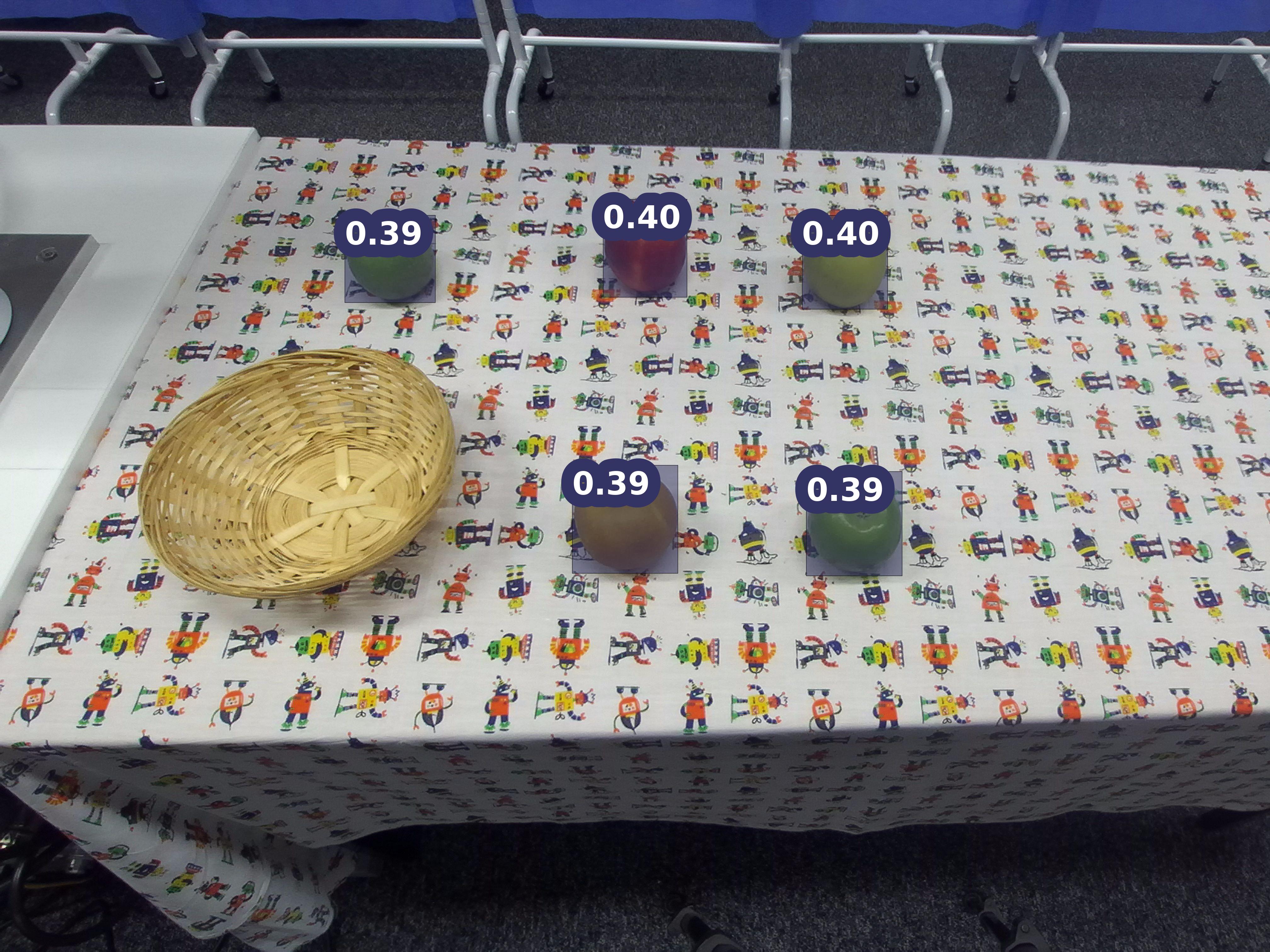} &
\fontsize{8}{9} \selectfont
User: Put the two granny smiths inside the basket. \newline
Agent: Okay. Let's try to achieve the task \newline
User: No. This is a red delicious. \newline
Agent. Sorry. Let's try again. \newline
User: No. This is a golden delicious. \newline
Agent. Sorry. Let's try again. \newline
Agent: I have finished the task. \newline
User: Correct. 
&
\includegraphics[width=0.23\textwidth]{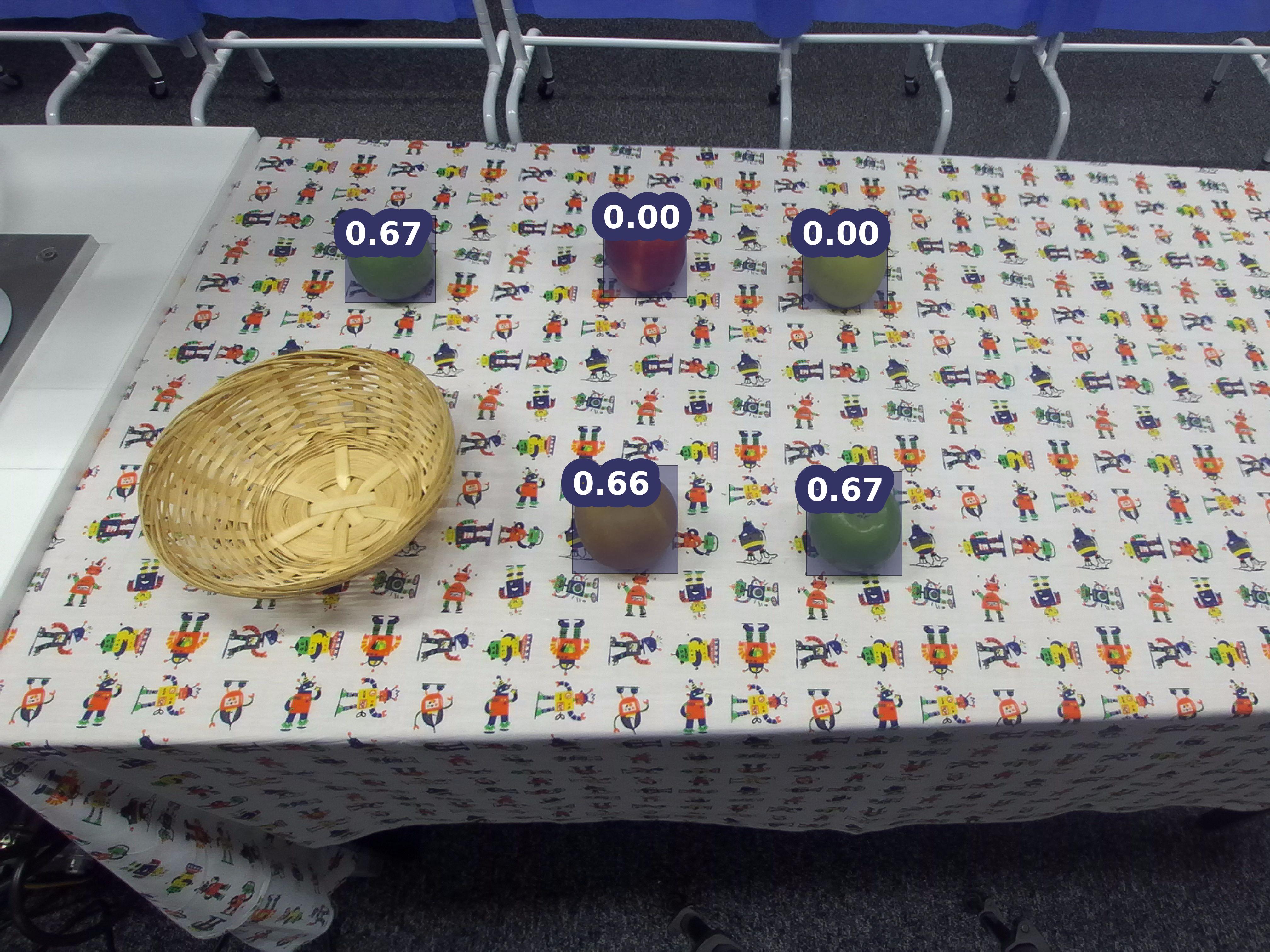} \\ 

\midrule
\multicolumn{3}{c}{\(\pi_{\mathrm{simple}}\)} \\

\includegraphics[width=0.23\textwidth]{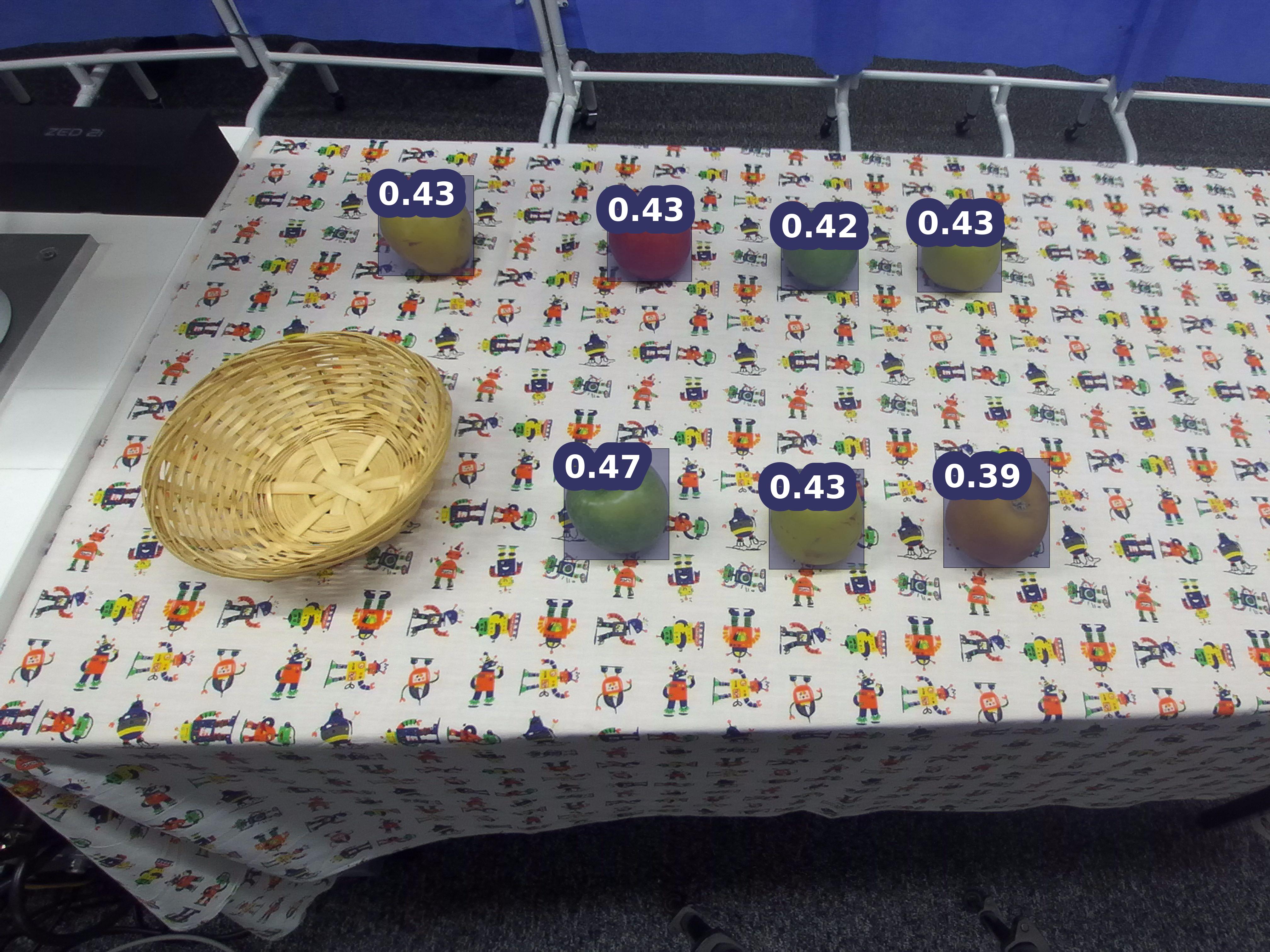} &
\fontsize{8}{9} \selectfont
User: Put the two granny smiths inside the basket. \newline
Agent: Before that, show me a granny smith. \newline
User: (designates one of the granny smiths) \newline 
Agent: Okay. Let's try to achieve the task \newline
User: No. This is a golden delicious. \newline
Agent. Sorry. Let's try again. \newline
User: No. This is a golden delicious. \newline
Agent. Sorry. Let's try again. \newline
Agent: I have finished the task. \newline
User: Correct. 
&
\includegraphics[width=0.23\textwidth]{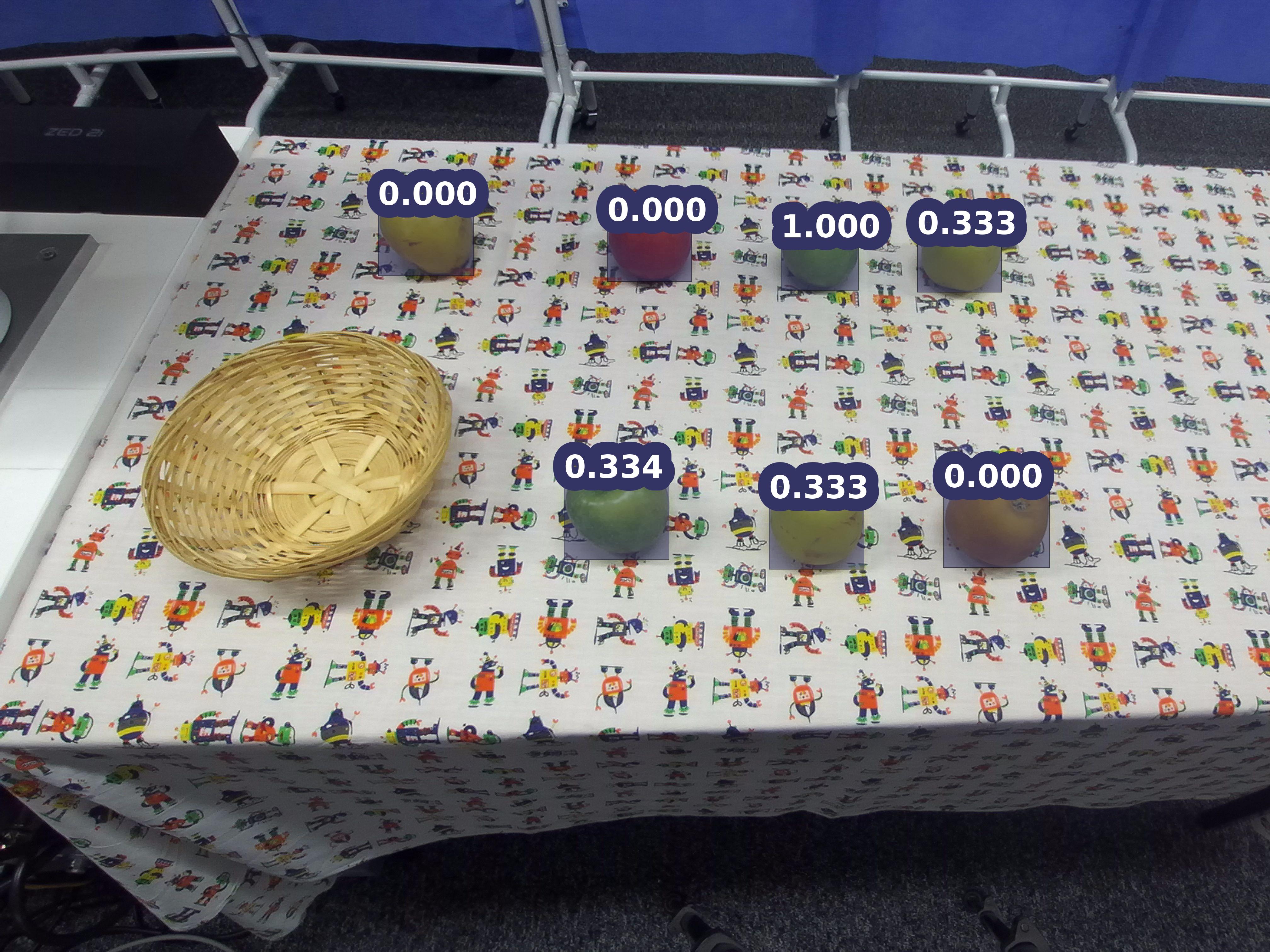} \\

\includegraphics[width=0.23\textwidth]{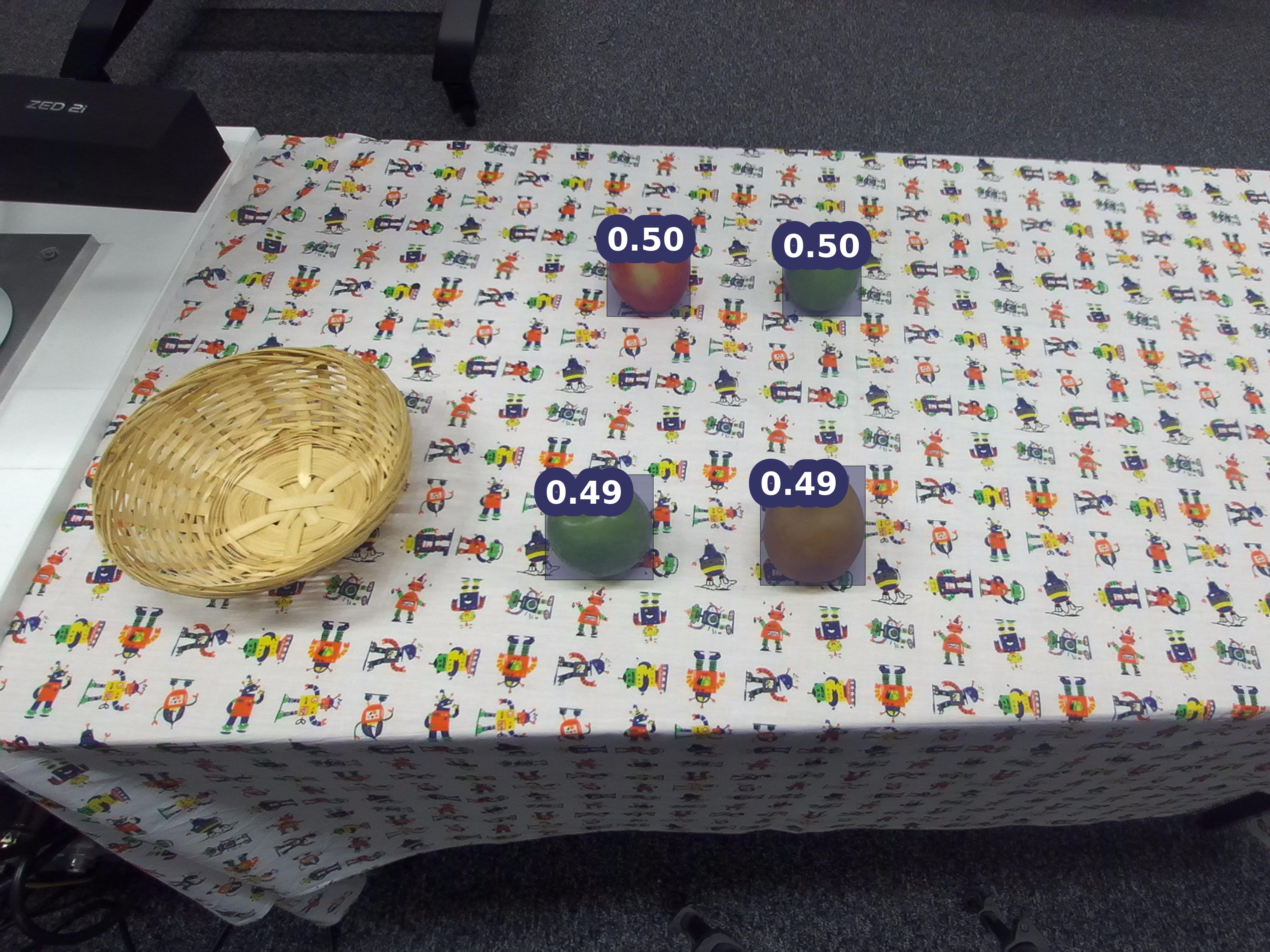} &
\fontsize{8}{9} \selectfont
User: Put the two granny smiths inside the basket. \newline
Agent: Okay. Let's try to achieve the task \newline
User: No. This is a russet. \newline
Agent. Sorry. Let's try again. \newline
Agent: I have finished the task. \newline
User: Correct. &
\includegraphics[width=0.23\textwidth]{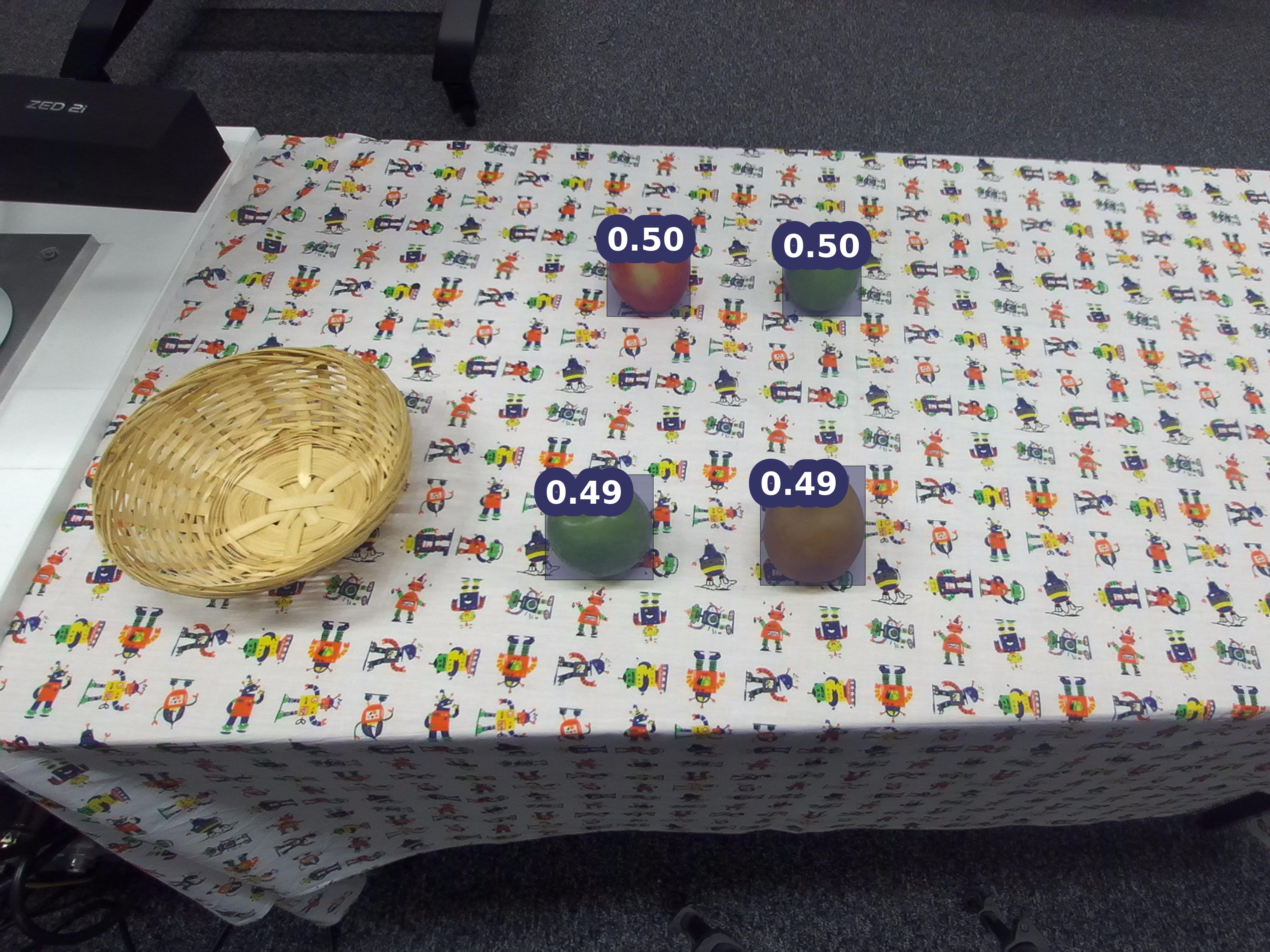} \\ 

\midrule
\multicolumn{3}{c}{\(\pi_{\mathrm{secure}}\)} \\

\includegraphics[width=0.23\textwidth]{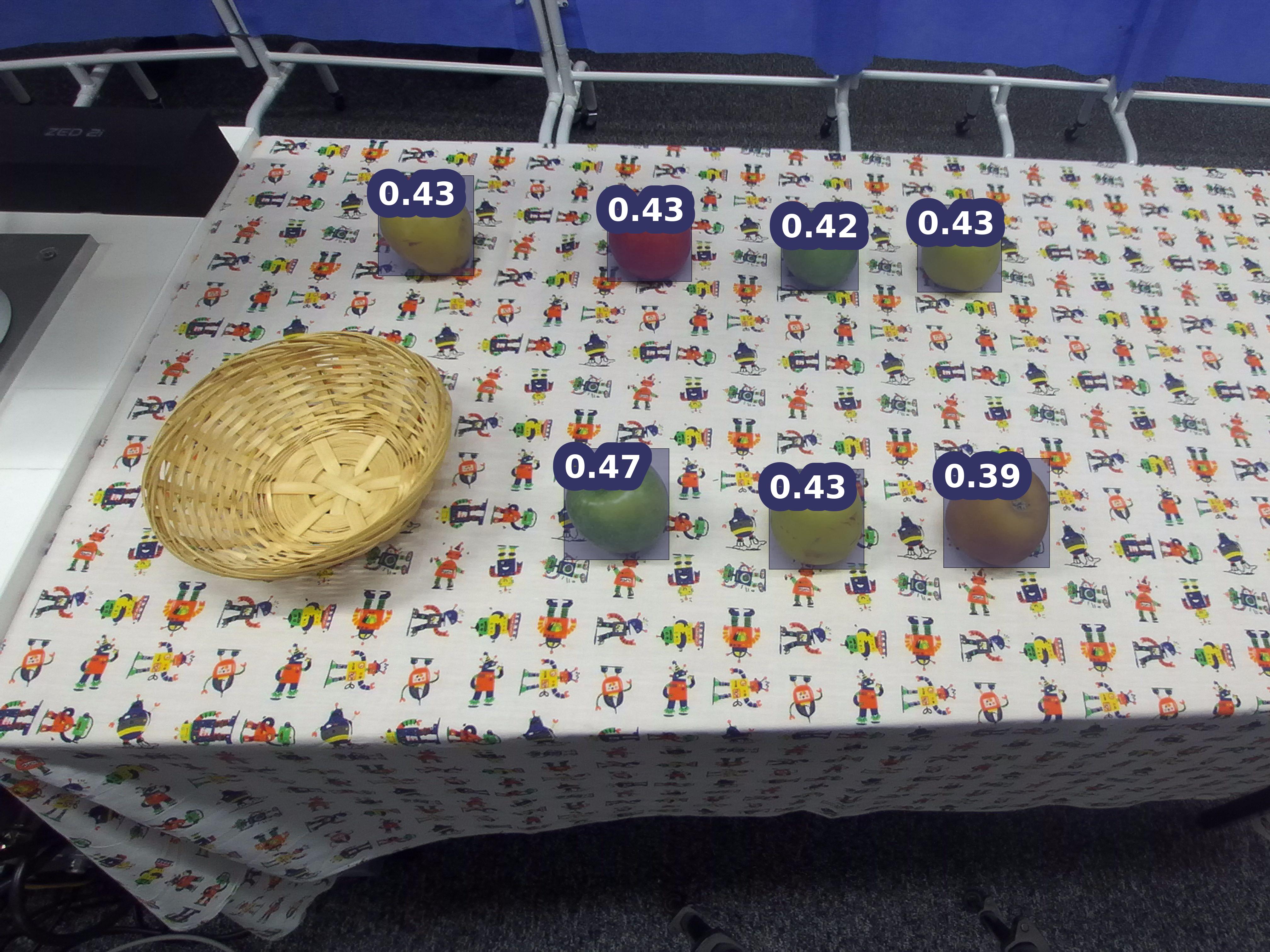} &
\fontsize{8}{9} \selectfont
User: Put the two granny smiths inside the basket. \newline
Agent: Before that, show me the two granny smiths. \newline
User: (designates the two granny smiths) \newline 
Agent: Okay. Let's try to achieve the task \newline
Agent: I have finished the task. \newline
User: Correct. 
&
\includegraphics[width=0.23\textwidth]{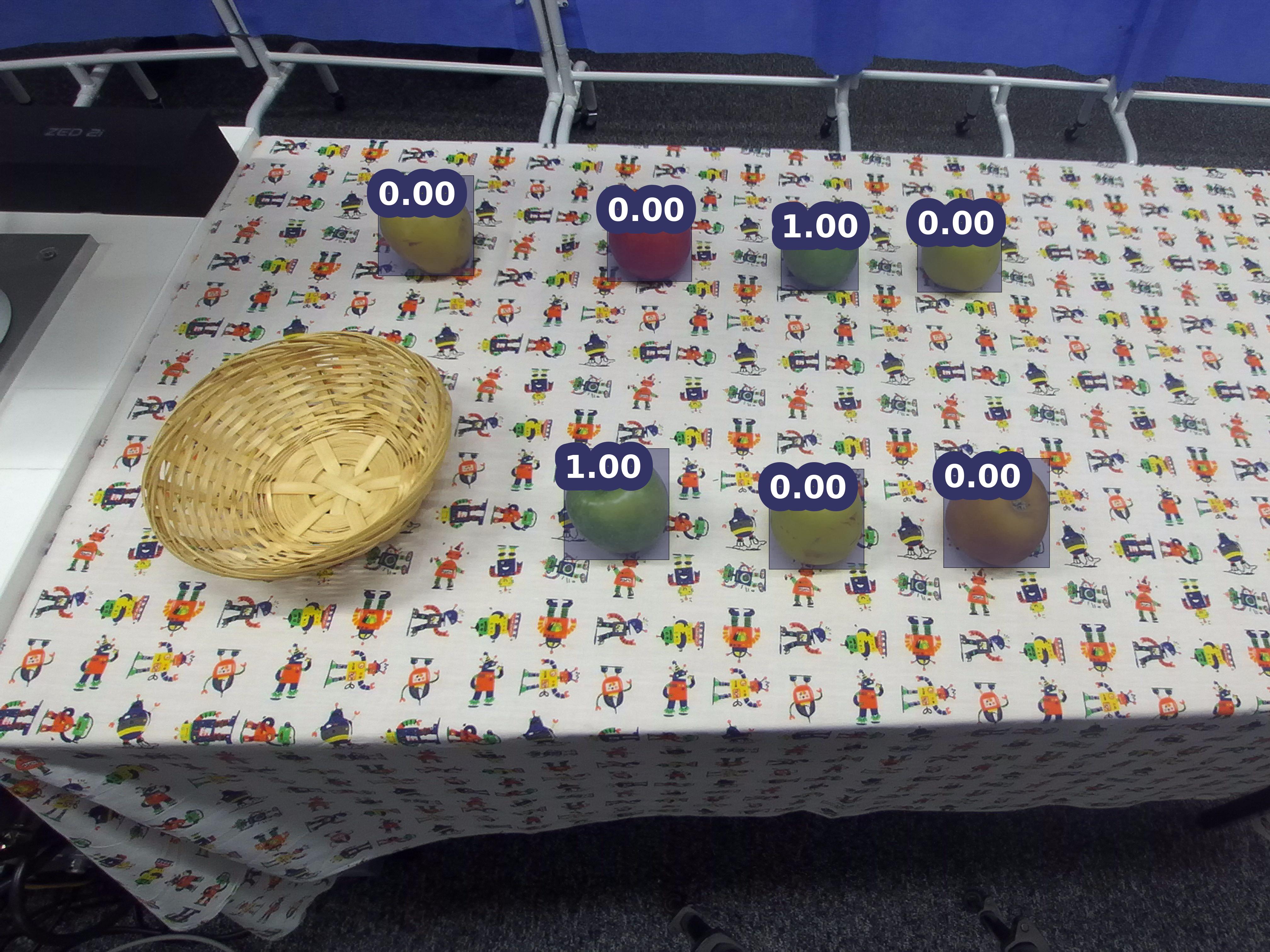} \\

\includegraphics[width=0.23\textwidth]{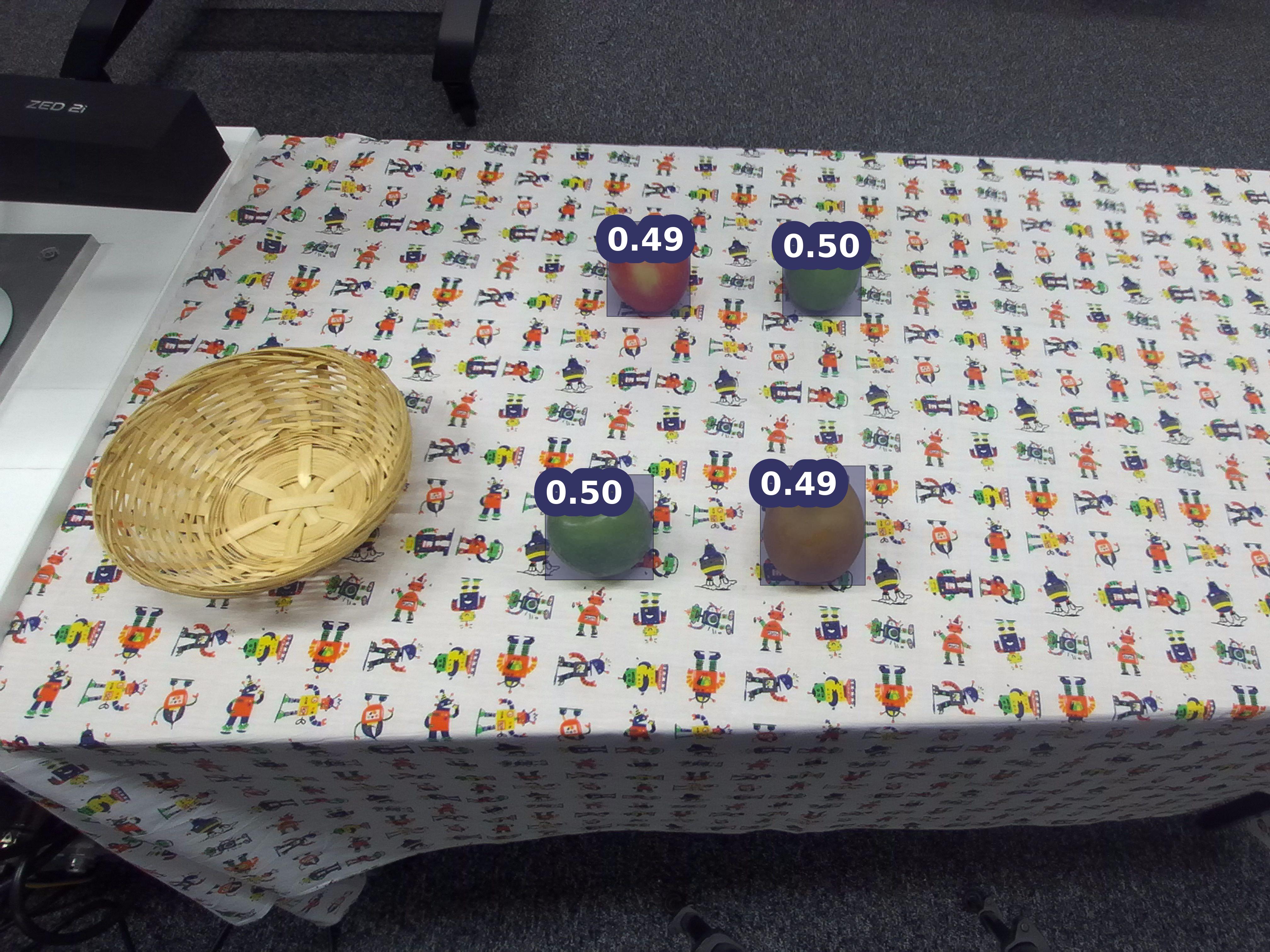} &
\fontsize{8}{9} \selectfont
User: Put the two granny smiths inside the basket. \newline
Agent: Okay. Let's try to achieve the task \newline
Agent: I have finished the task. \newline
User: Correct. &
\includegraphics[width=0.23\textwidth]{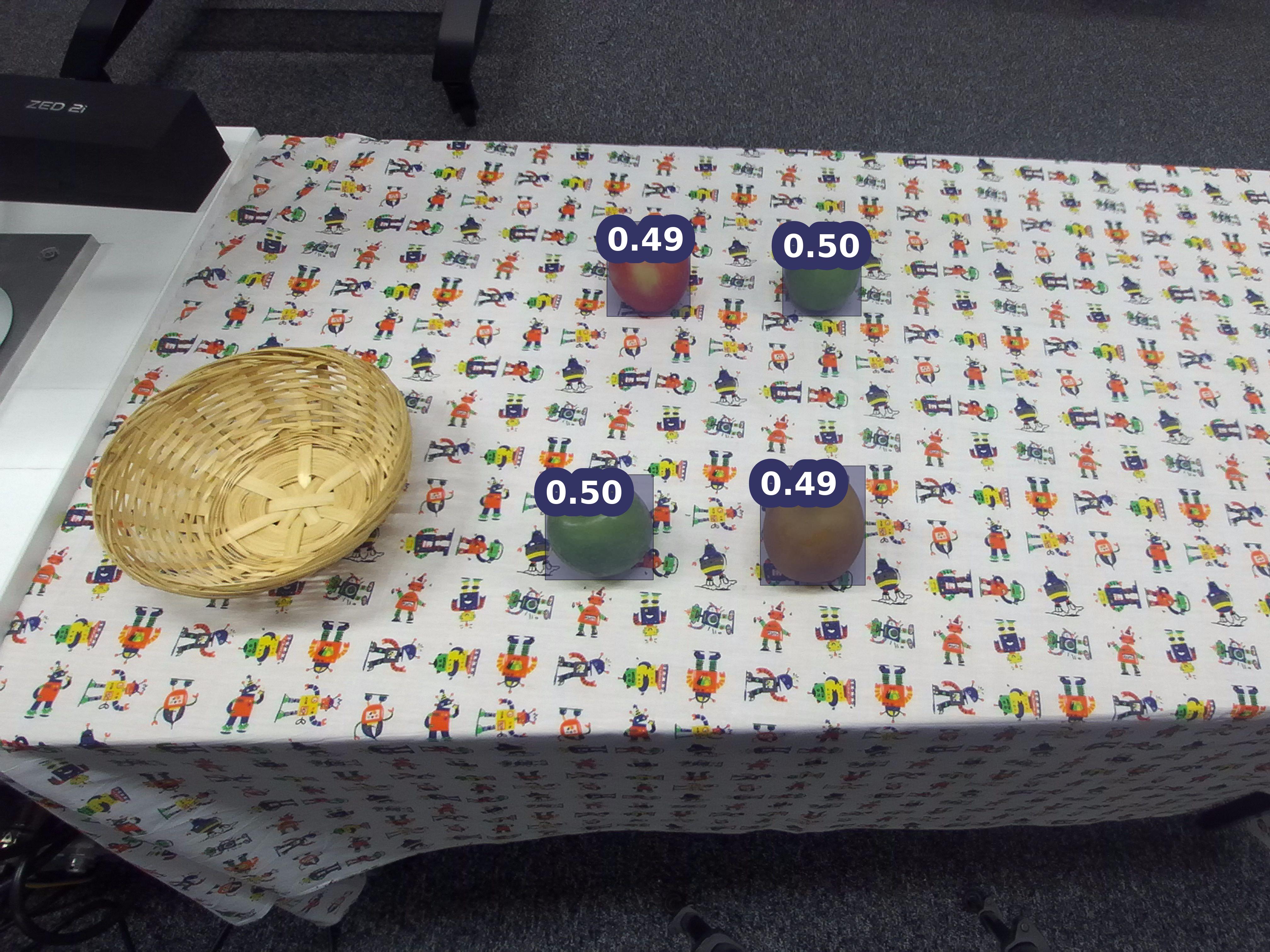} \\ 

\bottomrule
\end{tabular}
\caption{Interaction traces for real-world experiments showing different agent behaviour in two consequent environments when attempting to solve the task instructed. Each trace shows the initial, and updated belief state as well as the embodied conversation the user and agent engage in.}
\label{tab:traces}
\end{table}

\end{document}